\documentclass[10pt,twocolumn,letterpaper]{article}

\usepackage[pagenumbers]{cvpr}

\usepackage{amsmath,amsfonts,bm}

\def\eqref#1{equation~\ref{#1}}

\def\1{\bm{1}}

\DeclareMathAlphabet{\mathsfit}{\encodingdefault}{\sfdefault}{m}{sl}
\SetMathAlphabet{\mathsfit}{bold}{\encodingdefault}{\sfdefault}{bx}{n}

\newcommand{\yes}{\ding{51}}
\newcommand{\no}{\ding{55}}

\usepackage{url}
\usepackage{listings}
\usepackage{tikz}
\usetikzlibrary{backgrounds}
\usepackage{multirow}
\usepackage{pifont}
\usepackage{enumitem}
\usepackage{bbm}
\usepackage{capt-of}
\usepackage{colortbl}
\usepackage{array}
\usepackage{tabularx}
\newcolumntype{L}[1]{>{\raggedright\arraybackslash}p{#1}}%

\definecolor{cvprblue}{rgb}{0.21,0.49,0.74}
\usepackage[colorlinks,allcolors=cvprblue]{hyperref}

\hypersetup{
  pdftitle={HumanPhys: A Human-Centric Physical-World Perception Benchmark for Real-World Visual Reasoning},
  pdfauthor={Yiming Zhang, Enzhi Zhou, Junan Lin, Dulan Zhou, Weiqiang Wang, Yuge Huang, Jian Liu},
}

\def\confName{CVPR}
\def\confYear{2027}

\title{HumanPhys: A Human-Centric Physical-World Perception Benchmark for Real-World Visual Reasoning}
\author{%
Yiming Zhang$^{1,2,\ast}$ \quad
Enzhi Zhou$^{2,\ast}$ \quad
Junan Lin$^{2}$ \quad
Dulan Zhou$^{2}$\\[1pt]
Weiqiang Wang$^{2}$ \quad
Yuge Huang$^{2,\dagger}$ \quad
Jian Liu$^{2,\dagger}$\\[4pt]
\parbox[t]{0.94\textwidth}{\centering\small
$^{1}$Nanyang Technological University, Singapore\quad
$^{2}$Ant Group, Hangzhou, China\\[2pt]
\texttt{yzhang325@e.ntu.edu.sg},\\
\texttt{\{zhouenzhi.zez, linjunan.lja, zhoudulan.zdl, weiqiang.wwq, huangyuge.hyg, rex.lj\}@antgroup.com}\\[2pt]
$^{\ast}$Equal contribution.\quad
$^{\dagger}$Corresponding authors.%
}%
}

\makeatletter
\def\@maketitle{
   \newpage
   \null
   \iftoggle{cvprrebuttal}{\vspace*{-.3in}}{\vskip .375in}
   \begin{center}
      \iftoggle{cvprrebuttal}{{\large \bf \@title \par}}{{\Large \bf \@title \par}}
      \iftoggle{cvprrebuttal}{\vspace*{-22pt}}{\vspace*{24pt}}{
        \large
        \lineskip .5em
        \begin{tabular}[t]{c}
          \iftoggle{cvprfinal}{
            \@author
          }{
            \iftoggle{cvprrebuttal}{}{
              Anonymous \confName~submission
            }
          }
        \end{tabular}
        \par
      }
      \vskip .5em

      \begin{minipage}{\textwidth}
        \centering
        \includegraphics[width=\textwidth]{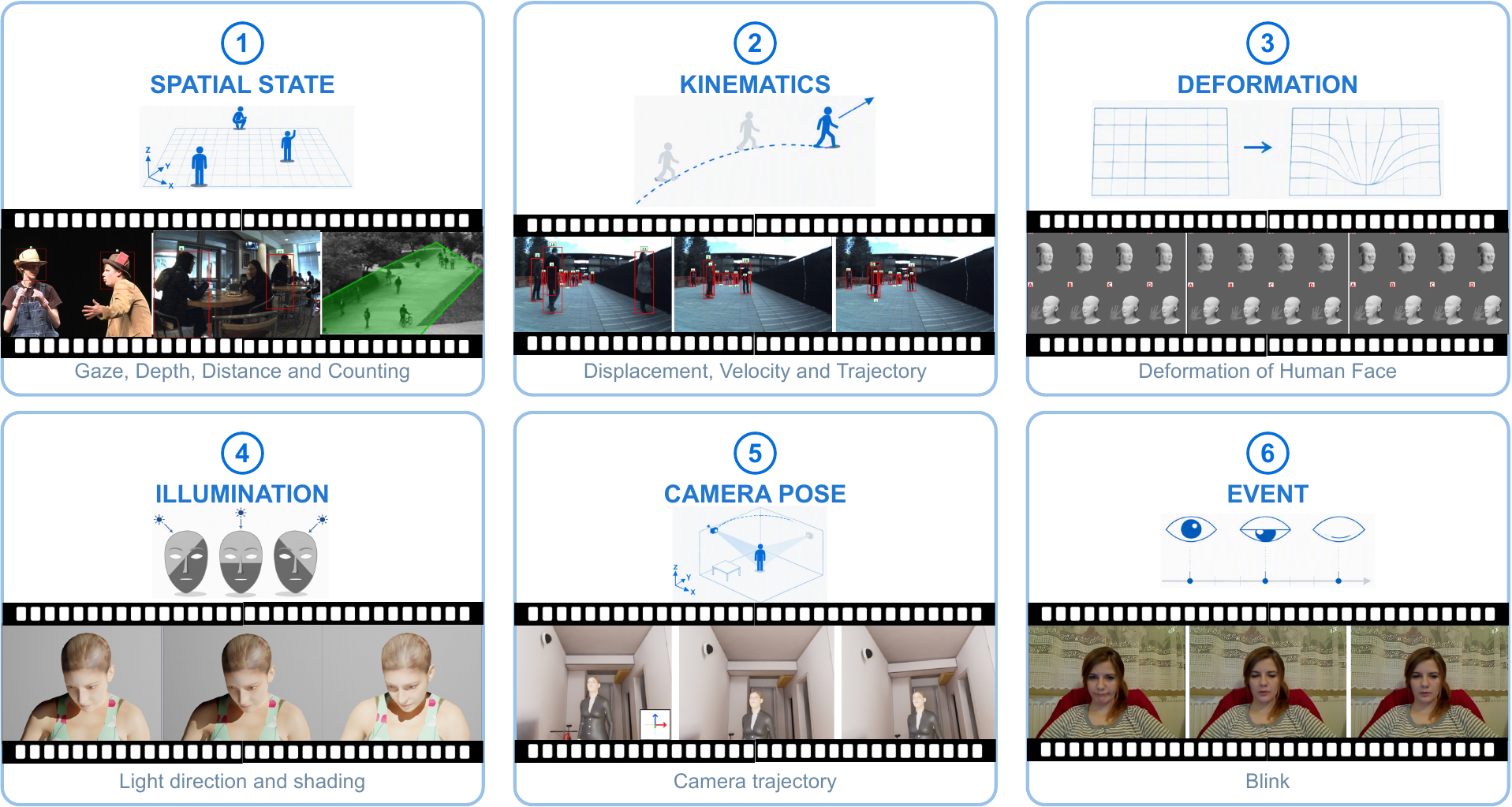}
        \vspace*{4pt}
        \captionof{figure}{Overview of the HumanPhys taxonomy. People are decomposed into measurable physical roles: \textbf{Rigid Mobile Entity}, \textbf{Active Spatial Projector}, \textbf{Non-Rigid Biomechanical Body}, \textbf{Photometric Reflector}, and \textbf{Ego-Motion Anchor}.}
        \label{fig:taxonomy}
      \end{minipage}

      \vspace*{12pt}
   \end{center}
}
\makeatother

\makeatletter
\output=\expandafter{\the\output\setbox\@tempboxa\box255 }
\makeatother

\begin{document}
\maketitle
\begin{abstract}
Can Multimodal Large Language Models (MLLMs) perceive humans as physical entities, rather than merely semantic categories? We study this question with HumanPhys, a diagnostic benchmark suite and training corpus organized around person-anchored physical measurability. HumanPhys spans 13 sub-tasks across six complementary dimensions---spatial state, kinematics, events, deformation, illumination, and camera ego-motion---over real-world videos and synthetic sequences; each task asks for a measurable physical quantity or discrete event, and a paired human-free control isolates camera ego-motion. Answer labels are programmatically derived from sensor calibration, source metadata, or renderer state, with task-specific noise sources explicitly disclosed, rather than manually guessed from benchmark frames; the two exceptions where an upstream fit contributes are disclosed per sub-task (\textit{Trajectory}'s triage classifier, \textit{Deform}'s DECAF anchor). We benchmark a broad set of open- and closed-source MLLMs and show that the six dimensions are only partially dissociable and that all remain challenging. Beyond evaluation, we show that HumanPhys labels and deterministic scorers support supervised training, reward-based optimization, and tool-assisted inference. Project homepage: \url{https://zhangym0213.github.io/HumanPhys/}.
\end{abstract}

\begin{figure*}[t]
\centering
\includegraphics[width=0.86\linewidth]{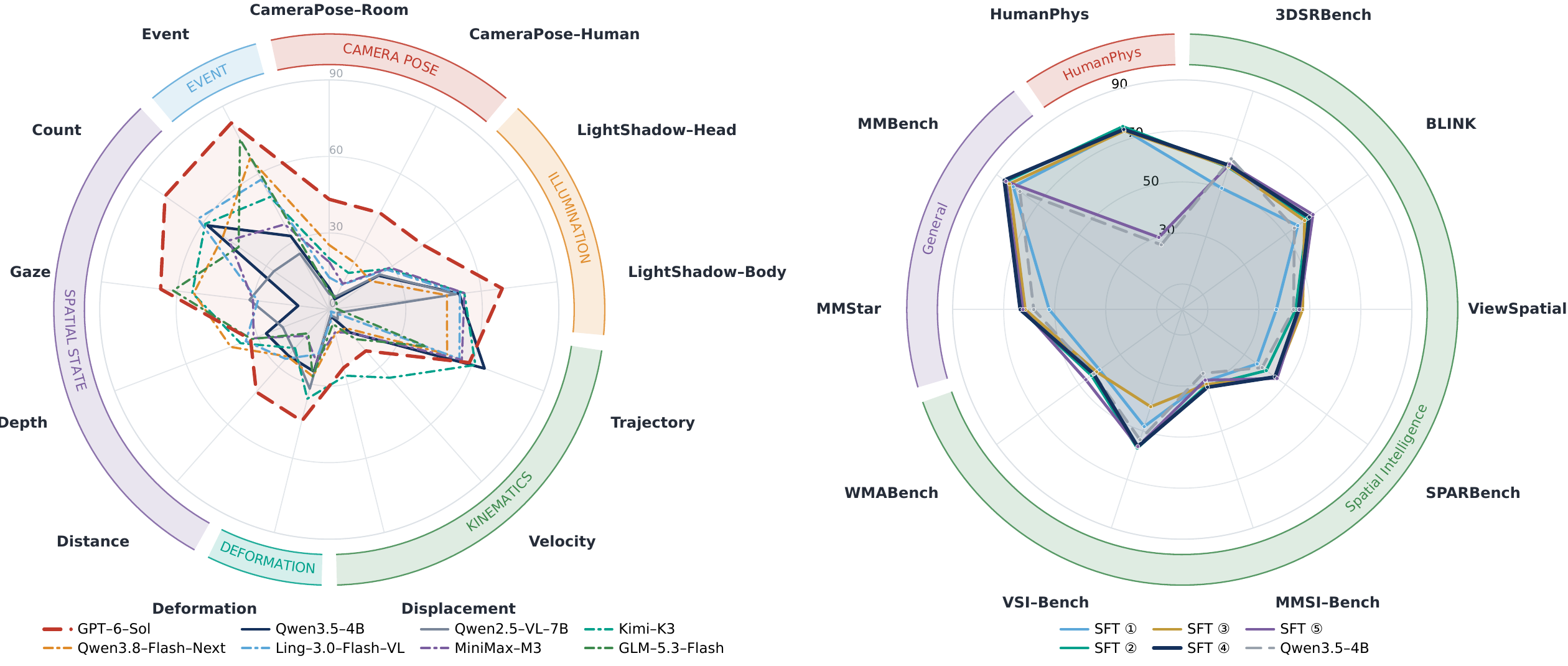}
\caption{\textbf{Left}: zero-shot performance radar of eight of the twelve evaluated MLLMs across the 13 sub-tasks---the closed-source GPT-6-Sol (bold red dashes), the two locally served base models (solid), and the five hosted-API models (dash-dot); the four spatial-intelligence fine-tunes of Table~\ref{tab:main_results} are tabulated only, for legibility (numerical results in Table~\ref{tab:main_results}). \textbf{Right}: data-mixture ablation with Qwen3.5-4B across ten benchmarks (tabulated in Table~\ref{tab:ablation} and analysed in \S\ref{sec:mixture}).}
\label{fig:radar}
\end{figure*}

\section{Introduction}
The recent rise of spatial intelligence benchmarks has raised a natural question: can Multimodal Large Language Models (MLLMs)~\citep{qwen35, internvl3, kimivl, glm45v, gemini25, mllmsurvey}, which perform strongly on 2D visual understanding, also perceive and reason about the physical world? Recent work has consequently explored physical spatial perception from object attributes (count, size) to inter-object relations (position, distance)~\citep{vsibench, 3dsrbench, blink, viewspatial, mmsibench, spatialscore}. These efforts have progressively expanded spatial-intelligence evaluation from semantic recognition toward explicit geometric understanding of objects and their spatial configurations. In parallel, recent training efforts have tried to elicit such reasoning through reinforcement learning over spatial rewards or tool-augmented inference~\citep{spacer, spatialssrl, vilasr, sagent, glm5vturbo, spatialthinker}. Yet most existing spatial benchmarks remain object- or scene-centric, leaving human-anchored quantitative physical perception underexamined.

However, understanding the physical world requires more than treating scenes as collections of rigid objects: humans are dynamic, living physical entities whose perception demands more than bounding-box geometry. Estimating only a person's depth or velocity reduces the person to a moving cylinder. Going beyond coarse person geometry requires measuring multiple physical roles across scales: (i) a \textbf{Rigid Mobile Entity}, positioned in and moved through metric space; (ii) an \textbf{Active Spatial Projector}, whose gaze links the head to an external fixation target as a directional geometric vector; (iii) a \textbf{Non-Rigid Biomechanical Body}, exhibiting contact deformation and transient events such as eye blinking; and (iv) a \textbf{Photometric Reflector and Ego-Motion Anchor}, where the body carries illumination information and anchors an observer's camera trajectory. Public human-centric resources capture several of these cues in isolation~\citep{decaf, eyeblink8, bedlam, DyVim}. We do not collapse these roles into a single latent skill; rather, HumanPhys organizes them under a shared criterion of person-anchored physical measurability.

Despite this progress, coverage of quantitative human physical perception remains narrow and uneven rather than absent. Video benchmarks mainly emphasize activities, actions, and events~\citep{humanvbench, humanvideomme, humanmme, maben, finebench}; spatial benchmarks focus on object- and scene-centric attributes and relations~\citep{sparbench, stibench, wmabench}; and even when humans appear, they usually serve as semantic subjects or action performers rather than physical entities whose states can be measured quantitatively. HumanMoveVQA~\citep{humanmovevqa} is the closest prior benchmark and asks geometric questions about the root translation and orientation of a single target person through discretized multiple-choice labels derived from estimated mesh tracks (\S\ref{sec:related_human}). By reducing a person to a coarse root trajectory, however, it leaves unmeasured multi-person metric relations, directional gaze vectors, transient blinks, contact-induced deformation, body-cast illumination, and camera ego-motion anchored by a person in the scene. HumanPhys therefore extends human-centric physical evaluation not merely by adding more task categories, but by changing the unit of supervision: from coarse, discretized motion labels to open-ended measurable quantities and event counts. It derives answer labels programmatically from source metadata or renderer state, with task-specific residual noise disclosed rather than hidden.

\begin{table}[t]
\centering
\caption{\textbf{Comparison of representative spatial reasoning benchmarks.} \textbf{Human-Centric}: humans as perception targets; \textbf{Video}: temporally sampled frames; \textbf{Static/Dynamic}: frame-level vs.\ temporal queries (\S\ref{sec:tasks}). For HumanMoveVQA, \textbf{Train~Data~Ava.} refers to its internal training split (detailed comparison in \S\ref{sec:related_human} and Appendix~\ref{app:closest}).}
\label{tab:benchmark_comparison}
\setlength{\tabcolsep}{3pt}
\resizebox{\linewidth}{!}{%
\begin{tabular}{lcccccccc}
\toprule
\multirow{2}{*}{Benchmark} & \multirow{2}{*}{Human-Centric}
 & \multicolumn{2}{c}{Input Type} & \multicolumn{2}{c}{Source} & \multicolumn{2}{c}{Task Taxonomy}
 & \multirow{2}{*}{\shortstack{Train\\Data~Ava.}} \\
\cmidrule(lr){3-4}\cmidrule(lr){5-6}\cmidrule(lr){7-8}
 & & Image & Video & Real & Simulation & Static & Dynamic & \\
\midrule
VSI-Bench~\citep{vsibench}           & \no  & \no  & \yes & \yes & \no  & \yes & \no  & \no  \\
STI-Bench~\citep{stibench}           & \no  & \no  & \yes & \yes & \no  & \yes & \yes & \no  \\
3DSRBench~\citep{3dsrbench}           & \no  & \yes & \no  & \yes & \yes & \yes & \no  & \no  \\
BLINK~\citep{blink}               & \no  & \yes & \no  & \yes & \yes & \yes & \no  & \no  \\
ViewSpatial~\citep{viewspatial}         & \no  & \yes & \no  & \yes & \no  & \yes & \no  & \no  \\
SPARBench~\citep{sparbench}           & \no  & \yes & \no  & \yes & \no  & \yes & \no  & \no  \\
MMSI-Bench~\citep{mmsibench}          & \no  & \yes & \no  & \yes & \no  & \yes & \yes & \no  \\
WMABench~\citep{wmabench}            & \no  & \yes & \yes & \no  & \yes & \yes & \yes & \no  \\
SpaceR~\citep{spacer}              & \no  & \no  & \yes & \yes & \no  & \yes & \no  & \yes \\
SpatialScore~\citep{spatialscore}        & \no  & \yes & \yes & \yes & \yes & \yes & \yes & \yes \\
HumanMoveVQA~\citep{humanmovevqa} & \yes & \no & \yes & \yes & \no & \no & \yes & \yes \\
\midrule
\textbf{HumanPhys} & \textbf{\yes} & \textbf{\yes} & \textbf{\yes} & \textbf{\yes} & \textbf{\yes} & \textbf{\yes} & \textbf{\yes} & \textbf{\yes} \\
\bottomrule
\end{tabular}%
}
\end{table}

\begin{figure*}[t]
\centering
\includegraphics[width=0.84\linewidth]{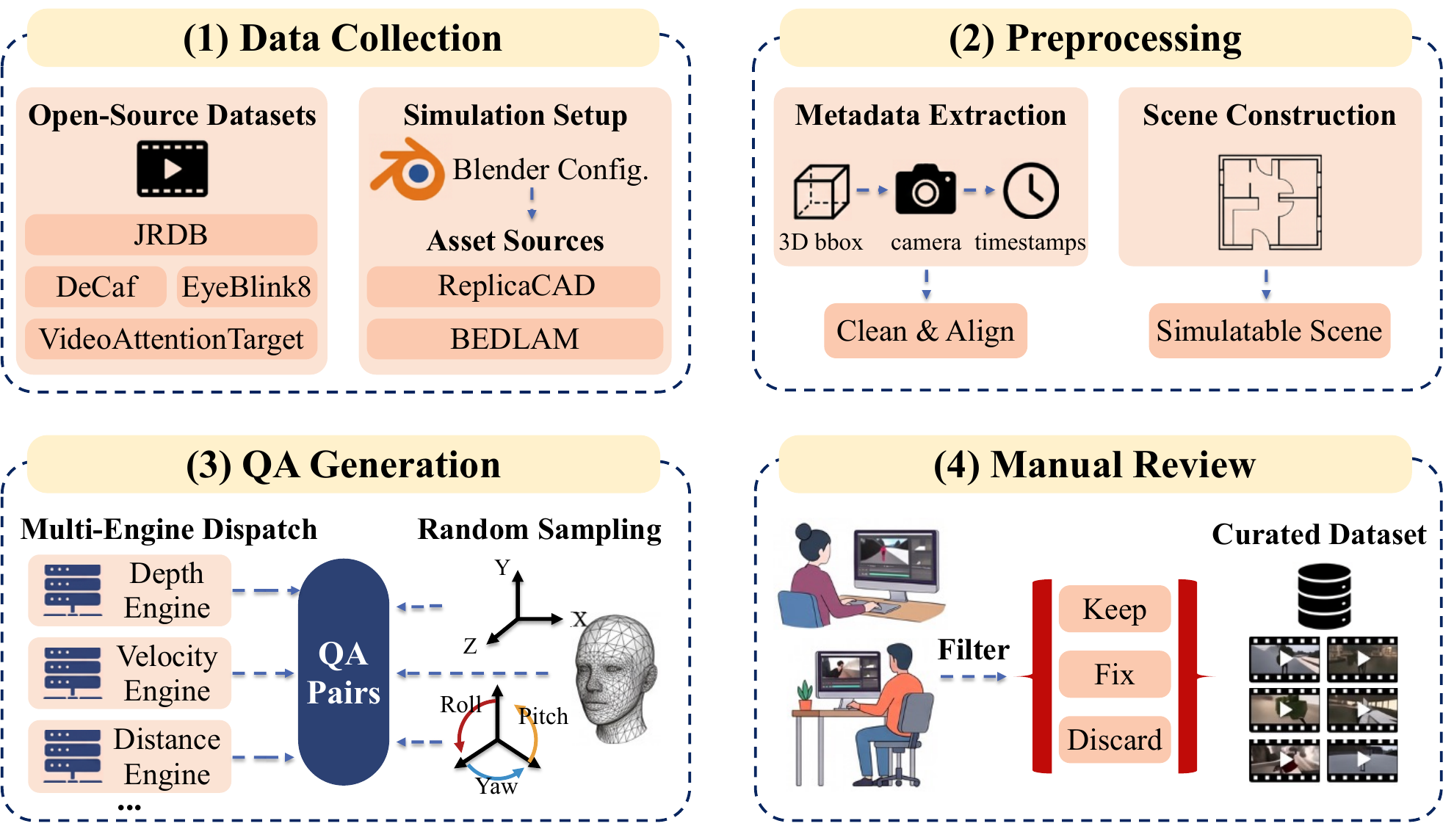}
\caption{The construction pipeline of HumanPhys: (1) data collection from open-source datasets and Blender-based simulation; (2) preprocessing with metadata extraction and scene construction; (3) multi-engine question answering (QA) generation with random sampling; (4) manual review and iterative refinement.}
\label{fig:pipeline}
\end{figure*}

\begin{figure*}[t]
\centering
\begin{minipage}[t]{0.48\linewidth}
\vspace{0pt}
\centering
\includegraphics[width=\linewidth]{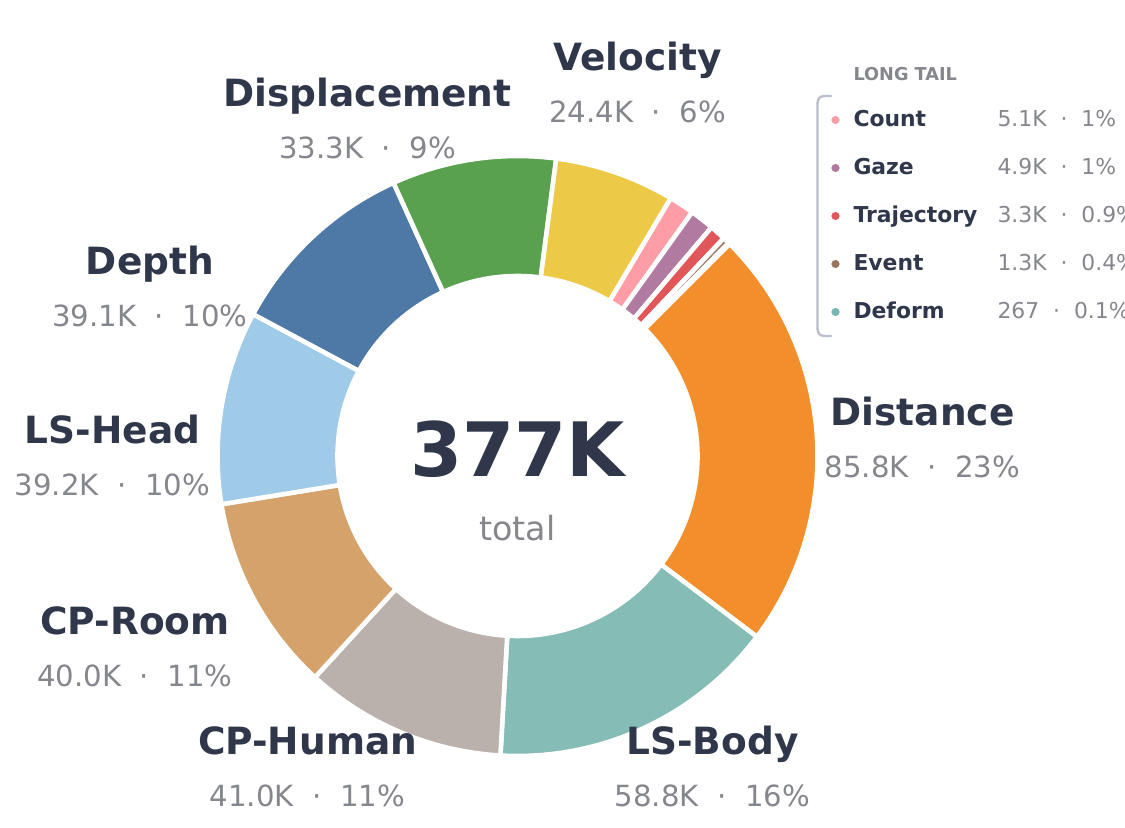}
\captionof{figure}{Task distribution of the HumanPhys training corpus (about
377K samples), distance-related tasks dominate, with long-tail categories such as Count, Gaze, Trajectory and Deformation.}
\label{fig:statistic}
\end{minipage}\hfill
\begin{minipage}[t]{0.51\linewidth}
\vspace{0pt}
\centering
\scriptsize
\setlength{\tabcolsep}{1.3pt}
\resizebox{\linewidth}{!}{\begin{tabular}{@{}clllp{42pt}lc@{}}
\toprule
Group & Dimension & Sub-task & Source & Input & Format \\
\midrule
\multirow{8}{*}{I} & \multirow{4}{*}{Spatial state}
  & Depth      & JRDB-Activity        & frame/video & Num \\
  & & Distance   & JRDB-Activity        & frame/video & Num \\
  & & Count      & DanceTrack, UCSD     & frame/video & Num \\
  & & Gaze       & Video\hspace{0pt}Attention\hspace{0pt}Target & frame & Num \\
\cmidrule(lr){2-6}
  & \multirow{3}{*}{Kinematics}
  & Displacement & JRDB-Activity        & video       & Num \\
  & & Velocity     & JRDB-Activity        & video       & Num \\
  & & Trajectory   & JRDB-Activity        & video       & Tuple \\
\cmidrule(lr){2-6}
  & Event & Blink      & EyeBlink8            & video       & Num \\
\midrule
\multirow{5}{*}{II} & Deformation & Deform   & DECAF (FLAME, MANO)  & video & MC \\
\cmidrule(lr){2-6}
  & \multirow{2}{*}{Illumination}
  & LS-Head  & 200 BEDLAM identities & video & Tuple \\
  & & LS-Body  & BEDLAM, 8 wall cams & video & Tuple \\
\cmidrule(lr){2-6}
  & \multirow{2}{*}{Camera pose}
  & CP-Room\,(human-free) & ReplicaCAD (84 stages) & video & Tuple \\
  & & CP-Human\,(anchored) & ReplicaCAD (84 stages) & video & Tuple \\
\bottomrule
\end{tabular}}
\par\vspace{\abovecaptionskip}
\captionof{table}{\textbf{HumanPhys composition.}
\textbf{Group}: I, measured human cues on real-world footage; II, controlled rendered probes for phenomena real footage cannot be labelled for reliably.
\textbf{Dimension}: taxonomy category of Figure~\ref{fig:taxonomy}; \textbf{Sub-task}: the queried quantity; \textbf{Source}: dataset or renderer.
\textbf{Input}: \textit{frame}, one image; \textit{video}, ordered frames.
\textbf{Format}: Num, scalar scored by mean relative accuracy (MRA) or the angular scorer; MC, multiple-choice option accuracy; Tuple, sequence scorers of \S\ref{sec:protocol}.}
\label{tab:composition}
\end{minipage}
\end{figure*}

\begin{table*}[t]
\centering
\scriptsize
\setlength{\tabcolsep}{1.2pt}
\caption{\textbf{Main results on HumanPhys.} Best per-column scores among evaluated MLLMs are in \textbf{bold}. \textit{Dist} = Distance, \textit{Disp} = Displacement, \textit{Traj} = Trajectory, \textit{LS} = Light-Shadow, \textit{CP} = Camera-Pose; SpaceR, SpatialEvo, SSRL (Spatial-SSRL-7B), and ViLaSR are Qwen2.5-VL-7B fine-tunes. Open-source models are grouped into \textbf{Base Models}, \textbf{For Spatial Intelligence}, and \textbf{API}; the API group contains open-source models accessed through hosted API endpoints under the same prompts and scoring protocol. \textbf{Overall} is the question-weighted average over all 2{,}500 items (unparseable outputs score~$0$); the \textit{Human} row is the one exception: it pools the 647 voluntary reference responses that the scorer could evaluate (Appendix~\ref{app:taskdetails}, \ref{app:human_study}) rather than question-weighting its columns, which would give 52.51. \textbf{Random} averages 50 blind-guess seeds over training priors; \textbf{Text Only} queries Qwen3.5-4B without visual input.}
\label{tab:main_results}
\begin{tabular*}{\linewidth}{@{}l@{\extracolsep{\fill}}cccccccccccccc@{}}
\toprule
 & \multicolumn{4}{c}{Spatial state} & \multicolumn{3}{c}{Kinematics} & Event & Deform. & \multicolumn{2}{c}{Illumination} & \multicolumn{2}{c}{Camera pose} & \\
\cmidrule(lr){2-5}\cmidrule(lr){6-8}\cmidrule(lr){9-9}\cmidrule(lr){10-10}\cmidrule(lr){11-12}\cmidrule(lr){13-14}
Model & Depth & Dist & Count & Gaze & Disp & Vel & Traj & Blink & Deform & LS-Head & LS-Body & CP-Room & CP-Human & Overall \\
\midrule
Random & 17.13 & 9.69 & 22.15 & 13.62 & 4.31 & 5.06 & 24.27 & 18.70 & 24.28 & 27.08 & 53.47 & 8.80 & 5.29 & 17.74 \\
Text Only & 10.35 & 1.70 & 0.00 & 0.00 & 0.00 & 0.00 & 18.20 & 0.00 & 23.00 & 20.86 & 52.35 & 8.61 & 3.73 & 10.18 \\
Human & 46.38 & 33.75 & 79.47 & 58.00 & 37.31 & 32.35 & 60.42 & 77.88 & 36.00 & 52.97 & 70.07 & 50.46 & 39.29 & 52.48 \\
\midrule
\multicolumn{15}{@{}l@{}}{\cellcolor{blue!10}\textbf{Closed-source models}}\\
GPT-6-Sol & 33.00 & \textbf{43.15} & \textbf{78.25} & \textbf{66.60} & 23.50 & 21.65 & 58.75 & \textbf{82.60} & \textbf{45.00} & \textbf{44.60} & \textbf{68.31} & \textbf{43.20} & \textbf{42.75} & \textbf{50.31} \\
\midrule
\multicolumn{15}{@{}l@{}}{\cellcolor{green!12}\textbf{Open-source models (Base Models)}}\\
Qwen3.5-4B~\citep{qwen35} & 26.45 & 24.15 & 58.00 & 12.30 & 2.70 & 11.95 & 65.10 & 32.55 & 25.00 & 23.44 & 51.29 & 8.43 & 4.64 & 26.68 \\
Qwen2.5-VL-7B~\citep{qwen25vl} & 19.50 & 21.10 & 26.40 & 31.50 & 2.30 & 4.95 & 3.53 & 24.80 & 32.00 & 24.13 & 50.08 & 6.13 & 5.83 & 18.90 \\
\midrule
\multicolumn{15}{@{}l@{}}{\cellcolor{green!12}\textbf{Open-source models (For Spatial Intelligence)}}\\
ViLaSR~\citep{vilasr} & 5.00 & 6.35 & 47.25 & 4.60 & 23.40 & \textbf{53.50} & 66.51 & 27.80 & 28.00 & 16.03 & 47.52 & 6.37 & 6.58 & 25.99 \\
SpaceR~\citep{spacer} & 23.70 & 15.95 & 22.25 & 27.00 & 8.45 & 16.00 & 66.11 & 28.20 & 26.00 & 25.17 & 50.17 & 6.69 & 9.87 & 25.01 \\
SpatialEvo~\citep{spatialevo} & 11.70 & 13.55 & 34.20 & 3.90 & 21.35 & 24.40 & \textbf{67.16} & 4.10 & 25.00 & 21.31 & 44.11 & 7.38 & 5.67 & 21.71 \\
SSRL~\citep{spatialssrl} & 20.95 & 17.25 & 29.45 & 25.50 & 0.00 & 3.40 & 5.76 & 36.85 & 28.00 & 27.38 & 51.75 & 8.26 & 6.48 & 19.76 \\
\midrule
\multicolumn{15}{@{}l@{}}{\cellcolor{green!12}\textbf{Open-source models (API)}}\\
Kimi-K3 & 37.20 & 20.20 & 59.10 & 54.10 & \textbf{26.70} & 35.70 & 61.26 & 49.85 & 36.00 & 28.12 & 53.33 & 20.04 & 16.21 & 38.38 \\
Qwen3.8-Flash-Next & \textbf{41.30} & 24.60 & 50.60 & 53.50 & 9.80 & 9.70 & 49.40 & 66.85 & 27.00 & 19.69 & 46.47 & 25.28 & 20.89 & 34.53 \\
Ling-3.0-Flash-VL & 35.00 & 25.80 & 62.60 & 28.20 & 3.80 & 1.00 & 54.54 & 57.50 & 18.00 & 27.49 & 51.59 & 12.57 & 11.03 & 30.41 \\
MiniMax-M3 & 31.75 & 13.75 & 47.70 & 30.10 & 9.25 & 12.05 & 55.58 & 37.70 & 21.00 & 28.93 & 53.35 & 18.61 & 11.39 & 28.85 \\
GLM-5.3-Flash & 32.65 & 12.45 & 43.15 & 61.70 & 5.85 & 15.50 & 40.75 & 75.35 & 27.00 & 3.56 & 2.81 & 7.15 & 4.99 & 25.55 \\
\bottomrule
\end{tabular*}
\end{table*}

To bridge this gap, we introduce \textbf{HumanPhys}, a diagnostic benchmark suite and corpus for human-centered physical perception in MLLMs on real-world and rendered videos, spanning spatial state, kinematics, events, deformation, illumination, and camera ego-motion. Every question is scored as a physical quantity or event count, while a paired human-free control isolates observer ego-motion. Our evaluation shows that the six dimensions are only partially dissociable: models specialized for coarse 3D geometry surpass proprietary frontier models on whole-body trajectory, yet score near floor on fine-grained directional gaze and transient blink events (\S\ref{sec:findings}), so coarse spatial training does not automatically transfer to human physical perception.

In summary, our contributions are as follows.
(1) We present HumanPhys, a diagnostic benchmark suite and training corpus for person-anchored physical perception, organized around measurable physical quantities and discrete events with a paired human-free ego-motion control.
(2) We provide a systematic evaluation of MLLMs on this benchmark and show that HumanPhys dimensions expose complementary failure modes rather than reducing to a single spatial-ability factor.
(3) Because HumanPhys provides deterministic scorers and actionable labels, we validate their utility through supervised data mixtures, metric-guided reinforcement learning with verifiable rewards (RLVR), and training-free geometric-tool use; these studies are utility validations rather than separate method contributions.

\section{Related Work}
\subsection{Spatial Intelligence}
Spatial intelligence probes whether MLLMs can recover geometric and physical properties beyond semantic recognition: object positions, sizes, orientations, distances and metric QA \citep{vsibench, 3dsrbench, blink, viewspatial, spatial457, statesight, spatialquery, embspatial}; simulator- or 3D-annotation-based supervision enables scalable construction and fine-tuning \citep{spatial457, spatialscore, qwen3d, spatialvid, openspatial, spatialvlm, spatialbot, spatialmllm}; region-level cues support object-centric analysis \citep{spatialrgpt}; video extensions infer layouts and dynamics from frame sequences \citep{vsibench, stibench, spacer, ovosbench, crossviewsuite}; and recent benchmarks stress spatial intelligence under visual degradation and low-level embodied viewpoints~\citep{spacedg, embodied3dbench}.

Despite these advances, existing spatial data remain predominantly object- or scene-centric, including comprehensive consolidations such as SpatialScore \citep{spatialscore}, and therefore rarely treat humans as physical entities whose states, motion, and interactions can be quantitatively perceived from videos; HumanPhys is designed to target this gap. Even SpatialScore's 331K-sample corpus and SpaceR-151K measure object and scene geometry rather than human body properties---omitting inter-person distance, physiological events, contact deformation, and body-cast illumination. We therefore use them as external complements in \S\ref{sec:mixture} rather than as competing breadth-oriented resources.

\subsection{Human-Centric Perception and Understanding}
\label{sec:related_human}
Existing human-centric video benchmarks mainly target semantic, behavioral, or cognitive understanding, while HumanMoveVQA~\citep{humanmovevqa} is the closest prior work to ours because it asks geometric questions about human motion. HumanMoveVQA evaluates global trajectory and orientation reasoning for a single target person from exocentric video through discretized multiple-choice questions derived from the SMPL-X tracks of its source datasets. HumanPhys is complementary rather than redundant and differs from HumanMoveVQA in measured quantities, answer format and scoring, ground-truth provenance, and generalization protocol, as detailed in Table~\ref{tab:closest} of Appendix~\ref{app:closest}. Table~\ref{tab:benchmark_comparison} positions HumanPhys among them.

\section{HumanPhys}
\label{sec:benchmark}

HumanPhys evaluates whether an MLLM can recover physical information from human-centric images and videos along six dimensions: spatial state, kinematics, event, deformation, illumination (light--shadow), and camera pose (Figure~\ref{fig:taxonomy}). Each answer is programmatically derived from source metadata or renderer state, including 3D boxes, camera parameters, timestamps, and source annotations for the real-world datasets and full scene state for the synthetic ones. This avoids frame-level annotator guessing at label time, but it does not imply zero label noise: inherited upstream or pipeline noise is disclosed per sub-task (provenance in \S\ref{sec:tasks} and Appendix~\ref{app:taskdetails}). In particular, \textit{Trajectory} inherits residual noise from its triage classifier, while \textit{Deform} is anchored to DECAF's monocular fit. HumanPhys spans \textbf{13 sub-tasks} over \textbf{8 sources} (5
real-world video datasets and 3 rendered scene collections), and includes a
\textbf{2,500-question evaluation split drawn by equal per-sub-task quota
sampling (200 questions, 100 for \textit{Deform})}, plus a
\textbf{376,582-question} training pool emitted by the same construction
pipeline. Training and evaluation are split by disjoint source groups, with question-level quotas sampled only from held-out groups; \S\ref{sec:construction} states the split-and-sampling rule and Appendix~\ref{app:split} gives the full split details. We evaluate twelve models zero-shot (Table~\ref{tab:main_results}).

These six dimensions operationalize the physical roles introduced in the Introduction under a single unifying criterion---\textbf{measurability}: every answer is a quantity in physical units or a discrete event count read programmatically from metadata or renderer state. All sub-tasks are anchored on a person, except \textit{CP-Room}, which serves as a paired controlled human-free baseline for ego-motion. This taxonomy is diagnostic rather than merely broad, revealing distinct cross-model profiles (\S\ref{sec:findings}) and enabling step-level failure attribution across motion axes (\S\ref{sec:diagnosis}).

\subsection{Task Definition}
\label{sec:tasks}

The 13 sub-tasks form two complementary groups (Table~\ref{tab:composition}). \textbf{Group~I} evaluates real-world physical cues recovered from sensor metadata on actual footage. \textbf{Group~II} uses controlled rendering (or FLAME/MANO re-rendering over real DECAF fits) to probe contact deformation, hidden-light walks, and atomic camera ego-motion that cannot be reliably annotated in the wild, with labels read from the simulator state.

\paragraph{Group I: real-world measured human-centered physical cues.}
\begin{itemize}[leftmargin=1.2em,itemsep=1pt]
  \item \textbf{Spatial state}: \textit{Depth} --- distance of people to the reference sensor; \textit{Distance} --- pairwise inter-person distance at the foot point; \textit{Count} --- humans satisfying an explicitly stated spatial criterion; \textit{Gaze} --- clockwise image-plane angle from upward to the head-centre--to--fixation vector.
  \item \textbf{Kinematics}: \textit{Displacement} --- cumulative ground-plane path length of a tracklet; \textit{Velocity} --- that path length over the clip's elapsed time; \textit{Trajectory} --- heading changes and step lengths as $(\text{turn\_deg},\text{distance\_m})$ tuples.
  \item \textbf{Event}: \textit{Blink} --- number of discrete eye-blink events in a clip, integrating evidence across frames.
\end{itemize}

\paragraph{Group II: controlled physically rendered probes.}
\begin{itemize}[leftmargin=1.2em,itemsep=1pt]
  \item \textbf{Deformation} (\textit{Deform}): among a $2\times4$ panel of a FLAME head deformed by MANO hand contact at four strengths, pick the strength whose deformation best matches the physically plausible one.
  \item \textbf{Illumination} (\textit{LS-Head}, \textit{LS-Body}): infer the lattice-coordinate walk of a single hidden point light from shadow motion on the face or on the body and floor, respectively.
  \item \textbf{Camera pose} (\textit{CP-Room}, \textit{CP-Human}): recover the sequence of atomic camera actions --- lattice translations or in-place rotations --- across indoor trajectories. \textit{CP-Room} is a controlled human-free baseline (control group) matched to \textit{CP-Human} in scene family, 0.20\,m lattice step and nine-token action grammar, but not in translation scoring: \textit{CP-Room} grades direction only (one lattice step per translation), whereas \textit{CP-Human} allows 1--3 steps and also grades the translated magnitude (Appendix~\ref{app:scoring_details}). \textit{CP-Human} places a static SMPL-X person in the frame, testing whether a human anchor helps or distracts ego-motion recovery.
\end{itemize}

Appendix~\ref{app:taskdetails} provides detailed construction procedures for each sub-task, including ego-motion filtering, coordinate and angle conventions, blink-window compression.

\subsection{Benchmark Construction}
\label{sec:construction}

As illustrated in Figure~\ref{fig:pipeline}, the construction of HumanPhys follows a four-stage pipeline: data collection, preprocessing, QA generation, and manual review. In the first stage, we aggregate open-source human-centric datasets (e.g., JRDB~\citep{jrdbact}, DECAF~\citep{decaf}, EyeBlink8~\citep{eyeblink8}, VideoAttentionTarget~\citep{videoattentiontarget}, DanceTrack~\citep{dancetrack} and UCSD~\citep{ucsd}) together with synthetic scenes configured in Blender and using assets from sources such as ReplicaCAD~\citep{replicacad} in Habitat-Sim~\citep{habitat} and BEDLAM~\citep{bedlam}. The collected videos are then preprocessed by extracting metadata (3D bounding boxes, camera parameters, and timestamps) and constructing simulatable scenes, followed by cleaning and alignment.

\paragraph{Group-level split and sampling validity.}
Splits are made before question-level sampling and at the source-\textbf{group} level, rather than by randomly partitioning questions. A source group denotes shared provenance---video, sequence, session, subject, identity, scene, or show---and is the unit kept disjoint between training and evaluation. Benchmark quotas are sampled only from held-out groups; every unsampled question from those groups is also excluded from supervised fine-tuning (SFT). Question sampling controls task balance, while per-source validity rules define which observations can become questions: JRDB keeps only static-camera spans, and \textit{Trajectory} filters unsuitable straight or ambiguous tracklets. Full split levels and audits are in Appendix~\ref{app:split}; construction filters are in Appendix~\ref{app:taskdetails}.

Finally, manual review is a quality gate rather than a labeling step: it flags rendering artefacts, mis-aligned metadata or ill-posed prompts for regeneration, but does not directly edit benchmark answers. Figure~\ref{fig:statistic} shows the training-corpus task distribution. The long tail reflects acquisition constraints rather than a design preference: tracked-scene metadata supports dense distance and motion QA, whereas calibrated contact deformation and transient-event recordings are inherently scarce. The benchmark therefore enforces an equal evaluation quota per sub-task, preventing high-resource training categories from masking low-resource physical capabilities.

\section{Experiments}
The experiments answer a progression of questions. First, what fails when current MLLMs perceive people as physical entities? Second, how can this capability be learned: from data mixtures (\S\ref{sec:mixture}), from optimization through deterministic physical rewards (\S\ref{sec:metric_rl}), or from inference-time geometric tools (\S\ref{sec:training_free})? The same physical scorers connect all three, imposing a consistent quantitative criterion across benchmarking and post-training.

\subsection{Setup}

\begin{figure*}[t]
\centering
\includegraphics[width=0.994\linewidth]{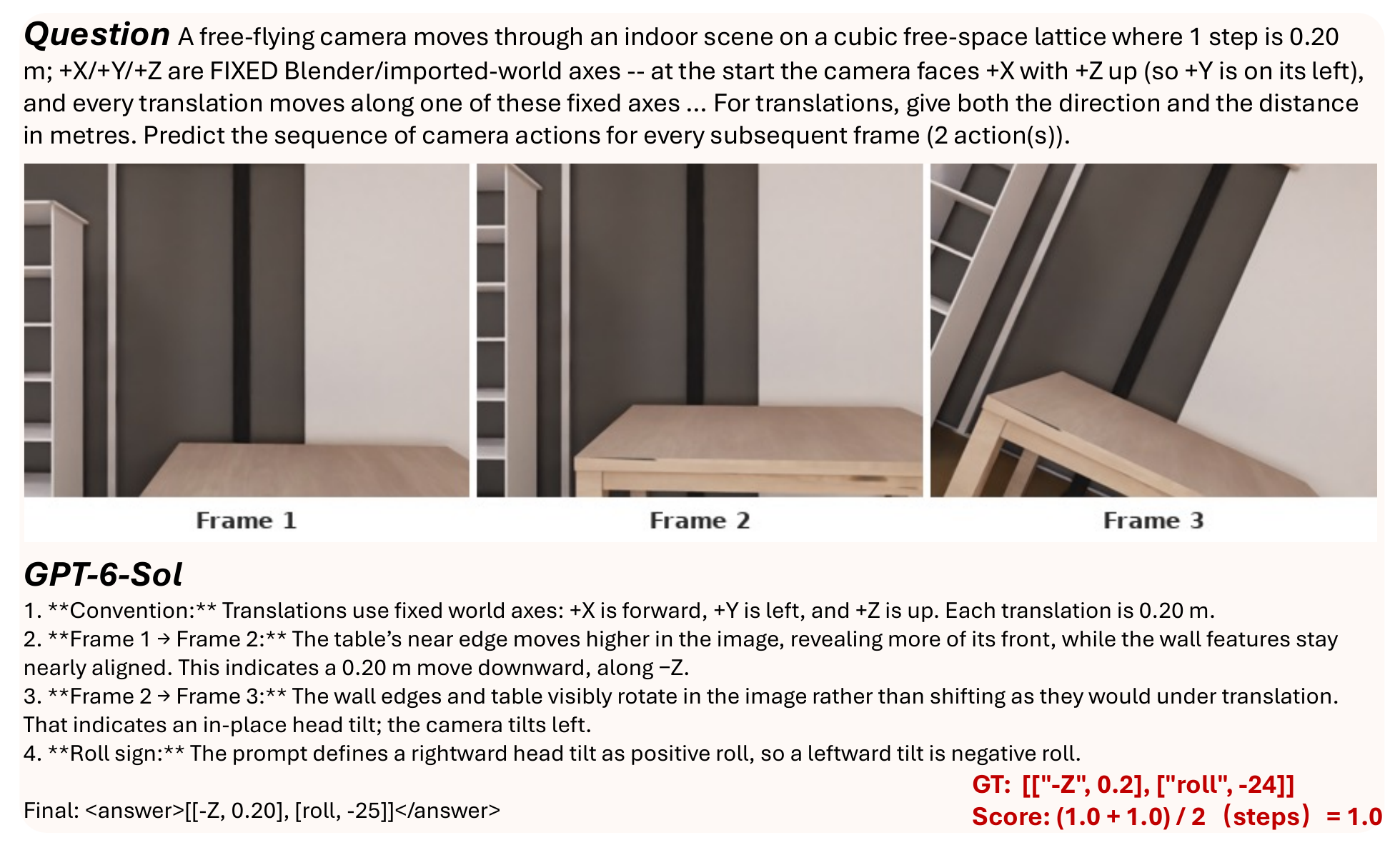}
\caption{\textbf{One scored \textit{CP-Room} QA.} The two-step prediction is \texttt{[[-Z, 0.20], [roll, -25]]} against gold \texttt{[[``-Z'', 0.2], [``roll'', -24]]}. Step~1 matches the gold translation axis ($-Z$) with an exact magnitude ($0.20$ vs.\ $0.2$\,m) for $1.0$; Step~2's roll has circular angular error $1^{\circ}$ ($-25^{\circ}$ vs.\ $-24^{\circ}$), below all five $A_\angle$ thresholds ($5^{\circ}$--$45^{\circ}$), for $1.0$; the sequence score averages over gold steps: $(1.0+1.0)/2=1.0$.}
\label{fig:demo}
\end{figure*}
\textbf{Training setup.} We use Qwen3.5-4B as the base model and fine-tune it for one epoch with the vision transformer (ViT) and aligner frozen, using a cosine scheduler, 5\% warm-up, a global batch size of 1024, and a learning rate of $1\times10^{-5}$ on 256 Ascend HPU-910C.

\textbf{Inference setup.} We implement all inference on top of
lmms-eval, extending it with prompt templates and answer parsers for each
sub-task. Every model uses a generation budget of up to 1,024 new tokens and default sampling; open-source checkpoints
(Qwen3.5-4B~\citep{qwen35}, Qwen2.5-VL~\citep{qwen25vl}, SpaceR~\citep{spacer},
Spatial-SSRL~\citep{spatialssrl}, SpatialEvo~\citep{spatialevo} and
ViLaSR~\citep{vilasr}) are served locally and the proprietary model (GPT-6-Sol) is queried through its official API.
Five further open-source models (Kimi-K3, Qwen3.8-Flash-Next,
Ling-3.0-Flash-VL, MiniMax-M3 and GLM-5.3-Flash)
are queried through hosted API endpoints under the same prompts and scoring,
forming the \textbf{API} group of Table~\ref{tab:main_results}.

\subsection{Evaluation Metrics}
\label{sec:protocol}

For \textbf{Multiple-choice answers} (\textit{Deform}), we assign a score of 1 to an exact option match and 0 otherwise.
\textbf{Numeric answers} are scored by MRA over a threshold set, as in VSI-Bench~\citep{vsibench}:
\[
\begin{aligned}
  C&=\{0.50,0.55,\dots,0.95\},\\
  \mathrm{MRA}&=\frac{1}{|C|}\sum_{\theta\in C}\mathbbm{1}\!\left[\,\frac{|\hat y-y|}{|y|}<1-\theta\,\right],
\end{aligned}
\]
This grants partial credit for continuous quantities while assigning 0 to missing or unparseable predictions.
\textbf{Angle-valued quantities} use absolute circular error in degrees rather than the relative-error denominator of MRA. For predicted and gold angles $\hat\alpha$ and $\alpha$, let $\Delta\alpha=\hat\alpha-\alpha$ and define
\[
\begin{aligned}
  d_\angle(\hat\alpha,\alpha)&=
  \min\!\left(|\Delta\alpha|,\,360^{\circ}-|\Delta\alpha|\right),\\
  A_\angle(\hat\alpha,\alpha)&=\frac{1}{5}
  \sum_{\tau\in\{5,15,25,35,45\}}\mathbbm{1}\!\left[d_\angle<\tau^{\circ}\right].
\end{aligned}
\]
The shortest circular distance makes $1^{\circ}$ and $359^{\circ}$ differ by $2^{\circ}$, and strict threshold comparisons follow the implementation. Because the error is already measured in degrees, it is not normalized by $|\alpha|$ or $360^{\circ}$; the only denominator is the number of thresholds, five. \textit{Gaze} and camera-pose rotation magnitudes use $A_\angle$, while \textit{Trajectory} compares circular turn differences. The detailed step-level scorers for \textit{Trajectory} and the light--shadow coordinates are given in Appendix~\ref{app:scoring_details}.

For \textbf{Camera-pose actions}, we assign 0 when the predicted axis is wrong; otherwise, rotation magnitudes use the angular score above, while translations retain their translation rule (CP-Room translations ignore magnitude once the direction is right), averaged over the gold step count; the per-axis hit rates this gate produces are analysed in \S\ref{sec:diagnosis}. Figure~\ref{fig:demo} shows one scored \textit{CP-Room} QA.

Taken together, the six scorers (multiple-choice match, MRA, the angular
score, and the three sequence rules of Appendix~\ref{app:scoring_details}) are
deterministic, bounded in $[0,1]$, and require no human judgment at scoring time.
The same scoring functions also serve directly as the reward of the RLVR run in
\S\ref{sec:metric_rl}~\citep{deepseekr1, sftmemrl, omnimath, bigmath, openmathinstruct, mmeureka, visualrft, r1onevision, videochatr1, wanr1, sslr1}.

%

\begin{table*}[!t]
\centering
\scriptsize
\setlength{\tabcolsep}{2.2pt}
\caption{\textbf{Data-mixture ablation.} Scores (best per column in bold)
of the Qwen3.5-4B base model and five SFT
checkpoints on HumanPhys, seven spatial-intelligence benchmarks
and two general multimodal benchmarks. The Data columns mark which training
corpora each run mixes (HP = HumanPhys corpus, SpR = SpaceR-151K, SpC =
SpatialCorpus-331K); SFT~\ding{175} is the final model, and SFT~\ding{176}
is the public-space control (SpR + SpC without HP).}
\label{tab:ablation}
\begin{tabular*}{\linewidth}{@{}l@{\extracolsep{\fill}}ccccccccccccc@{}}
\toprule
 & \multicolumn{3}{c}{Data} & HumanPhys & \multicolumn{7}{c}{Spatial intelligence} & \multicolumn{2}{c}{General} \\
\cmidrule(lr){2-4}\cmidrule(lr){5-5}\cmidrule(lr){6-12}\cmidrule(lr){13-14}
Model & HP & SpR & SpC & Overall & 3DSRB. & BLINK & ViewSp. & SPAR & MMSI & VSI & WMA & MMStar & MMB. \\
\midrule
Qwen3.5-4B & -- & -- & -- & 26.68 & \textbf{62.09} & 54.30 & 43.63 & 38.69 & 26.30 & 53.66 & 42.90 & 58.23 & 78.69  \\
SFT \ding{172} & \ding{51} & -- & -- & 74.14 & 50.01 & 55.96 & 36.87 & 36.25 & 29.50 & 48.21 & 40.13 & 52.17 & 81.87  \\
SFT \ding{173} & \ding{51} & \ding{51} & -- & \textbf{75.41} & 58.60 & 60.72 & 44.47 & 40.70 & 31.30 & \textbf{57.07} & 43.98 & 63.44 & 85.40  \\
SFT \ding{174} & \ding{51} & -- & \ding{51} & 73.62 & 58.89 & 59.29 & \textbf{47.18} & 45.34 & 30.80 & 40.05 & 41.54 & 61.54 & 83.76  \\
SFT \ding{175} & \ding{51} & \ding{51} & \ding{51} & 74.21 & 59.43 & 61.56 & 45.59 & 44.94 & \textbf{32.00} & 56.55 & 42.92 & \textbf{63.52} & \textbf{86.25}  \\
SFT \ding{176} & -- & \ding{51} & \ding{51} & 29.74 & 59.90 & \textbf{63.31} & 45.94 & \textbf{45.89} & 29.10 & 56.42 & \textbf{46.76} & 62.45 & 85.48  \\
\bottomrule
\end{tabular*}
\end{table*}

\subsection{Zero-Shot Findings}
\label{sec:findings}

\paragraph{Object-centric spatial training transfers only weakly.} The four checkpoints released for object- and scene-centric spatial reasoning reach 19.76--25.99 overall, 24.32--30.55 points below GPT-6-Sol. Human-physical perception is therefore at best weakly induced by existing spatial corpora, motivating the HumanPhys training pool and our use of SpaceR-151K and SpatialCorpus-331K as complements in \S\ref{sec:mixture}.

\paragraph{The six dimensions are only partially dissociable.} Rankings reverse sharply across measurement targets: SpatialEvo and ViLaSR exceed GPT-6-Sol on trajectory (67.16 and 66.51 vs.\ 58.75), yet SpatialEvo collapses to 3.90 on gaze and 4.10 on blink events, and ViLaSR reaches only 4.60 on gaze. Conversely, GPT-6-Sol leads gaze and blink, while SpaceR and Spatial-SSRL lead the spatially fine-tuned models on gaze and \textit{CP-Room}, respectively. An aggregate score would hide these complementary failures; the per-sub-task columns of Table~\ref{tab:main_results}, of which Figure~\ref{fig:radar} (left) plots the closed-source, base-model and hosted-API eight, attribute each failure to a concrete physical measurement target. The difficulty is not confined to MLLMs: the human reference panel takes a median of 91.8\,s per \textit{CP-Human} question yet reaches only 39.29, highlighting the difficulty of mental 3D lattice transformation and motivating the tool-assisted setting in \S\ref{sec:training_free}.

\paragraph{Hosted-API endpoints interpolate between the base models and the frontier, yet keep idiosyncratic profiles.} The five API-served models reach 25.55--38.38 overall---11.93--24.76 points below GPT-6-Sol, and four of them above both base models (26.68 and 18.90)---while their radar profiles in Figure~\ref{fig:radar} (left) stay as uneven as those of the fine-tunes: Kimi-K3 leads every evaluated MLLM on \textit{Disp} (26.70) and Qwen3.8-Flash-Next on \textit{Depth} (41.30), while GLM-5.3-Flash pairs the strongest blink event score among the MLLM rows outside GPT-6-Sol (75.35) with the two weakest illumination scores in the table (3.56 on \textit{LS-Head}, 2.81 on \textit{LS-Body}). Their radar profiles are thus no more even than those of the locally served fine-tunes; we do not evaluate the same checkpoint behind both routes, so this describes the profiles we measure rather than an effect of hosting itself.

\subsection{Preserving General Spatial Abilities}
\label{sec:mixture}

Supervised training on the HumanPhys corpus introduces a specialization trade-off: SFT~\ding{172} (HumanPhys only) raises Qwen3.5-4B from 26.68 to 74.14 on HumanPhys, but transfer to some object- and scene-centric benchmarks remains uneven—3DSRBench drops from 62.09 to 50.01 and ViewSpatial from 43.63 to 36.87, while MMSI-Bench improves from 26.30 to 29.50 (Table~\ref{tab:ablation}); the sub-task-level transfer gains behind these averages are itemised in Appendix~\ref{app:transfer}. To preserve broader spatial abilities we mix SpaceR-151K~\citep{spacer} and SpatialCorpus-331K~\citep{spatialscore} into the pool. The SpaceR mixture, SFT~\ding{173}, is the strongest HumanPhys checkpoint in this setting (75.41, 1.27 above SFT~\ding{172}) and recovers ViewSpatial to 44.47, while the full three-corpus mixture, SFT~\ding{175}, keeps most of the HumanPhys gain (74.21) and improves MMSI-Bench to 32.00, MMStar to 63.52 and MMBench to 86.25. HumanPhys supervision is therefore effective for the target capability, while complementary spatial corpora are needed to avoid narrow specialisation. Appendix~\ref{app:sft_subtasks} decomposes the same five SFT runs into all 13 HumanPhys sub-tasks.
\begin{figure}[t]
\centering
\includegraphics[width=\linewidth]{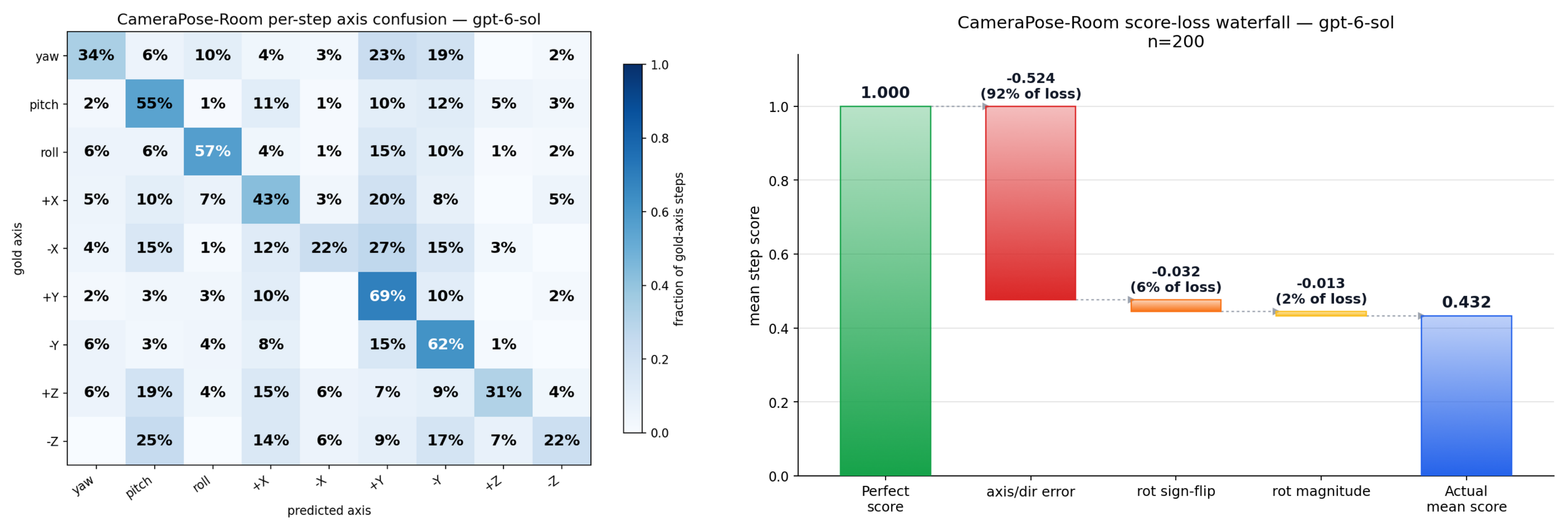}
\caption{\textbf{Diagnosis of \textit{CP-Room} errors (GPT-6-Sol, $n=200$).} Left: per-step gold-vs-predicted axis confusion. Right: score-loss waterfall from a perfect mean per-step score of $1.0$ to the actual $0.432$.}
\label{fig:diagnosis}
\end{figure}

\subsection{Diagnosis}
\label{sec:diagnosis}

Figure~\ref{fig:diagnosis} analyses the failure modes of GPT-6-Sol on \textit{CP-Room} ($n=200$). Rotational-axis recovery varies sharply across axes (per-axis diagonal hit rates 57\%, 55\%, 34\% for roll, pitch, yaw, i.e.\ the fraction of gold-axis steps predicted correctly); translational motions are also frequently confused with rotations or other directions (e.g.\ gold $-Z$ steps predicted as pitch in 25\% of cases). The score breakdown attributes most of the error to choosing the wrong axis or motion type (92.2\% of the total loss), followed by sign errors (5.6\%) and magnitude errors (2.2\%): identifying the correct axis and motion type remains the main challenge. The paired conditions also show that a human anchor is not uniformly helpful: GPT-6-Sol drops from 43.20 on \textit{CP-Room} to 42.75 on \textit{CP-Human}, and Qwen3.5-4B from 8.43 to 4.64, whereas SpaceR rises from 6.69 to 9.87 and ViLaSR is nearly unchanged; and since the two columns also grade translation differently (Appendix~\ref{app:scoring_details}), the \textit{CP-Room}$>$\textit{CP-Human} ordering is not a pure human-anchor effect. Thus, axis and motion-type confusion persists regardless of human presence rather than reducing to a single person-in-frame effect.

\subsection{Metric-Guided RL}
\label{sec:metric_rl}

\begin{figure}[!t]
\centering
\begin{minipage}[t]{0.48\linewidth}
  \centering
  \includegraphics[width=\linewidth]{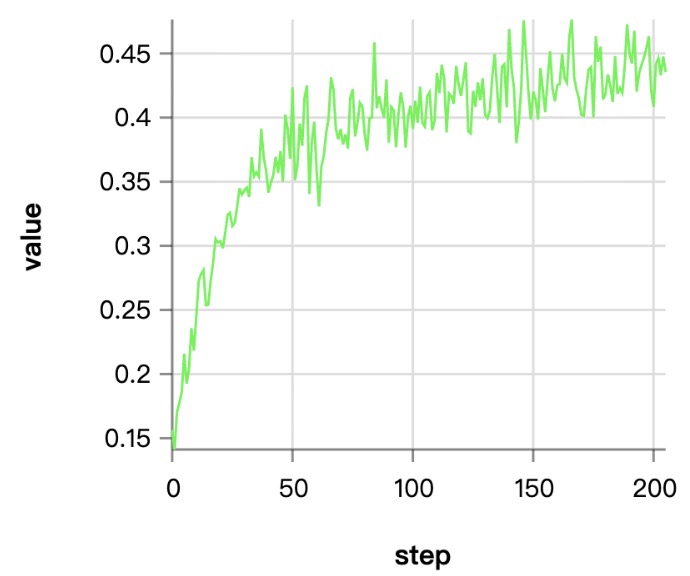}
\end{minipage}\hfill
\begin{minipage}[t]{0.52\linewidth}
  \centering
  \includegraphics[width=\linewidth]{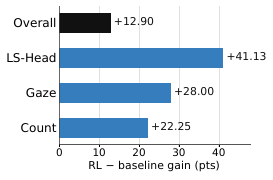}
\end{minipage}
\caption{\textbf{Metric-guided RL on HumanPhys.} \textbf{Left}: reward during GRPO/RLVR training, where the deterministic HumanPhys scorer is used as the reward. \textbf{Right}: overall and top-three sub-task gains of the RL checkpoint relative to the Qwen3.5-4B baseline.}
\label{fig:rl_reward}
\end{figure}

Beyond supervised post-training, we ask whether the benchmark's scoring functions can directly guide policy optimisation.
Starting from Qwen3.5-4B, we run RLVR with Group Relative Policy Optimization (GRPO)~\citep{grpo} on HumanPhys for around 200 steps on 128 HPUs, using the same deterministic metric-based HumanPhys scorer as the reward signal rather than manual preference labels (full configuration, reward design and per-sub-task results in Appendix~\ref{app:rl_details}).
The reward rises consistently during training (Fig.~\ref{fig:rl_reward}), and the resulting checkpoint improves the overall composite from 26.68 to 39.58 (+12.90) over the Qwen3.5-4B baseline.
Thus, the metric is not only an evaluator but also an effective learning signal for physical reasoning.
This makes HumanPhys useful both for measuring human-physical perception and for directly optimising models without human-annotated preferences.

\subsection{Tool-Enhanced Reasoning}
\label{sec:training_free}

To test external computation without parameter updates, we let GPT-6-Sol invoke a calculator, a pinhole camera model, and VGGT~\citep{vggt}. Pinhole is calibrated sensor-assisted (it uses official JRDB intrinsics and distortion coefficients), VGGT is uncalibrated, and No Tool has neither; Table~\ref{tab:training_free_reward} therefore compares information regimes rather than equal-information monocular inference. On 1,050 paired questions (150 per tool-supported sub-task), tools raise the mean reward from 41.2 to 56.6 (+15.4), with the largest gain on \textit{CP-Room} (+32.9) and positive gains on all seven sub-tasks. Appendix~\ref{app:tool_stats} gives the subset construction, explains why the \emph{No Tool} column is not comparable to Table~\ref{tab:main_results}, and audits all 1{,}069 tool calls. Finally, Appendix~\ref{app:agentic_rl} documents an \emph{agentic} variant of this tool-use policy: the teacher trajectories are distilled into a 4B policy via an SFT cold start and a tool-aware RLVR stage, yielding a small model that invokes the same tools itself.

\begin{table}[!tb]
\centering
\small
\setlength{\tabcolsep}{6pt}
\renewcommand{\arraystretch}{1.08}
\caption{\textbf{Training-free tool use.} Mean official HumanPhys reward for GPT-6-Sol on 150 paired questions per sub-task, with and without external tools, scaled to 0--100. Pinhole uses the official JRDB intrinsics and distortion coefficients; VGGT is an uncalibrated learned 3D-geometry tool; No Tool receives neither. $\Delta$ is Tool minus No Tool; all 1,050 questions are included in both conditions.}
\label{tab:training_free_reward}
\begin{tabular}{lcccc}
\toprule
Sub-task & No Tool & Tool & $\Delta$ & $N$ \\
\midrule
Depth            & 53.4 & 61.9 & +8.5 & 150 \\
Distance         & 41.7 & 47.1 & +5.4 & 150 \\
Displacement     & 27.8 & 43.1 & +15.3 & 150 \\
Velocity         & 31.0 & 45.1 & +14.1 & 150 \\
Trajectory       & 61.2 & 66.9 & +5.7 & 150 \\
CP-Room          & 42.3 & 75.2 & +32.9 & 150 \\
CP-Human         & 31.2 & 57.1 & +25.9 & 150 \\
\midrule
\textbf{Overall} & \textbf{41.2} & \textbf{56.6} & \textbf{+15.4} & \textbf{1,050} \\
\bottomrule
\end{tabular}
\end{table}

\section{Conclusion}
\label{sec:conclusion}

HumanPhys is a 13-task human-centric benchmark and training resource for deterministic human-physical perception evaluation and optimization. Its programmatically derived labels and bounded scorers make evaluation reproducible and support utility validation through SFT, metric-guided RL, and tool-assisted reasoning. Across these settings, current models still struggle with human-physical geometry, while targeted HumanPhys supervision and complementary geometric tools provide measurable improvements.

\section*{Ethics statement}

HumanPhys is built entirely by secondary use of publicly released research
datasets (JRDB, DanceTrack, UCSD Ped, VideoAttentionTarget, EyeBlink8, DECAF)
and of synthetic assets (FLAME/MANO/SMPL-X parametric bodies, BEDLAM
identities, ReplicaCAD scenes); we collect no new human-subjects data. The
human-reference panel consisted of ten adult volunteers affiliated with our
institution, who participated voluntarily and without compensation, and produced
the 647 valid answers behind the \emph{Human} row of
Table~\ref{tab:main_results}; only anonymised per-question answers and response
times were recorded, and no identity-linked or sensitive attribute was collected,
stored or released. We inherit and enforce the upstream licence terms: the
release comprises only the trimmed video segments each question consumes,
together with derived question--answer pairs, calibration metadata and
renderer-generated sequences, all under non-commercial research terms that
preserve upstream attribution (JRDB clips CC BY-NC-SA 3.0, DanceTrack
annotations CC BY 4.0). For sources whose terms do not clearly permit re-hosting
we ship fetch scripts pointing to the upstream archives instead, plus a
per-segment provenance manifest and a takedown procedure that honours any
upstream, creator or subject request. All rendered Group-II material inherits the
MPI non-commercial research licence of its parametric bodies and scenes and ships
with regeneration scripts, so that asset bytes never change hands. Our labels
describe geometry and transient physiological cues (depth, distance, image-plane
gaze angle, blink count) rather than identity or sensitive attributes: the
release contains no identity labels and we make no attempt to re-identify any
individual. We note the dual-use potential of fine-grained physical reading of
people and mitigate it by non-commercial licensing, by withholding
identity-linked attributes, and by restricting every task to short,
context-free clips without narrative or behavioural profiling.

\section*{Reproducibility statement}

Evaluation runs on the open-source lmms-eval framework with prompt templates for
each sub-task and the deterministic, judgement-free scorers of \S\ref{sec:protocol};
generation is capped at 1{,}024 tokens with otherwise default sampling. As in
\S\ref{sec:tasks}, every question carries its ordered sampled frames
(\emph{frame}: one image; \emph{video}: the clip's full sampled frame list), so
no frame-rate resampling and no per-video frame cap is ever applied to a
HumanPhys question, identically in training and evaluation; the only videos in our
pipeline belong to the external SpaceR-151K training mixture, where each video
is uniformly sub-sampled under a per-video cap of 16 frames and a shorter clip
may yield fewer frames. The SFT checkpoints update all parameters of Qwen3.5-4B except the ViT and aligner, which are frozen, with learning rate $1\times10^{-5}$, a cosine schedule with
5\% warm-up and global batch size 1024 for one epoch on 256 Ascend HPU-910C
accelerators, mixing the indicated corpora in full (\S\ref{sec:mixture}).
Construction is fully specified in \S\ref{sec:benchmark}: source datasets,
filtering thresholds, walk grammars, scorer definitions and the stratified
evaluation split. Code, prompt templates, scorers and the evaluation split will be
released upon acceptance to allow exact replication.

\section*{AI use statement}

In this work, we used generative AI tools only for tasks with recommended
disclosure: editing portions of the manuscript for readability and scripting
auxiliary data-processing code. Every AI-assisted passage and script was
reviewed, verified and approved by the authors. We have not used generative AI
tools for any task with required disclosure: the HumanPhys corpus is generated
programmatically from source-dataset metadata and simulator/renderer state rather than by an MLLM,
and no mathematical claims, hypotheses, experimental designs or result
interpretations were produced by generative models. The large vision--language
models evaluated in our experiments (e.g.\ GPT-6-Sol and Qwen3.5-4B)
serve solely as study subjects, not as assistants in the research
workflow. We take responsibility for the final content of this work, including
text, claims or artifacts produced with the aid of generative AI.

{
    \small
    \bibliographystyle{ieeenat_fullname}
    \bibliography{main}
}
\clearpage
\appendix

\section{Construction Details for Each Sub-task}
\label{app:taskdetails}

This appendix provides the full construction details summarized in \S\ref{sec:tasks}: the filtering
thresholds, coordinate conventions, walk grammars and worked examples behind each
of the thirteen sub-tasks.

\paragraph{Group I: real-world measured human-centered physical cues.}
\begin{itemize}[leftmargin=1.2em,itemsep=1pt]
  \item \textbf{Spatial state} (\textit{Depth}, \textit{Distance}, \textit{Count}, \textit{Gaze}).
  \textit{Depth} and \textit{Distance} are derived from JRDB~\citep{jrdbact}, whose five cameras ride a moving
  robot: point-cloud registration between consecutive samples flags every range
  whose per-step camera displacement exceeds 0.15\,m in a single step; questions are
  constructed only from frames outside those ranges (a candidate clip is kept only if all of its frames are static). \textit{Count} is read from the boxes of
  DanceTrack~\citep{dancetrack} and UCSD~\citep{ucsd}, and \textit{Gaze} from the head
  boxes and fixation points of VideoAttentionTarget~\citep{videoattentiontarget},
  as listed in Table~\ref{tab:composition}.
  \textit{Depth} is read directly from the calibration metadata, namely the
  distance to the reference sensor. \textit{Distance} is the pairwise
  inter-person distance measured at
  the \emph{foot} point $(c_x,c_y,c_z-\tfrac{h}{2})$ which reduces the effect of variation in human height.
  \textit{Count} is the number of humans satisfying an explicit spatial
  criterion (everybody dancing in frame, or inside a labelled walkway ROI),
  explicitly stated in the question. \textit{Gaze} is the clockwise angle
  $\operatorname{atan2}(\Delta y,\Delta x)+90^{\circ}$ between the upward
  image direction and the head-centre-to-fixation-point vector; because this
  angle lives in the image plane it is camera-dependent by construction, we
  treat it as a viewpoint-conditioned physical cue, not an invariant head
  direction.
  \emph{Ex:} \textit{``What is the minimum distance between pedestrian 29 and
  pedestrian 46 in the video?''}
  \item \textbf{Kinematics} (\textit{Displacement}, \textit{Velocity}, \textit{Trajectory}). \textit{Displacement} is the cumulative ground-plane path length
  $\sum_i\sqrt{\Delta x_i^2+\Delta y_i^2}$ between the first and last frame of
  the clip; \textit{Velocity} divides that path length by the clip's elapsed
  time, with the sampling interval stated in the question so the model must bind
  visual motion to a stated time base; and \textit{Trajectory} is the sequence
  of heading changes and step lengths as tuples
  $(\text{turn\_deg},\text{distance\_m})$ with the first step anchored at
  $0^{\circ}$ and positive turn meaning left. Curved paths are kept by a
  lightweight classifier that triages every candidate trajectory as straight,
  curved or unsuitable, so that low-quality tracks never become questions
  (Appendix~\ref{app:taskdetails}); its residual error is therefore a component
  of the \textit{Trajectory} gold noise.
  \item \textbf{Event} (\textit{Blink}). The number of discrete behaviour
  events (eye blinks) in a clip, requiring the model to integrate evidence
  across frames instead of reading a single canonical pose. Its source,
  EyeBlink8~\citep{eyeblink8}, runs at 30\,Hz and ships its own human blink annotations
  (408 blinks over 70{,}992 frames at 640$\times$480), which we use as published rather than re-labelling;
  consequently a raw 32-frame window would almost never contain more than one blink, so
  we down-sample, compressing each open-eye run to at most two evenly
  spaced frames and collapsing each closed-eye run to its middle frame.
  \emph{Ex:}
  \textit{``How many times did the person blink in this video?''}
\end{itemize}

\begin{table*}[!t]
\centering
\small
\setlength{\tabcolsep}{3pt}
\caption{\textbf{Human-reference study per sub-task.} Random and Text Only are full-bank baselines from Table~\ref{tab:main_results}. \emph{Responses} gives the number of valid panel answers and \emph{Median time (s)} the median response time over those same valid answers, with cell shading proportional to time (darker = slower). The \emph{Human} score is computed with the same deterministic scorer and strict type-parsing rule as the MLLM rows, using each item's own move-rule normaliser (Appendix~\ref{app:scoring_details}).}
\label{tab:human_time}
\begin{tabular*}{\linewidth}{@{}l@{\extracolsep{\fill}}ccccccccccccc@{}}
\toprule
 & Depth & Dist & Count & Gaze & Disp & Vel & Traj & Blink & Deform & LS-Head & LS-Body & CP-Room & CP-Human \\
\midrule
Random & 17.13 & 9.69 & 22.15 & 13.62 & 4.31 & 5.06 & 24.27 & 18.70 & 24.28 & 27.08 & 53.47 & 8.80 & 5.29 \\
Text Only & 10.35 & 1.70 & 0.00 & 0.00 & 0.00 & 0.00 & 18.20 & 0.00 & 23.00 & 20.86 & 52.35 & 8.61 & 3.73 \\
Responses & 47 & 56 & 57 & 60 & 52 & 51 & 52 & 52 & 25 & 35 & 46 & 57 & 57 \\
\emph{Median time (s)} & \cellcolor{black!7}10.8 & \cellcolor{black!8}12.1 & \cellcolor{black!8}12.9 & \cellcolor{black!8}12.7 & \cellcolor{black!8}11.8 & \cellcolor{black!8}12.7 & \cellcolor{black!13}37.6 & \cellcolor{black!8}12.8 & \cellcolor{black!8}12.1 & \cellcolor{black!50}\color{white}204.6 & \cellcolor{black!46}\color{white}186.9 & \cellcolor{black!25}92.1 & \cellcolor{black!25}91.8 \\
Human & 46.38 & 33.75 & 79.47 & 58.00 & 37.31 & 32.35 & 60.42 & 77.88 & 36.00 & 52.97 & 70.07 & 50.46 & 39.29 \\
\bottomrule
\end{tabular*}
\end{table*}

\paragraph{Triage filter and review statistics.}
Low-quality tracks are removed before any \textit{Trajectory} item is generated:
a lightweight classifier over ground-plane path descriptors sorts every
candidate trajectory as straight, curved or unsuitable, and only the first two
classes become questions. The filter is deliberately recall-oriented on curved
paths, so that genuinely curved walks survive it; its residual error is the
\textit{Trajectory} gold-noise component disclosed in \S\ref{sec:benchmark}. Of
1{,}435 clean JRDB candidate segments it retains 520 usable curved paths, the
pool from which all \textit{Trajectory} items are drawn. Sampled renders are visually checked per engine before acceptance (e.g.\
face framing of placed humans in \textit{CP-Human}); items flagged as ill-posed
are regenerated, never hand-edited (\S\ref{sec:construction}).

\paragraph{Group II: controlled physically rendered probes.}
\begin{itemize}[leftmargin=1.2em,itemsep=1pt]
  \item \textbf{Deformation} (\textit{Deform}). A FLAME~\citep{flame} head deformed by MANO~\citep{mano}
  hand contact, captured by DECAF~\citep{decaf}, is rendered per frame as a $2\times4$ panel grid (two head-following views
  $\times$ four deformation strengths), and each item is an 8-frame clip of such grids. Each option's strength $\alpha$ rescales
  the measured contact deformation by a Gaussian influence field
  $w_i=\exp(-d_i^2/2\sigma^2)$,
  $\mathbf{V}'_i=\mathbf{V}^{\text{und}}_i+(\alpha-1)w_i\mathbf{D}_i$; exactly
  one option ($\alpha=1.0$) matches physical reality, while the three
  distractor options sample their strength from
  $\alpha\in[0,0.5]\cup[2,3]$ under a deterministic per-(subject, window)
  seed, so that only the $\alpha{=}1$ option is consistent with the measured
  contact, while every distractor is clearly under- or over-deformed. DECAF's monocular fit is therefore the anchor of the $\alpha{=}1$
  option, so its fit error is a component of the \textit{Deform} gold noise,
  exactly as the triage classifier's error is for \textit{Trajectory}.
  \emph{Ex:}
  \textit{``Which face's deformation best matches physical reality?''}
  \item \textbf{Illumination} (\textit{LS-Head}, \textit{LS-Body}). A standing
  human (200 BEDLAM~\citep{bedlam} identities) inside a light stage of 27 point lights on a
  $3\times3\times3$ lattice; exactly one light is on per frame and its
  coordinate follows a lattice walk of 3--6 frames that the model must infer
  from the shadow motion on the face and body. The two variants probe complementary cue regimes and walk grammars. \textit{LS-Head} uses the frontal
  self-portrait camera, where the shading change on the face is subtle, and
  restricts each walk step to a single lattice axis ($\pm X$, $\pm Y$ or $\pm Z$
  by one step), keeping the facial cue readable; \textit{LS-Body} uses the 8
  peripheral wall cameras, where body shading and cast shadows are far more
  informative, and runs the \emph{fast} walk, whose steps may change up to three
  lattice axes at once (step lengths $1$, $\sqrt2$ or $\sqrt3$), so its legal step
  set is larger and its per-axis shading signals are mixed. Because cue strength
  and step grammar change together, we treat the two variants as separate
  sub-tasks throughout; the quantitative form of this caution is in
  Appendix~\ref{app:scoring_details}.
  \item \textbf{Camera pose} (\textit{CP-Room}, \textit{CP-Human}). A
  free-flying camera performs a random walk on a collision-checked cubic lattice (0.20\,m step; lattice
  nodes that intersect scene geometry are discarded beforehand) through 84
  baked indoor stages; every frame
  transition is exactly one \emph{atomic action}: a translation along
  $\pm X/\pm Y/\pm Z$ (exactly one 0.20\,m step for \textit{CP-Room}, 1--3 steps for
  \textit{CP-Human}, as in the released prompts), or an in-place yaw/pitch/roll rotation of
  $\pm1^{\circ}$--$\pm30^{\circ}$ (accumulated pitch capped at $+60^{\circ}$).
  Frame~1 shows a 3D axis gizmo plus the stated camera pose, and the question
  asks for the action sequence of all later frames. \textit{CP-Human} plants a
  SMPL-X~\citep{smplx} human and samples paths that keep the face framed.
\end{itemize}

\subsection{Rendering Parameters for the Synthetic Probes}
\label{app:render_params}

Every Group~II frame is ray-traced with headless Blender~3.6~\citep{blender} and the Cycles path
tracer, denoised, with motion blur and depth of field disabled; outputs are PNG
stills.
Table~\ref{tab:render_params} lists the exact per-family configuration so that
the probes can be regenerated.

\paragraph{Light--shadow stages.}
The light-stage files already contain the 27 point lights on the
$3\times3\times3$ lattice and all camera views, so the renderer zeroes
\emph{every} light and then re-enables only the frame's designated light (1000\,W,
Table~\ref{tab:render_params}) before writing the still. Illumination is therefore never ambient and
the active lamp is never visible, which is what makes the lattice walk
recoverable only from shading and cast-shadow motion. \textit{LS-Head} renders
the frontal self-portrait camera of the stage, \textit{LS-Body} one of its
eight peripheral wall cameras; in both cases the intrinsics are those baked
into the stage asset. A ``frame'' here is a lighting state rather than a moment in time.

\paragraph{Camera-pose walks.}
The world-from-camera $4\times4$ matrix of each sampled lattice walk is applied
verbatim, so the rendered view is exactly the pose the question asks about. Illumination is a \emph{fixed} three-point rig plus world ambient
(Table~\ref{tab:render_params}) that does not depend on the walk.

\paragraph{Deformation panels.}
Each option is rendered with a 50\,mm head-following lens at the resolution
of Table~\ref{tab:render_params}; the
$2\times4$ grid shown in the \textit{Deform} example card of Appendix~\ref{app:examples} combines the two
head-following views at $\pm45^{\circ}$ face angle with the four strength
levels $\alpha$. Two choices are specific to reading a contact deformation off
a mesh rather than off a photograph: per-face (flat) shading, which avoids
shading artefacts along the neck seam of the DECAF fit, and a hand rendered at
5\% opacity so that the cheek deformation it causes stays visible.
Four head-following area lights provide diffuse frontal illumination.
Finally, the Gaussian influence
field $w_i=\exp(-d_i^2/2\sigma^2)$ uses $\sigma=8.0$ measured in
\emph{mesh-BFS hop distance} to the nearest contact vertex, not in millimetres,
which is what keeps the influence scale resolution-independent across subjects.

\begin{table*}[!t]
\centering
\scriptsize
\setlength{\tabcolsep}{4pt}
\caption{\textbf{Rendering configuration of the three synthetic sub-task
families} (headless Blender~3.6, Cycles, PNG stills). ``Samples'' are Cycles path-tracing samples per
pixel; the point-light energies of the light stage and the area-light energies
of the camera-pose rig are stated in the last column, and every light not
listed is switched off. \textit{Deform}: each of the eight panels of a frame is rendered on a $600\times800$ canvas and
downscaled 50\,\% to $300\times400$; the shipped frame is the $2\times4$ composite (two head-following-view rows $\times$ four strength columns), i.e.\ $1200\times800$\,px.}
\label{tab:render_params}
\begin{tabular}{@{}p{2.6cm}p{3.9cm}p{3.3cm}p{6.0cm}@{}}
\toprule
Family & Device / camera & Output & Lighting \\
\midrule
\textit{LS-Head}\newline \textit{LS-Body}
  & Cycles, CUDA GPU; selfie camera and 8 wall cameras baked into the light stage
  & $640\times480$\,px, 16 samples, PNG still per lighting state
  & 1 of 27 point lights ($3\times3\times3$ lattice) at $1000$\,W, all others at $0$\,W \\
\textit{CP-Room}\newline \textit{CP-Human}
  & Cycles, CUDA GPU; horizontal FOV $60^{\circ}$, world-from-camera matrix applied verbatim
  & $640\times480$\,px, 16 samples, 15\,fps frame sequence
  & fixed rig: sun key $3.0$; area fill $1500$\,W, area rim $1000$\,W; camera-rigid point $400$\,W ($0.3$ soft-shadow radius); world background RGB $(0.40,0.42,0.45)$ at strength $3.0$ \\
\textit{Deform}
  & Cycles, CPU; $50$\,mm head-following lens, views at $\pm45^{\circ}$ face angle
  & $1200\times800$\,px composite ($2\times4$ panels of $300\times400$; 50\,\% of a $600\times800$ canvas), 32 samples, 8 frames per item
  & 4 head-following area lights: key $10$\,W, fill $20$\,W, rim $30$\,W, bottom $10$\,W \\
\bottomrule
\end{tabular}
\end{table*}

\subsection{Detailed Scoring Functions}
\label{app:scoring_details}

The sequence families are scored by deterministic step-level rules, which we omit from \S\ref{sec:protocol} and state here so that all three tuple families are given symmetrically. This appendix also documents the two vision-free reference baselines of Table~\ref{tab:main_results} and the score floor that the light--shadow metric induces.

\paragraph{Trajectory tuples.} Each gold step is described by relative turn $u_i$ and metric step length $a_i$, and the corresponding prediction is $(\hat u_i,b_i)$. The circular turn difference and vector error are
\[
\begin{gathered}
  \delta_i=d_\angle(\hat u_i,u_i),\qquad
  e_i=\sqrt{a_i^2+b_i^2-2a_ib_i\cos\delta_i},\\
  \bar e=\frac{1}{m}\Big(\sum_{i<\min(n,m)}e_i+\sum_{i=n}^{m-1}a_i\Big),\qquad
  s=\frac{1}{1+\bar e},
\end{gathered}
\]
where $m$ and $n$ are the gold and predicted step counts. Missing steps are
charged the full gold step length, and extra predicted steps are ignored, so
the score matches the relative-step semantics of the task.

\paragraph{Light--shadow coordinates.} For every light-shift sequence we compute
\[
\begin{gathered}
  s=\mathrm{clip}\!\left(1-\frac{\bar e}{2\,C_{\mathrm{mode}}},\,0,\,1\right),\\
  C_{\mathrm{mode}}=
  \begin{cases}
    1,&\textit{LS-Head},\\
    \sqrt{3},&\textit{LS-Body},
  \end{cases}
\end{gathered}
\]
where $\bar e$ is the mean per-step Euclidean coordinate error in lattice
units and $C_{\mathrm{mode}}$ is the maximum legal step length of the walk mode.
The clip keeps the score bounded and assigns zero to any prediction whose
mean step error exceeds $2C_{\mathrm{mode}}$.

\paragraph{Camera-pose action tuples.}
The gold answer is a sequence of $m$ atomic actions and the prediction a
sequence of $n$ actions, each a pair (token, magnitude) whose token is drawn
from the nine-token vocabulary, split into the rotation set
$\mathcal{A}=\{\text{yaw},\text{pitch},\text{roll}\}$ and the translation set
$\mathcal{T}=\{\pm X,\pm Y,\pm Z\}$, and whose magnitude is the translated distance in metres (an integer
multiple of the $0.20$\,m lattice step) for the translation tokens or a number of
degrees for the rotation tokens. Writing the gold sequence as $\{(d_i,a_i)\}_{i=1}^{m}$
and the prediction as $\{(\hat d_i,\hat a_i)\}_{i=1}^{n}$, the sequence score
is the mean over the gold steps,
$s=\frac{1}{m}\sum_{i=1}^{\min(m,n)} s_i$, with the per-step term
\[
  s_i=\begin{cases}
    0, & \hat d_i\neq d_i,\\[2pt]
    A_\angle(\hat a_i,a_i), & d_i\in\mathcal{A},\\[2pt]
    1, & d_i\in\mathcal{T}\ \text{and}\ (\textit{CP-Room}\ \text{or}\ a_i=0),\\[2pt]
    \mathrm{MRA}(\hat a_i,a_i), & d_i\in\mathcal{T},\ \textit{CP-Human},\ a_i\neq0.
  \end{cases}
\]
The branches are read in order, so the last three are reached only when the
predicted token is correct, $\hat d_i=d_i$. A wrong token scores $0$ for that step whatever its magnitude, which is why the per-axis hit rates this gate induces are analysed in \S\ref{sec:diagnosis}. Gold steps
that the prediction omits are charged $0$ but stay in the denominator $m$, and
steps predicted beyond $m$ are ignored, so a truncated answer cannot score
above the fraction of gold steps it covers. Rotation magnitudes use the
angular scorer $A_\angle$ of \S\ref{sec:protocol} in both variants, so the two
camera-pose sub-tasks differ \emph{only} in the translation branch:
\textit{CP-Room} credits a translation step as soon as its direction is right
and therefore measures the recovered shape of the ego-motion path, whereas
\textit{CP-Human} grades the translated distance with the same relative-error
$\mathrm{MRA}$ as the numeric sub-tasks and therefore measures direction and
metric magnitude jointly. The two columns are therefore not on a common scale for
translation steps: part of the \textit{CP-Room}$\to$\textit{CP-Human} drop of
\S\ref{sec:diagnosis} reflects the stricter magnitude requirement, while the
axis-confusion breakdown is unaffected, because the token gate and the rotation
branch are identical in both variants.

\paragraph{Vision-free reference baselines (\textit{Random}, \textit{Text Only}).}
Both baseline rows of Table~\ref{tab:main_results} pass through exactly the
same structural parser and the same scorers as every model row and never
receive a frame-specific privilege; they differ only in where the answer comes
from. \textit{Text Only} runs the Qwen3.5-4B checkpoint on the verbatim
question with all frames withheld, so any credit it earns comes from the
question text alone --- which, for light--shadow, already states the lattice
geometry, the move rule and the first-frame light coordinate. \textit{Random}
is a schema-legal blind guesser rather than an unconstrained one: for each
item it draws an answer from the answer space that the question text and the
training split permit, in the sub-task's native answer schema, and the draw is
repeated under 50 independent seeds whose scores are averaged. Concretely, the
draw is a uniform option letter for \textit{Deform}; a value from the
training-split answer distribution of that sub-task for the scalar quantities
scored by $\mathrm{MRA}$ or $A_\angle$; a uniformly sampled token with a
sampled magnitude per step for camera pose; a sampled (turn, distance) list
for \textit{Trajectory}; and, for light--shadow, a lattice walk that starts
from the first-frame coordinate given in the prompt and draws each step
uniformly among the moves allowed by that question's move rule. Every draw is
schema-legal by construction, so no item scores $0$ for a structural reason: a
zero can only come from the scorer itself (the light--shadow clip, or a relative
error outside the $\mathrm{MRA}$ threshold set).

\paragraph{Why \textit{LS-Body} has an intrinsic $\approx\!53$-point floor.}
The light--shadow score is affine in the mean per-step coordinate error and
clipped at zero, so a variant's blind level is set by the ratio between the error
of a schema-legal but vision-free answer and the normaliser
$2C_{\mathrm{mode}}$; the two variants share one $3\times3\times3$ lattice
while their normalisers differ by a factor of $\sqrt3$, so their scores are not
comparable to each other. For \textit{LS-Body},
$2C_{\mathrm{mode}}=2\sqrt3=3.464$ equals the lattice diameter, so the clip
essentially never engages and the blind legal walk defined above scores
$53.4\pm0.7$ over the 50 seeds, reproducing the $53.47$ of
Table~\ref{tab:main_results}. \textit{LS-Head} differs in kind rather than in
degree: its
single-axis move rule gives $2C_{\mathrm{mode}}=2$, smaller than the lattice
diameter, so the clip zeroes $12.8\%$ of sampled blind sequences and the same
strategy scores only $26.8\pm1.3$ (Table~\ref{tab:main_results}: $27.08$).
Two consequences follow.
(i)~An \textit{LS-Body} value in the forties or low fifties is what a
schema-correct, vision-free answer must obtain, so it is a floor rather than
evidence of light localisation: among the zero-shot rows of
Table~\ref{tab:main_results} only GPT-6-Sol ($68.31$) clears the blind-walk
level of $53.4$, and every other model sits at or below it (the API-served group
spans $46.5$--$53.4$ once GLM-5.3-Flash is set aside; see (ii)). (ii)~A
score \emph{below} the floor carries information of its own: every constant
in-lattice answer averages at least $40.2$ on \textit{LS-Body}, so a value near
zero is consistent only with off-lattice or non-integer coordinates, truncated
coordinate lists or unparseable outputs, each counted as $0$ for that item; the
stored outputs of the hosted endpoints are outside the format audit of
Table~\ref{tab:unparseable} (Appendix~\ref{app:prompts}).

\onecolumn
\raggedbottom
\section{HumanPhys Dataset Examples}
\label{app:examples}

Each sub-task is illustrated by one record of the released corpus, identified
on its card. Evaluation items are emitted by the same pipeline
and are identical in form; they differ only in being drawn from held-out source
groups (Appendix~\ref{app:split}) and in carrying no corpus identifier.

\definecolor{HPBlue}{HTML}{3568C0}
\definecolor{HPPurple}{HTML}{7654B8}
\definecolor{HPOrange}{HTML}{C86B27}
\definecolor{HPRose}{HTML}{AE4F6C}
\definecolor{HPTeal}{HTML}{247F86}
\definecolor{HPCamera}{HTML}{52677D}
\definecolor{HPGreen}{HTML}{147D68}
\definecolor{HPText}{HTML}{17243A}
\definecolor{HPMuted}{HTML}{65738A}
\definecolor{HPPanel}{HTML}{F4F6FA}
\definecolor{HPLine}{HTML}{DCE3EE}

\newcommand{\HPImage}[2]{%
  \tikz[baseline] \node[inner sep=0pt,outer sep=0pt,clip,rounded corners=1.5mm]
    {\includegraphics[#1]{#2}};}

\newcommand{\HPFrame}[4]{%
  \begin{minipage}[t]{#1}
    \vspace{0pt}\centering
    \HPImage{width=\linewidth,height=#2,keepaspectratio}{#3}\\[0.7mm]
    {\sffamily\scriptsize\bfseries\color{HPMuted}#4}
  \end{minipage}}

\newlength{\HPBoxTextWidth}
\newcommand{\HPInfoBox}[6]{%
  \setlength{\HPBoxTextWidth}{#1}\addtolength{\HPBoxTextWidth}{-4mm}%
  \begin{tikzpicture}
    \node[anchor=north west,inner sep=2mm,outer sep=0pt,
      rounded corners=1.5mm,draw=#4,fill=#3,line width=0.55pt,
      text width=\HPBoxTextWidth,minimum height=#2,
      align=flush left,font=\sffamily\color{HPText}]
      {{\scriptsize\bfseries\color{HPMuted}#5}\par
       \vspace{0.8mm}{\small\color{HPText}#6}};
  \end{tikzpicture}}

\newcommand{\HPSampleCard}[4]{%
  \par\vspace{3mm}\noindent
  \begin{tikzpicture}
    \node[anchor=north west,inner sep=0pt,outer sep=0pt] (hpbody)
      {\begin{minipage}[t]{0.998\linewidth}
         \vspace*{11mm}
         \hspace*{4mm}\begin{minipage}{0.96\linewidth}
           \sffamily\color{HPText}#4
         \end{minipage}
         \vspace*{4mm}
       \end{minipage}};
    \begin{pgfonlayer}{background}
      \fill[black!10,rounded corners=2.4mm]
        ([xshift=0.25mm,yshift=-0.70mm]hpbody.north west)
        rectangle ([xshift=0.25mm,yshift=-0.70mm]hpbody.south east);
      \fill[#1!2,rounded corners=2.4mm]
        (hpbody.north west) rectangle (hpbody.south east);
      \begin{scope}
        \clip[rounded corners=2.4mm] (hpbody.north west) rectangle (hpbody.south east);
        \fill[#1] (hpbody.north west) rectangle
          ([yshift=-9mm]hpbody.north east);
        \fill[#1!14] ([yshift=-9mm]hpbody.north west)
          rectangle ([yshift=-9.8mm]hpbody.north east);
      \end{scope}
      \draw[#1!60,rounded corners=2.4mm,line width=0.65pt]
        (hpbody.north west) rectangle (hpbody.south east);
    \end{pgfonlayer}
    \node[anchor=west,text=white,font=\sffamily\bfseries\Large]
      at ([xshift=4mm,yshift=-4.5mm]hpbody.north west) {#2};
    \node[anchor=east,text=white!94,font=\sffamily\scriptsize]
      at ([xshift=-4mm,yshift=-4.5mm]hpbody.north east) {#3};
  \end{tikzpicture}\par}

\newcommand{\HPVisualLabel}[1]{%
  {\sffamily\scriptsize\bfseries\color{HPMuted}#1}\par\vspace{1.1mm}}
\newcommand{\HPAnswer}[2]{%
  {\LARGE\bfseries\color{HPGreen}#1}\quad{\small\color{HPMuted}#2}}
\newcommand{\HPSampleNote}[1]{%
  \par\vspace{0.8mm}{\scriptsize\color{HPMuted}#1}}
\newcommand{\HPSingleNote}[1]{%
  \par\vspace{-1.2mm}{\scriptsize\color{HPMuted}#1}}

\noindent{\color{HPBlue}\sffamily\bfseries\scriptsize 13 SUB-TASKS $\bullet$ REAL SFT SAMPLES}\par
\vspace{1mm}
\noindent One representative SFT example is shown for each of the 13 sub-tasks.
Frames appear in temporal order, and every frame associated with the selected sample
is displayed. Questions retain task-relevant definitions but omit repetitive
output-format boilerplate; ground-truth values are unchanged. The source and SFT
sample ID appear below each card.

\HPSampleCard{HPBlue}{Count}{SPATIAL STATE \quad$\bullet$\quad DANCETRACK \quad$\bullet$\quad 1 FRAME}{%
  \HPVisualLabel{INPUT FRAME}\vspace{-2.5mm}
  \begin{center}
    \HPImage{width=\linewidth,height=49mm,keepaspectratio}{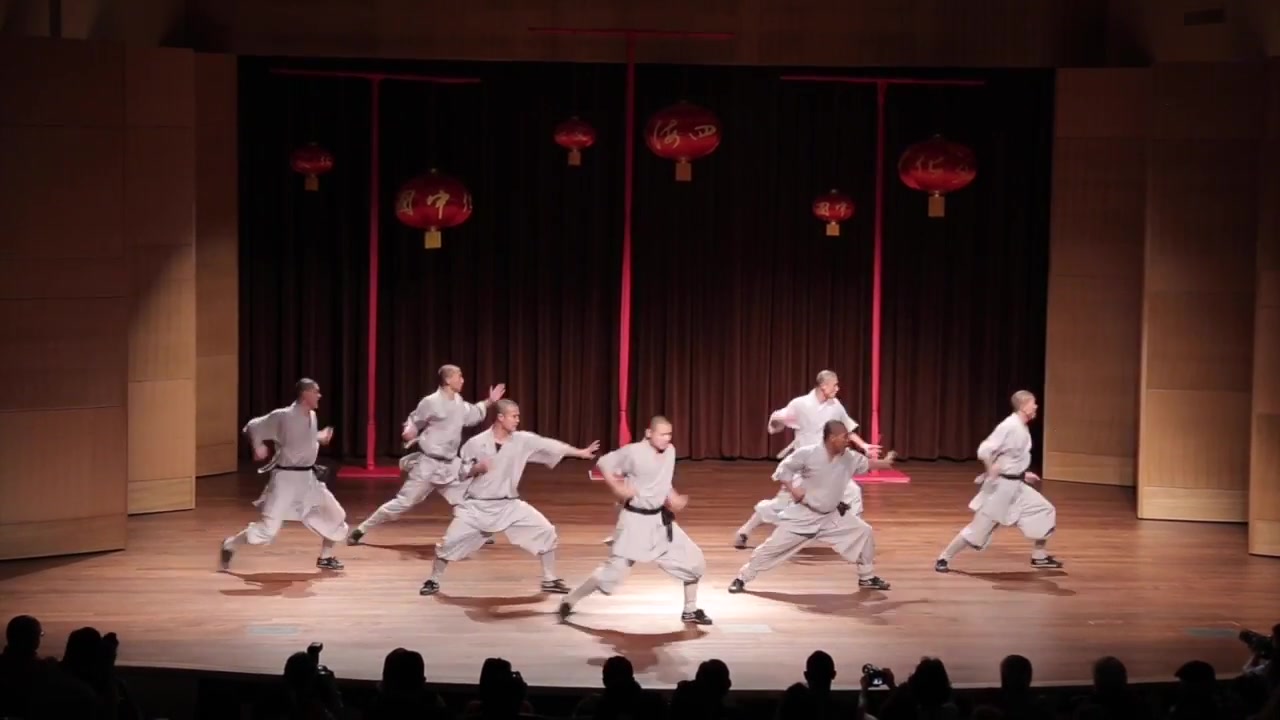}
  \end{center}
  \vspace{-1.5mm}
  \noindent\begin{minipage}[t]{0.72\linewidth}
    \HPInfoBox{\linewidth}{17mm}{HPPanel}{HPLine}{QUESTION}{How many people are dancing in this picture?}
  \end{minipage}\hfill
  \begin{minipage}[t]{0.26\linewidth}
    \HPInfoBox{\linewidth}{17mm}{HPGreen!7}{HPGreen!35}{GROUND TRUTH}{\HPAnswer{7}{people}}
  \end{minipage}
  \HPSingleNote{SFT ID 000000 \quad$\bullet$\quad Numeric answer}
}

\HPSampleCard{HPBlue}{Depth}{SPATIAL STATE \quad$\bullet$\quad JRDB-ACTIVITY \quad$\bullet$\quad 1 FRAME}{%
  \HPVisualLabel{INPUT FRAME}\vspace{-2.5mm}
  \begin{center}
    \HPImage{width=\linewidth,height=49mm,keepaspectratio}{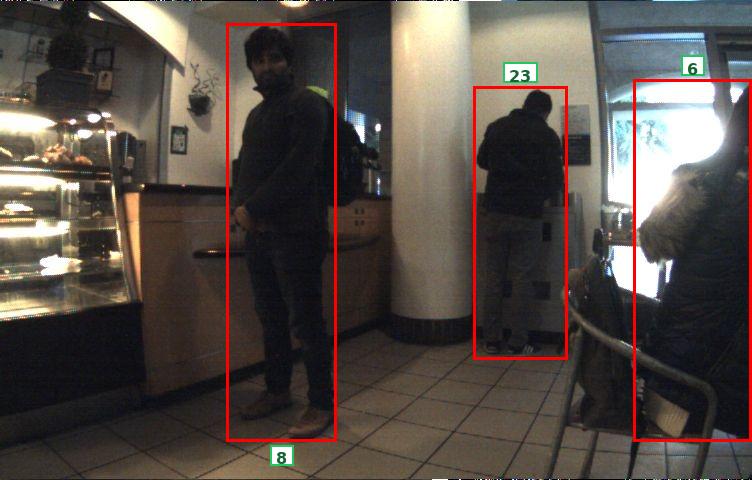}
  \end{center}
  \vspace{-1.5mm}
  \noindent\begin{minipage}[t]{0.72\linewidth}
    \HPInfoBox{\linewidth}{17mm}{HPPanel}{HPLine}{QUESTION}{What is the distance from pedestrian 23 to the camera in this picture? (in metres, to 1 decimal place)}
  \end{minipage}\hfill
  \begin{minipage}[t]{0.26\linewidth}
    \HPInfoBox{\linewidth}{17mm}{HPGreen!7}{HPGreen!35}{GROUND TRUTH}{\HPAnswer{3.1}{metres}}
  \end{minipage}
  \HPSingleNote{SFT ID 000001 \quad$\bullet$\quad Metric depth estimate}
}

\HPSampleCard{HPBlue}{Distance}{SPATIAL STATE \quad$\bullet$\quad JRDB-ACTIVITY \quad$\bullet$\quad 1 FRAME}{%
  \HPVisualLabel{INPUT FRAME}\vspace{-2.5mm}
  \begin{center}
    \HPImage{width=\linewidth,height=58mm,keepaspectratio}{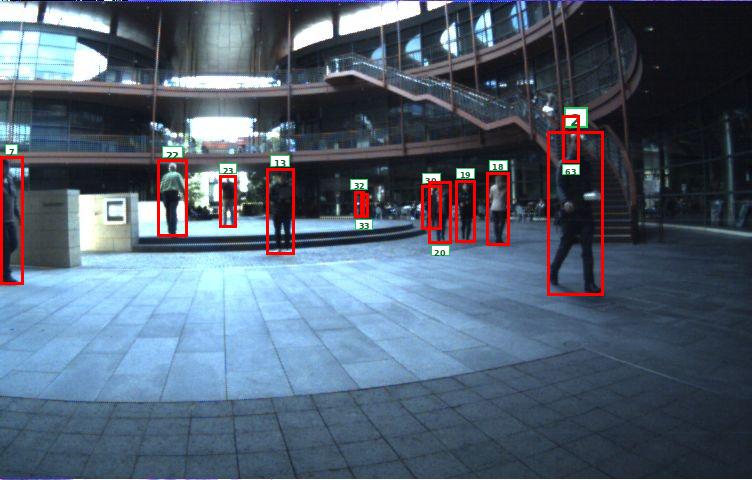}
  \end{center}
  \vspace{-1.5mm}
  \noindent\begin{minipage}[t]{0.72\linewidth}
    \HPInfoBox{\linewidth}{17mm}{HPPanel}{HPLine}{QUESTION}{What is the distance between pedestrian 13 and pedestrian 18 in this picture? (in metres, to 1 decimal place)}
  \end{minipage}\hfill
  \begin{minipage}[t]{0.26\linewidth}
    \HPInfoBox{\linewidth}{17mm}{HPGreen!7}{HPGreen!35}{GROUND TRUTH}{\HPAnswer{5.4}{metres}}
  \end{minipage}
  \HPSingleNote{SFT ID 001480 \quad$\bullet$\quad Foot-point distance between two pedestrians}
}

\HPSampleCard{HPBlue}{Gaze}{SPATIAL STATE \quad$\bullet$\quad VIDEOATTENTIONTARGET \quad$\bullet$\quad 1 FRAME}{%
  \HPVisualLabel{INPUT FRAME}\vspace{-2.5mm}
  \begin{center}
    \HPImage{width=\linewidth,height=58mm,keepaspectratio}{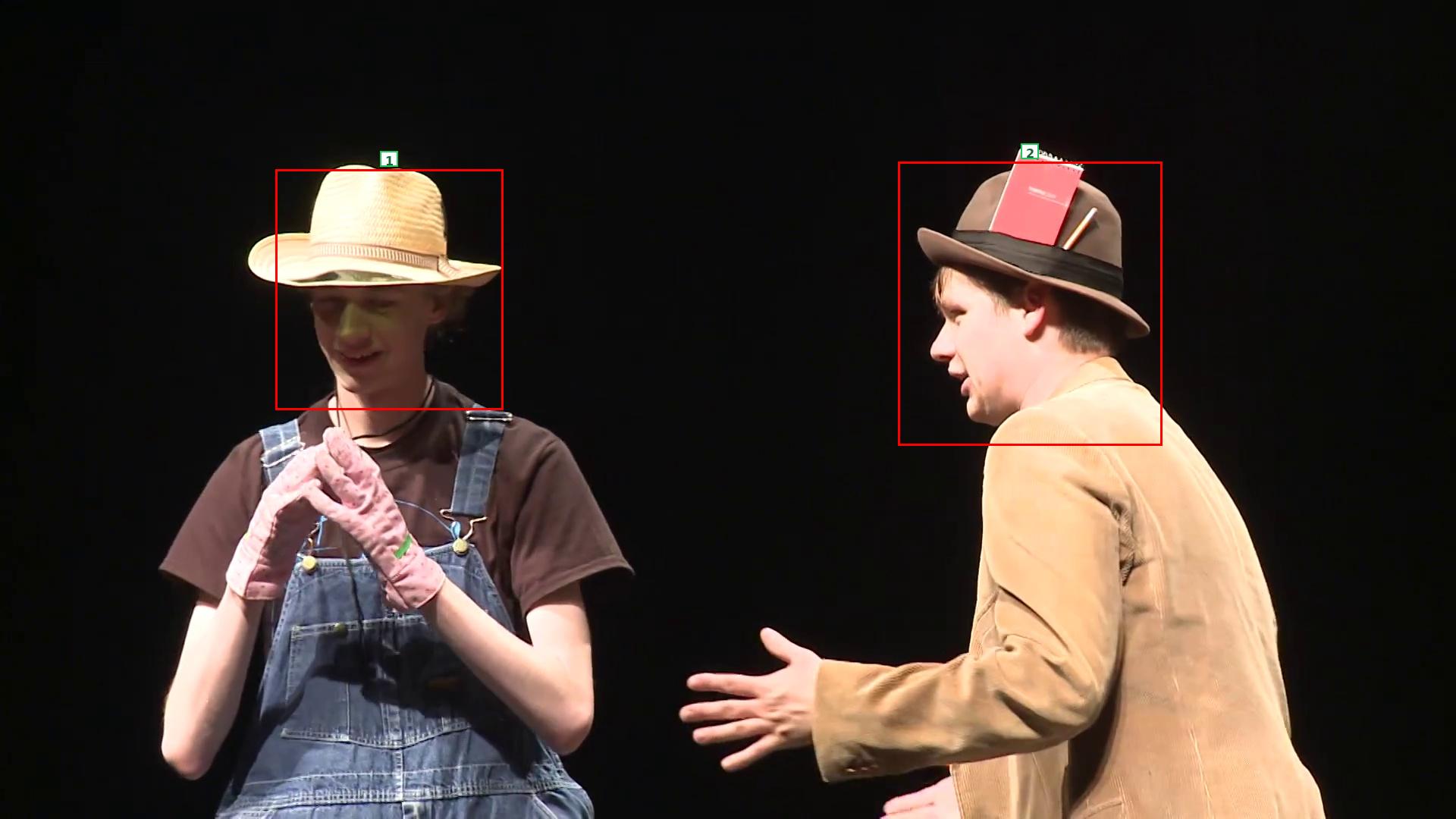}
  \end{center}
  \vspace{-1.5mm}
  \noindent\begin{minipage}[t]{0.72\linewidth}
    \HPInfoBox{\linewidth}{17mm}{HPPanel}{HPLine}{QUESTION}{What is the clockwise gaze angle of person 1 in this frame, where the upward direction is $0^\circ$? (in degrees)}
  \end{minipage}\hfill
  \begin{minipage}[t]{0.26\linewidth}
    \HPInfoBox{\linewidth}{17mm}{HPGreen!7}{HPGreen!35}{GROUND TRUTH}{\HPAnswer{$137^\circ$}{clockwise}}
  \end{minipage}
  \HPSingleNote{SFT ID 000000 \quad$\bullet$\quad Target: person 1 \quad$\bullet$\quad Image-plane gaze angle}
}

\clearpage

\HPSampleCard{HPPurple}{Displacement}{KINEMATICS \quad$\bullet$\quad JRDB-ACTIVITY \quad$\bullet$\quad 2 FRAMES}{%
  \HPVisualLabel{INPUT VIDEO FRAMES \quad\textnormal{\scriptsize (TEMPORAL ORDER)}}
  \noindent
  \HPFrame{0.49\linewidth}{52mm}{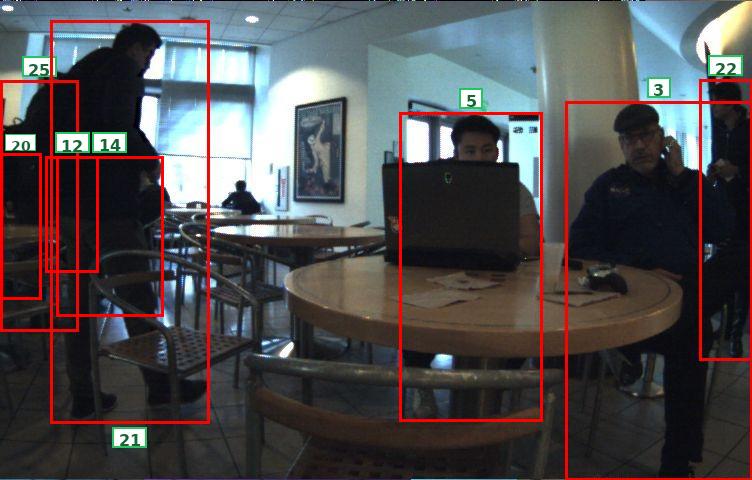}{FRAME 1}\hfill
  \HPFrame{0.49\linewidth}{52mm}{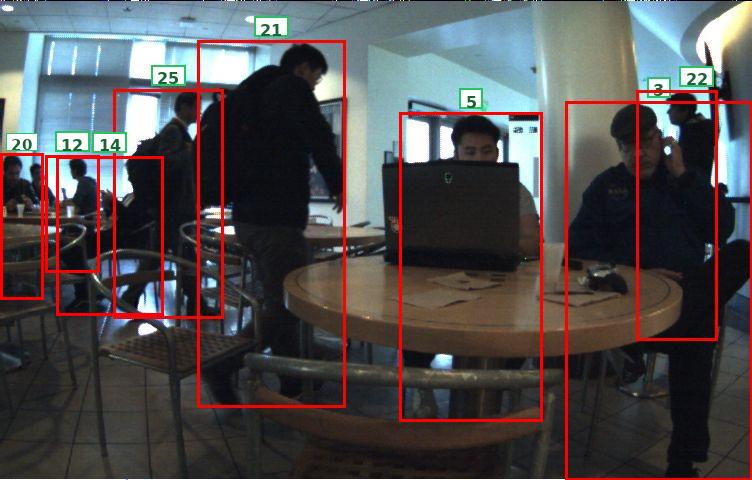}{FRAME 2}
  \par\vspace{1.2mm}
  \noindent\begin{minipage}[t]{0.72\linewidth}
    \HPInfoBox{\linewidth}{20mm}{HPPanel}{HPLine}{QUESTION}{How far has pedestrian 21 moved in the video? (in metres, to 1 decimal place)}
  \end{minipage}\hfill
  \begin{minipage}[t]{0.26\linewidth}
    \HPInfoBox{\linewidth}{20mm}{HPGreen!7}{HPGreen!35}{GROUND TRUTH}{\HPAnswer{0.8}{metres}}
  \end{minipage}
  \HPSampleNote{SFT ID 000001 \quad$\bullet$\quad Target: pedestrian 21 \quad$\bullet$\quad Same motion clip as Velocity below}
}

\HPSampleCard{HPPurple}{Velocity}{KINEMATICS \quad$\bullet$\quad JRDB-ACTIVITY \quad$\bullet$\quad 2 FRAMES}{%
  \HPVisualLabel{INPUT VIDEO FRAMES \quad\textnormal{\scriptsize (1.00-SECOND INTERVAL)}}
  \noindent
  \HPFrame{0.49\linewidth}{52mm}{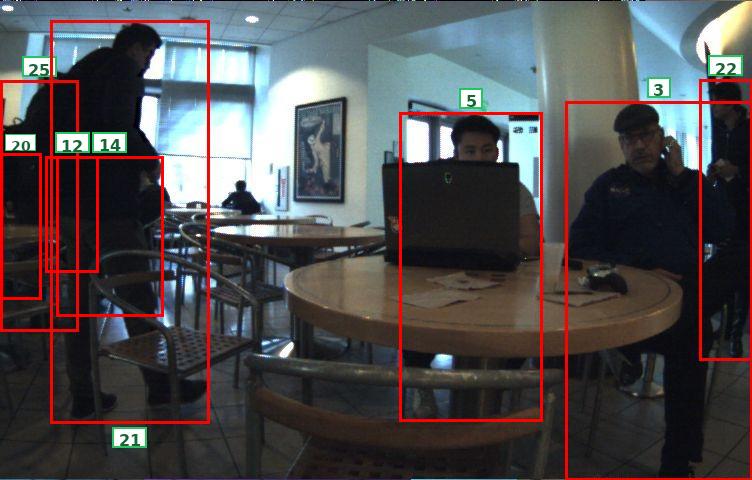}{FRAME 1 $\bullet$ $t=0$ s}\hfill
  \HPFrame{0.49\linewidth}{52mm}{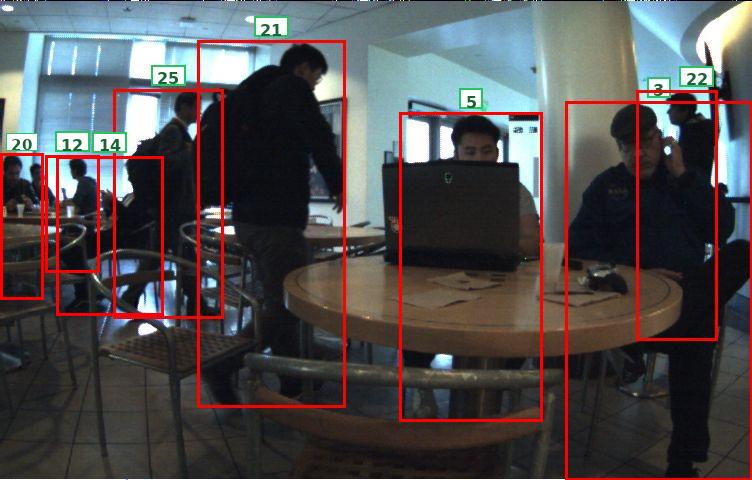}{FRAME 2 $\bullet$ $t=1$ s}
  \par\vspace{1.2mm}
  \noindent\begin{minipage}[t]{0.72\linewidth}
    \HPInfoBox{\linewidth}{20mm}{HPPanel}{HPLine}{QUESTION}{The video frames are sampled at 1.00-second intervals. What is the velocity of pedestrian 21 in the video? (in m/s, to 1 decimal place)}
  \end{minipage}\hfill
  \begin{minipage}[t]{0.26\linewidth}
    \HPInfoBox{\linewidth}{20mm}{HPGreen!7}{HPGreen!35}{GROUND TRUTH}{\HPAnswer{0.8}{m/s}}
  \end{minipage}
  \HPSampleNote{SFT ID 000000 \quad$\bullet$\quad Target: pedestrian 21 \quad$\bullet$\quad Path length divided by elapsed time}
}

\clearpage

\HPSampleCard{HPPurple}{Trajectory}{KINEMATICS \quad$\bullet$\quad JRDB-ACTIVITY \quad$\bullet$\quad 3 FRAMES}{%
  \HPVisualLabel{INPUT VIDEO FRAMES \quad\textnormal{\scriptsize (TEMPORAL ORDER)}}
  \noindent
  \HPFrame{0.327\linewidth}{45mm}{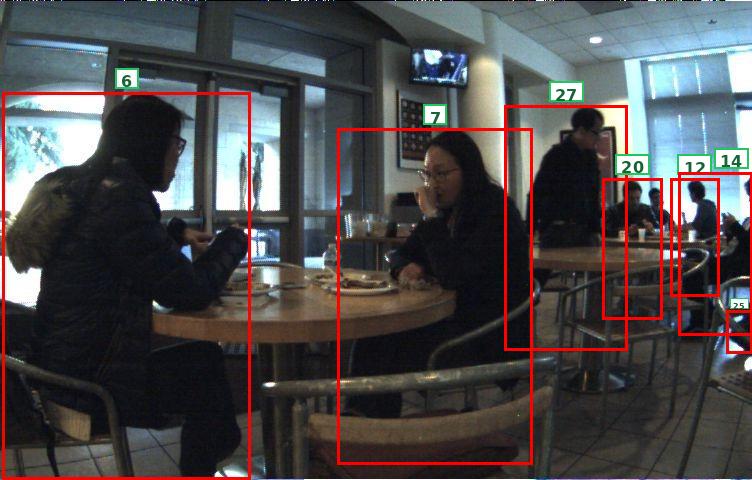}{FRAME 1}\hfill
  \HPFrame{0.327\linewidth}{45mm}{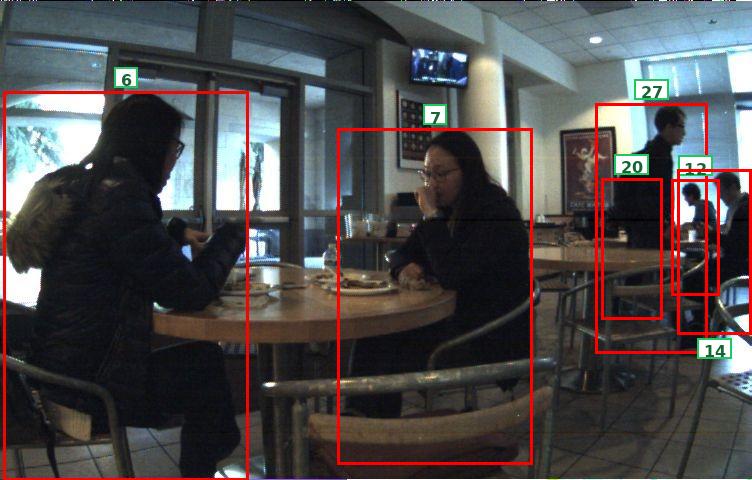}{FRAME 2}\hfill
  \HPFrame{0.327\linewidth}{45mm}{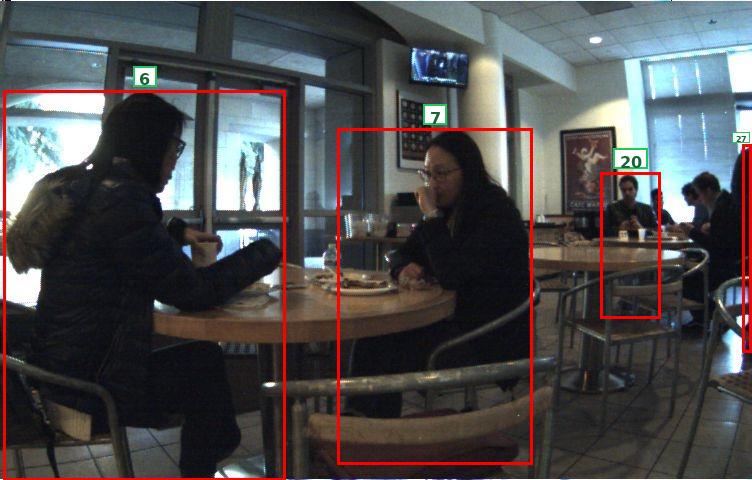}{FRAME 3}
  \par\vspace{1.2mm}
  \noindent\begin{minipage}[t]{0.63\linewidth}
    \HPInfoBox{\linewidth}{24mm}{HPPanel}{HPLine}{QUESTION}{What is the trajectory of pedestrian 27 in this video? Report \texttt{[turn\_deg, distance\_m]} for each step: signed turn from the previous displacement (positive = left) and displacement in metres; the first turn is $0$.}
  \end{minipage}\hfill
  \begin{minipage}[t]{0.35\linewidth}
    \HPInfoBox{\linewidth}{24mm}{HPGreen!7}{HPGreen!35}{GROUND TRUTH}{{\ttfamily\bfseries\footnotesize\color{HPGreen}[[0.0, 0.7],\newline[-23.6, 0.8]]}}
  \end{minipage}
  \HPSampleNote{SFT ID 000014 \quad$\bullet$\quad Target: pedestrian 27 \quad$\bullet$\quad Ordered turn/step sequence}
}

\HPSampleCard{HPOrange}{Blink}{EVENT \quad$\bullet$\quad EYEBLINK8 \quad$\bullet$\quad 4 FRAMES}{%
  \HPVisualLabel{INPUT VIDEO FRAMES \quad\textnormal{\scriptsize (TEMPORAL ORDER)}}
  \noindent
  \HPFrame{0.24\linewidth}{42mm}{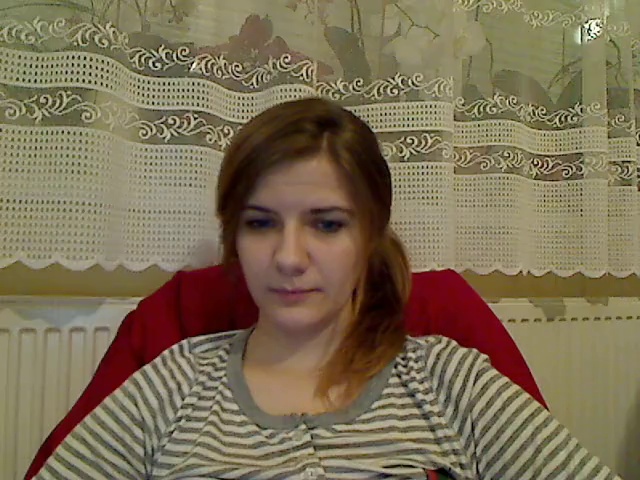}{FRAME 1}\hfill
  \HPFrame{0.24\linewidth}{42mm}{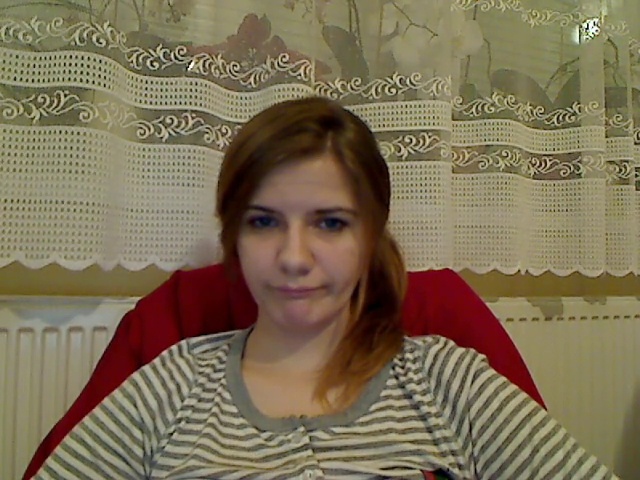}{FRAME 2}\hfill
  \HPFrame{0.24\linewidth}{42mm}{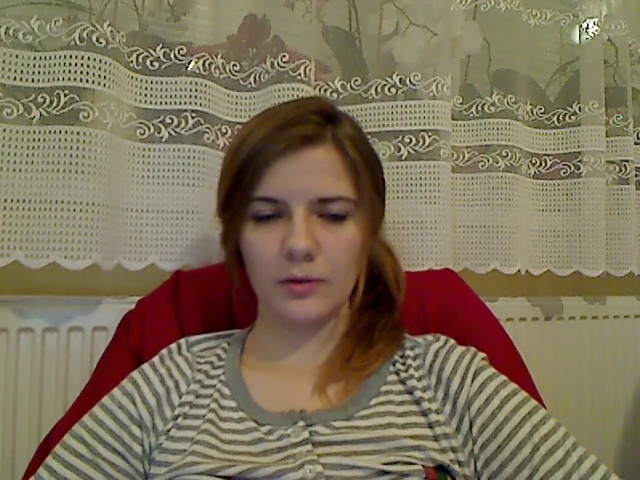}{FRAME 3}\hfill
  \HPFrame{0.24\linewidth}{42mm}{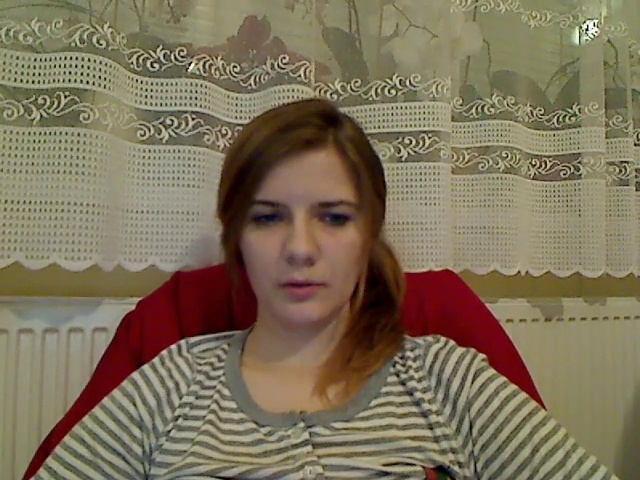}{FRAME 4}
  \par\vspace{1.2mm}
  \noindent\begin{minipage}[t]{0.72\linewidth}
    \HPInfoBox{\linewidth}{17mm}{HPPanel}{HPLine}{QUESTION}{How many times did the person blink in this video?}
  \end{minipage}\hfill
  \begin{minipage}[t]{0.26\linewidth}
    \HPInfoBox{\linewidth}{17mm}{HPGreen!7}{HPGreen!35}{GROUND TRUTH}{\HPAnswer{1}{blink}}
  \end{minipage}
  \HPSampleNote{SFT ID 000000 \quad$\bullet$\quad Discrete event count over the complete clip}
}

\clearpage

\HPSampleCard{HPRose}{Deform}{DEFORMATION \quad$\bullet$\quad DECAF \quad$\bullet$\quad 8 FRAMES}{%
  \HPVisualLabel{INPUT VIDEO FRAMES \quad\textnormal{\scriptsize (EACH PANEL: TWO VIEWS $\times$ OPTIONS A--D)}}
  \noindent
  \HPFrame{0.24\linewidth}{27mm}{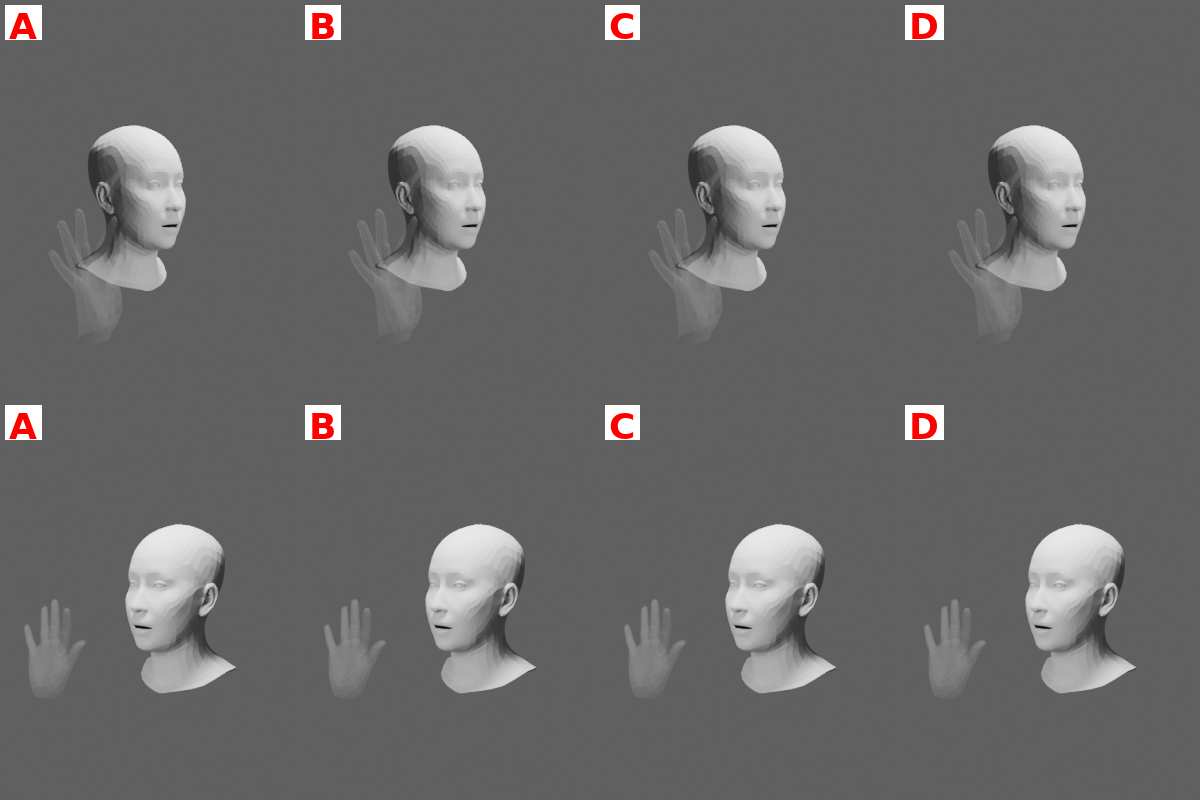}{FRAME 1}\hfill
  \HPFrame{0.24\linewidth}{27mm}{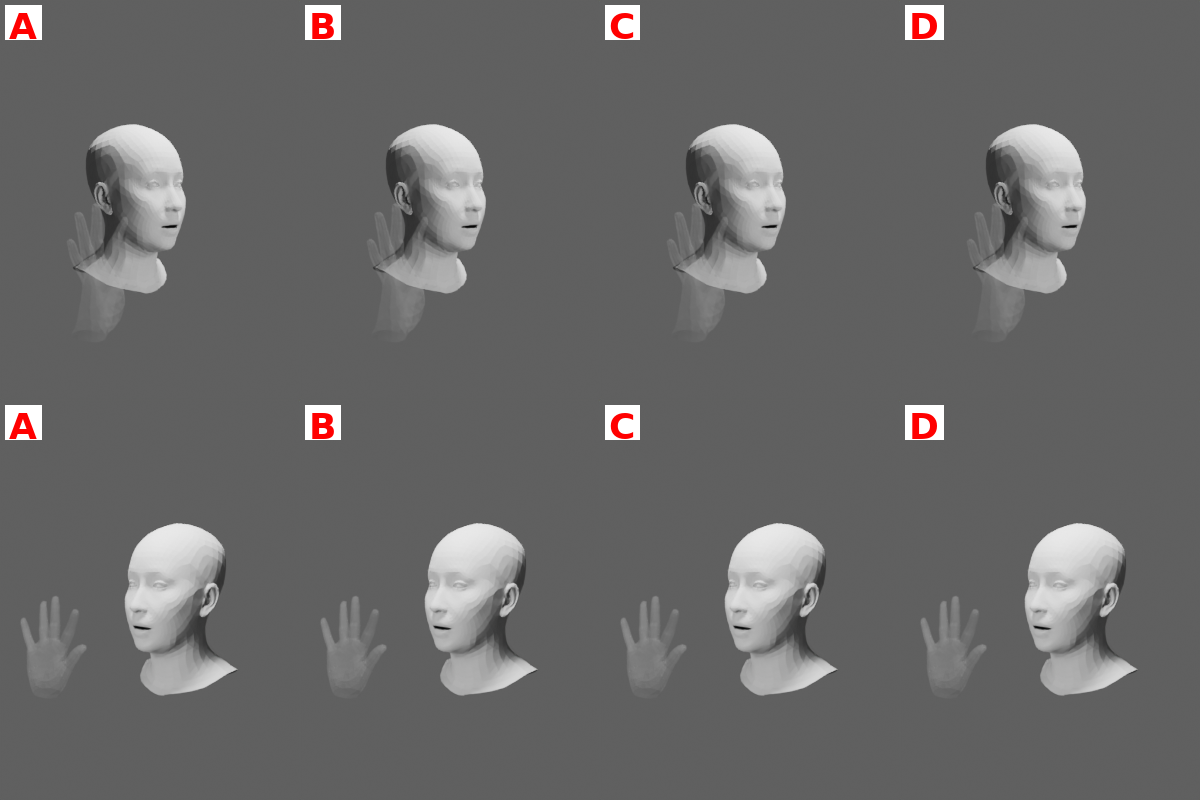}{FRAME 2}\hfill
  \HPFrame{0.24\linewidth}{27mm}{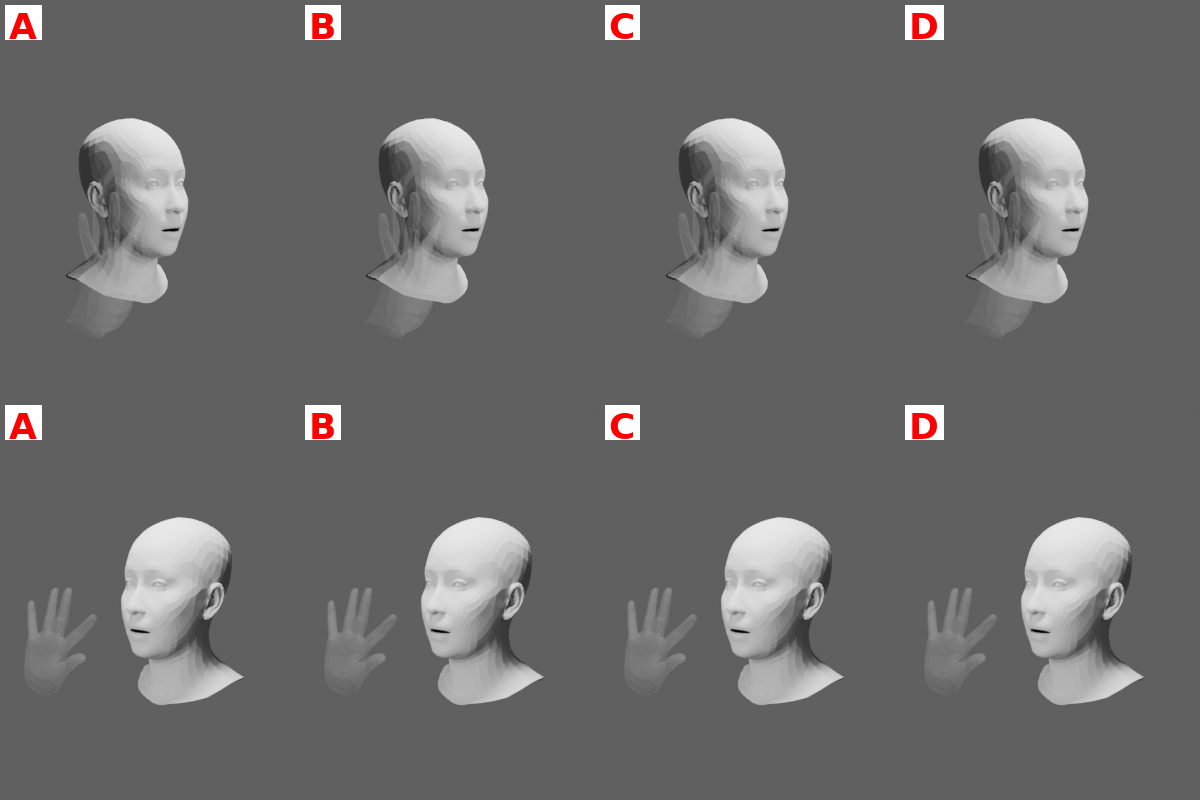}{FRAME 3}\hfill
  \HPFrame{0.24\linewidth}{27mm}{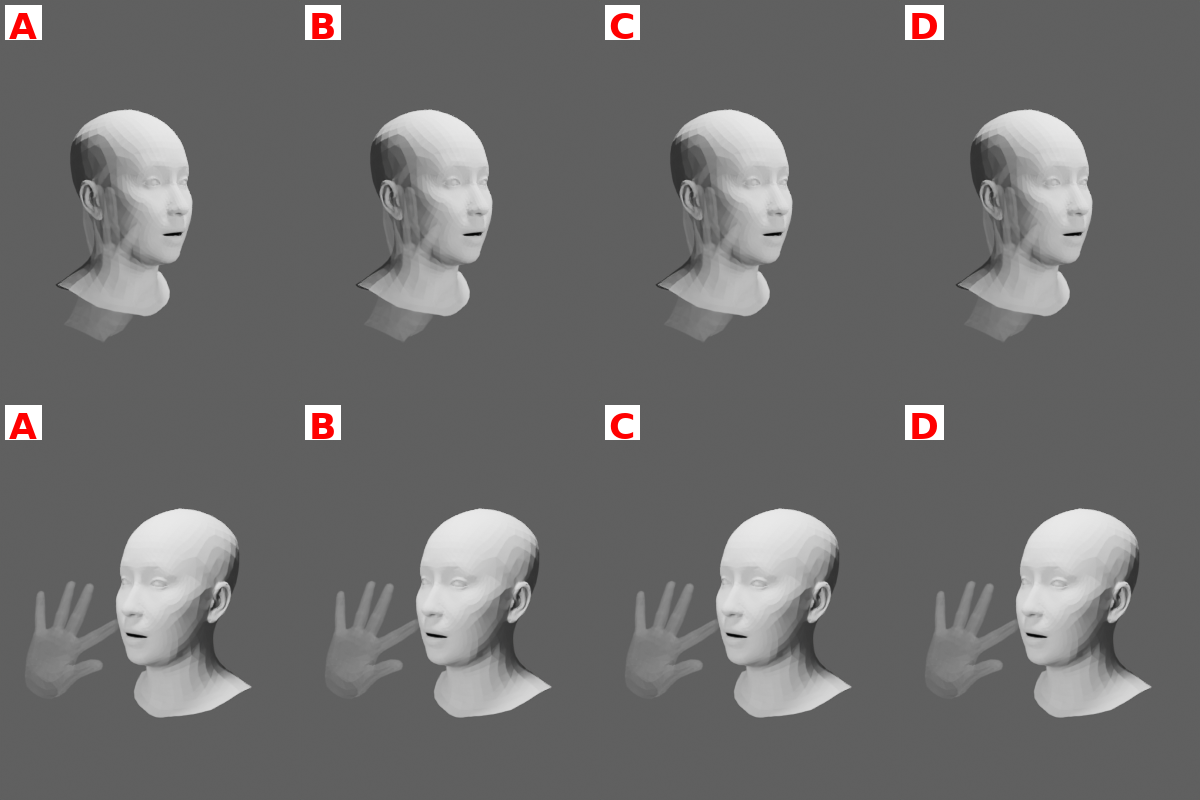}{FRAME 4}
  \par\vspace{0.8mm}\noindent
  \HPFrame{0.24\linewidth}{27mm}{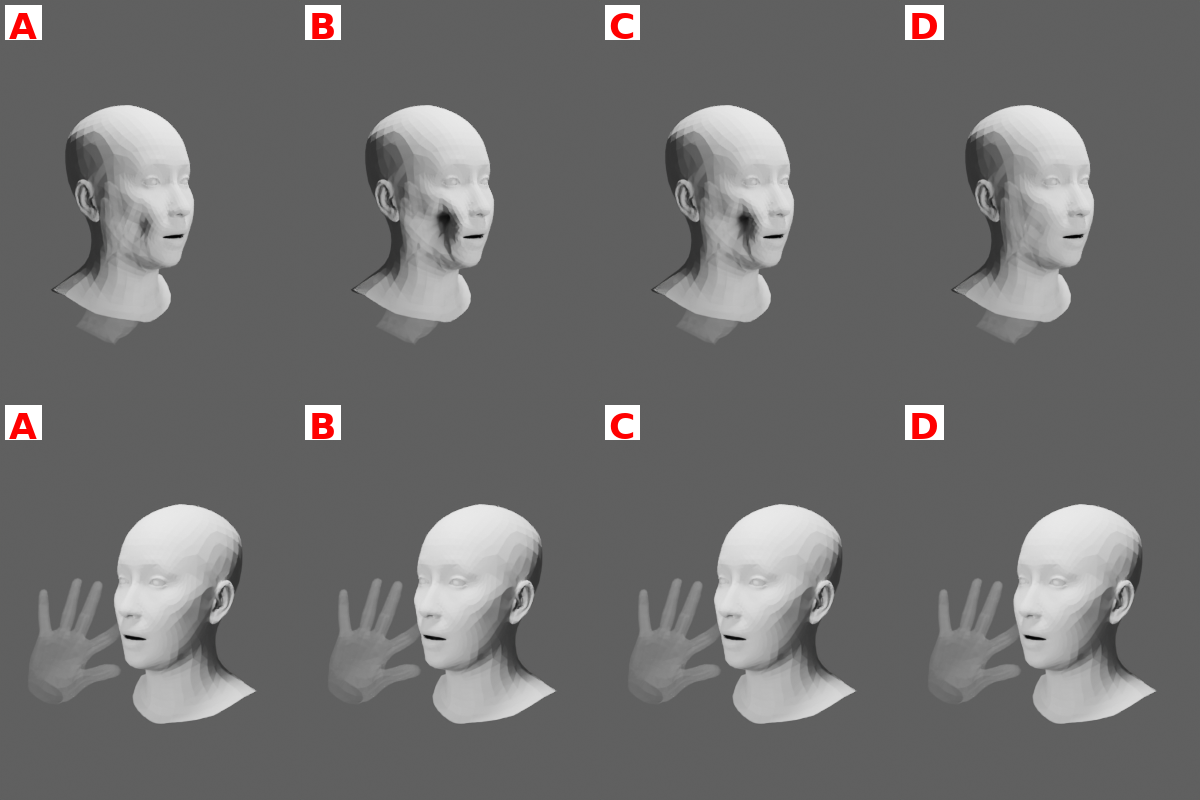}{FRAME 5}\hfill
  \HPFrame{0.24\linewidth}{27mm}{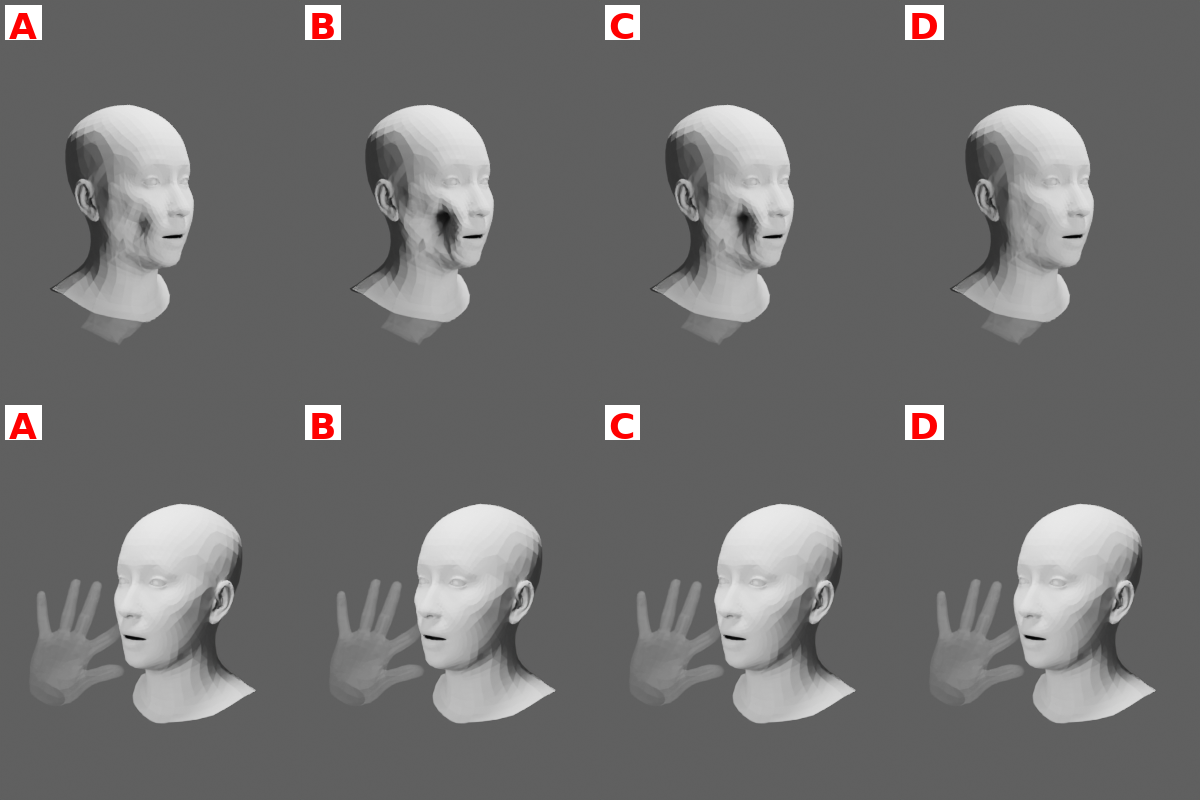}{FRAME 6}\hfill
  \HPFrame{0.24\linewidth}{27mm}{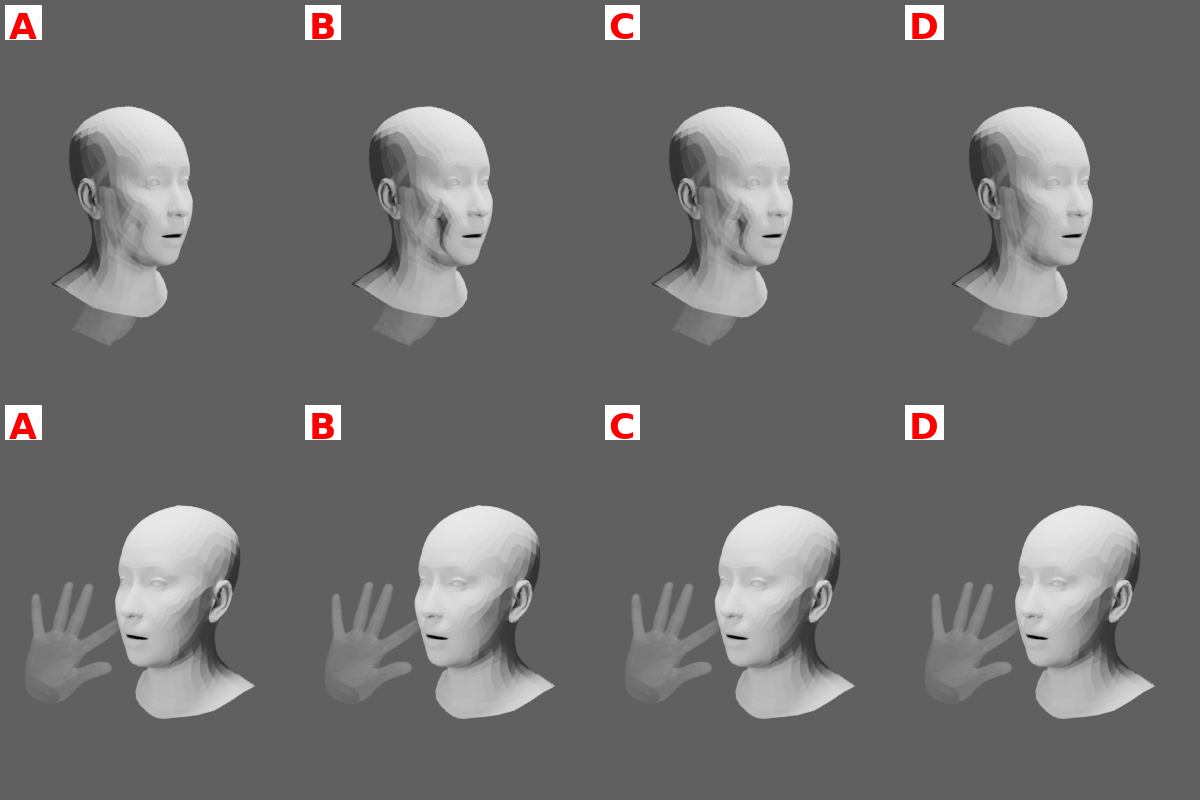}{FRAME 7}\hfill
  \HPFrame{0.24\linewidth}{27mm}{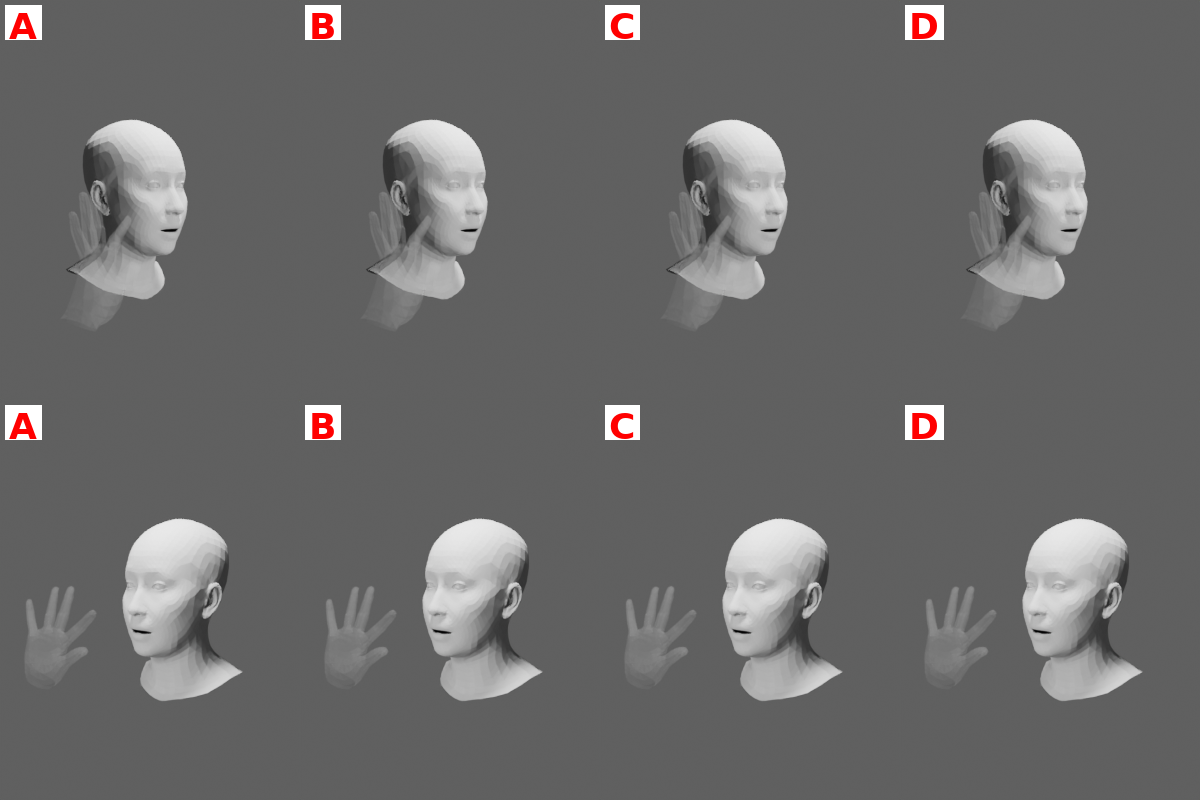}{FRAME 8}
  \par\vspace{1.2mm}
  \noindent\begin{minipage}[t]{0.63\linewidth}
    \HPInfoBox{\linewidth}{21mm}{HPPanel}{HPLine}{QUESTION}{Which face's deformation best matches physical reality?}
  \end{minipage}\hfill
  \begin{minipage}[t]{0.35\linewidth}
    \HPInfoBox{\linewidth}{21mm}{HPGreen!7}{HPGreen!35}{GROUND TRUTH}{\HPAnswer{A}{$\alpha=1.0$}\par\vspace{0.3mm}{\footnotesize\color{HPMuted}Peak deformation: $\approx2.4$ cm}}
  \end{minipage}
  \par\vspace{1.1mm}
  \HPInfoBox{\linewidth}{24mm}{HPRose!4}{HPRose!22}{GROUND-TRUTH RATIONALE}{\footnotesize $\alpha$ scales the measured contact deformation: $\alpha=1$ is physically real, $\alpha>1$ over-deforms, and $\alpha<1$ under-deforms. \textbf{A} ($\alpha=1.0$) matches reality; \textbf{B} ($\alpha=3.0$) and \textbf{C} ($\alpha=2.3$) produce 300\% and 230\% of the real deformation; \textbf{D} ($\alpha=0.3$) produces only 30\%. Therefore the correct face is \textbf{A}.}
  \HPSampleNote{SFT ID 000055 \quad$\bullet$\quad FLAME head with MANO hand contact \quad$\bullet$\quad Multiple-choice answer with deformation-strength rationale}
}

\clearpage

\HPSampleCard{HPTeal}{LS-Head}{ILLUMINATION \quad$\bullet$\quad BEDLAM SELFIE \quad$\bullet$\quad 4 FRAMES}{%
  \HPVisualLabel{INPUT VIDEO FRAMES \quad\textnormal{\scriptsize (ONE ACTIVE LIGHT PER FRAME)}}
  \noindent
  \HPFrame{0.24\linewidth}{32mm}{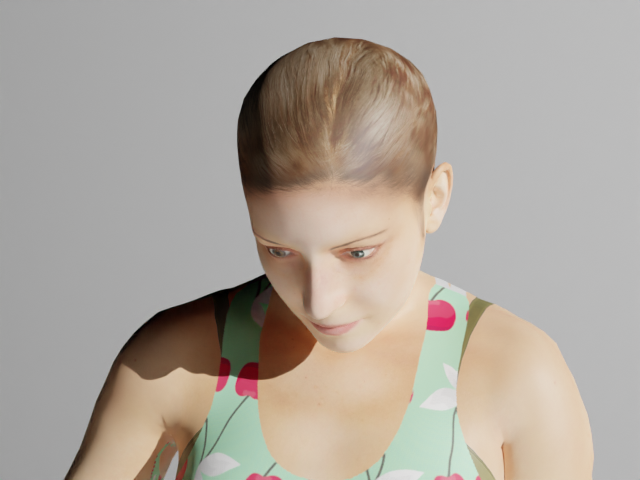}{FRAME 1 $\bullet$ $(-1,1,0)$}\hfill
  \HPFrame{0.24\linewidth}{32mm}{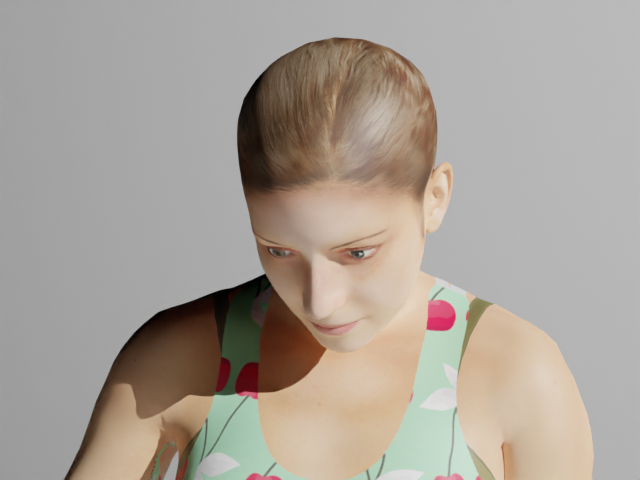}{FRAME 2}\hfill
  \HPFrame{0.24\linewidth}{32mm}{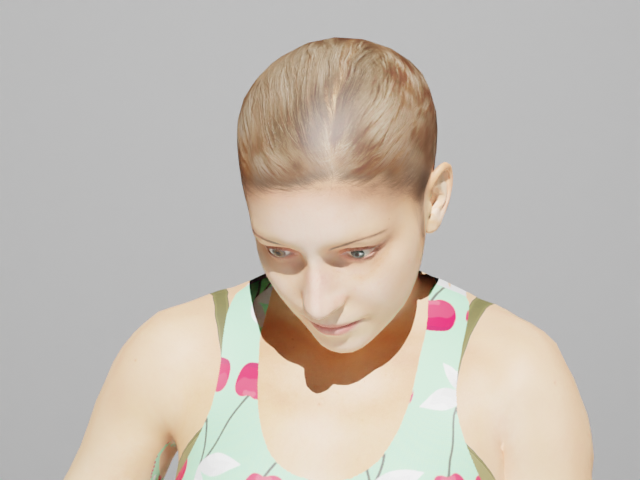}{FRAME 3}\hfill
  \HPFrame{0.24\linewidth}{32mm}{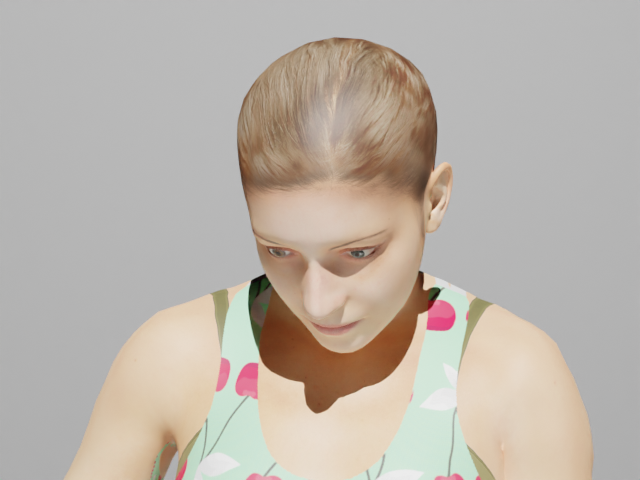}{FRAME 4}
  \par\vspace{1.2mm}
  \noindent\begin{minipage}[t]{0.69\linewidth}
    \HPInfoBox{\linewidth}{34mm}{HPPanel}{HPLine}{QUESTION}{\footnotesize A $3\!\times\!3\!\times\!3$ grid of 27 point lights surrounds the person: $x,y\in\{-1,0,1\}$, $z\in\{0,1,2\}$, $(0,0,0)$ is above the head, and $+z$ is up. The camera faces $-y$ and image-left is $+x$; unit spacing is 2.35 m in $x/y$ and 0.50 m in $z$. The active light moves one unit along a single axis per step. Starting from $(-1,1,0)$, predict the light coordinates for the remaining 3 frames.}
  \end{minipage}\hfill
  \begin{minipage}[t]{0.29\linewidth}
    \HPInfoBox{\linewidth}{34mm}{HPGreen!7}{HPGreen!35}{GROUND TRUTH}{{\ttfamily\bfseries\scriptsize\color{HPGreen}[[-1, 1, 1],\newline\ [0, 1, 1],\newline\ [0, 1, 2]]}}
  \end{minipage}
  \HPSampleNote{SFT ID 000301 \quad$\bullet$\quad Facial shading cue \quad$\bullet$\quad Single-axis light walk}
}

\HPSampleCard{HPTeal}{LS-Body}{ILLUMINATION \quad$\bullet$\quad BEDLAM WALL CAMERA \quad$\bullet$\quad 3 FRAMES}{%
  \HPVisualLabel{INPUT VIDEO FRAMES \quad\textnormal{\scriptsize (BODY SHADING AND CAST-SHADOW CUES)}}
  \noindent
  \HPFrame{0.327\linewidth}{38mm}{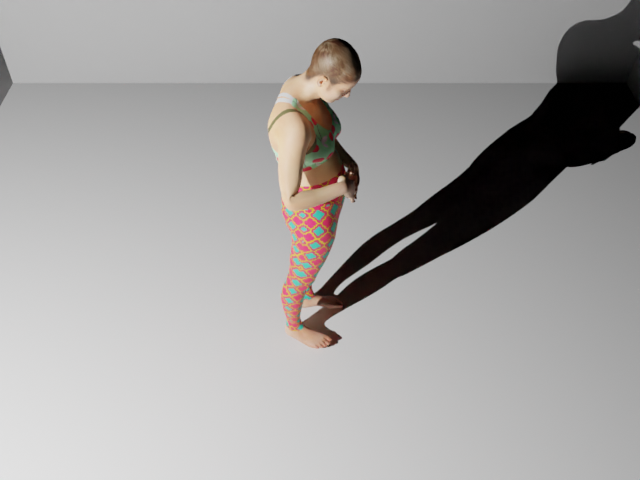}{FRAME 1 $\bullet$ $(1,-1,0)$}\hfill
  \HPFrame{0.327\linewidth}{38mm}{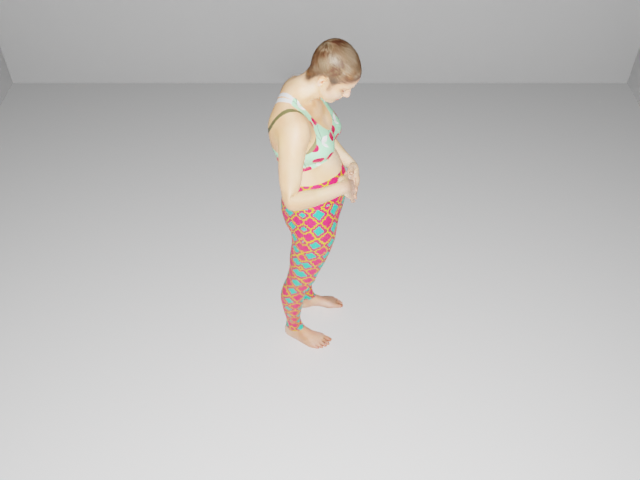}{FRAME 2}\hfill
  \HPFrame{0.327\linewidth}{38mm}{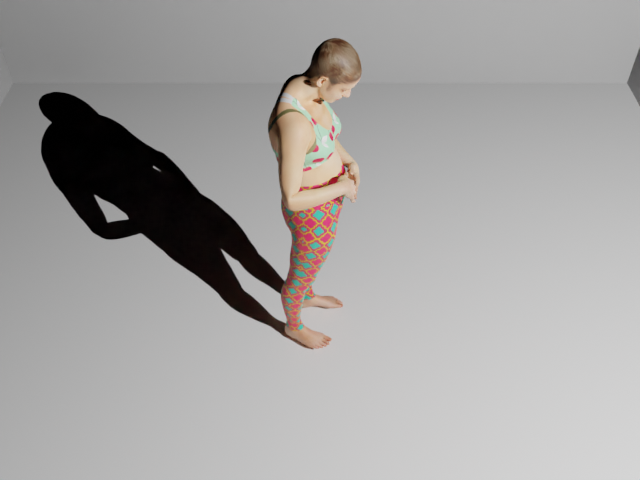}{FRAME 3}
  \par\vspace{1.2mm}
  \noindent\begin{minipage}[t]{0.69\linewidth}
    \HPInfoBox{\linewidth}{34mm}{HPPanel}{HPLine}{QUESTION}{\footnotesize A $3\!\times\!3\!\times\!3$ grid of 27 point lights surrounds the person: $x,y\in\{-1,0,1\}$, $z\in\{0,1,2\}$, $(0,0,0)$ is above the head, and $+z$ is up. The camera faces $-x$ and image-left is $-y$; unit spacing is 2.35 m in $x/y$ and 0.50 m in $z$. Each step may move along one axis, a face diagonal, or a body diagonal (maximum $\sqrt{3}$ units). Starting from $(1,-1,0)$, predict the light coordinates for the remaining 2 frames.}
  \end{minipage}\hfill
  \begin{minipage}[t]{0.29\linewidth}
    \HPInfoBox{\linewidth}{34mm}{HPGreen!7}{HPGreen!35}{GROUND TRUTH}{{\ttfamily\bfseries\scriptsize\color{HPGreen}[[1, 0, 1],\newline\ [1, 1, 2]]}}
  \end{minipage}
  \HPSampleNote{SFT ID 000004 \quad$\bullet$\quad Body and cast-shadow cues \quad$\bullet$\quad Fast lattice walk}
}

\clearpage

\HPSampleCard{HPCamera}{CP-Room}{CAMERA POSE \quad$\bullet$\quad REPLICACAD \quad$\bullet$\quad 3 FRAMES}{%
  \HPVisualLabel{INPUT VIDEO FRAMES \quad\textnormal{\scriptsize (SCENE-LEVEL EGO-MOTION)}}
  \noindent
  \HPFrame{0.327\linewidth}{38mm}{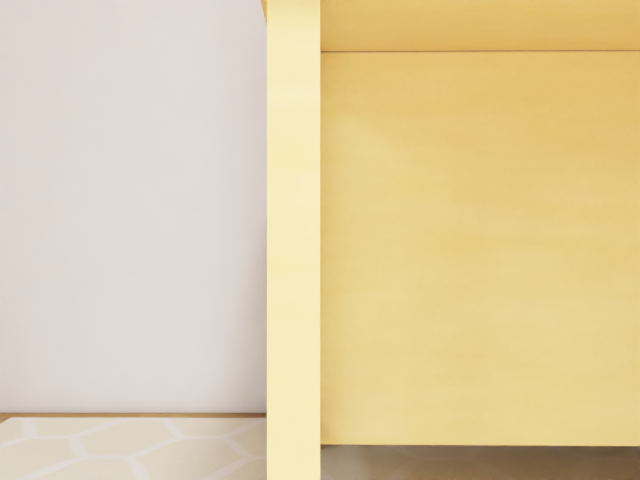}{FRAME 1 $\bullet$ REFERENCE POSE}\hfill
  \HPFrame{0.327\linewidth}{38mm}{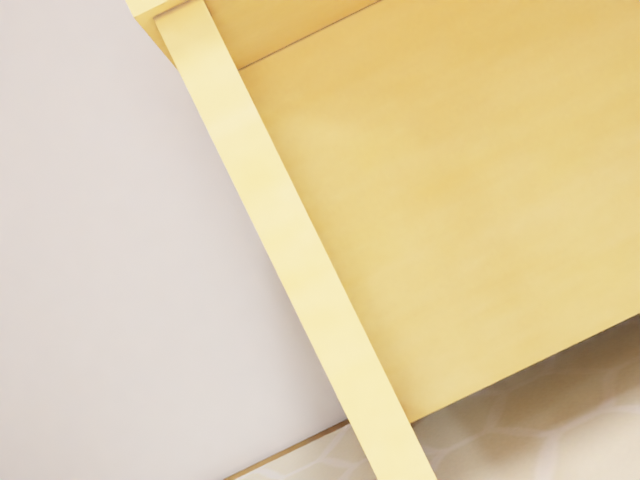}{FRAME 2}\hfill
  \HPFrame{0.327\linewidth}{38mm}{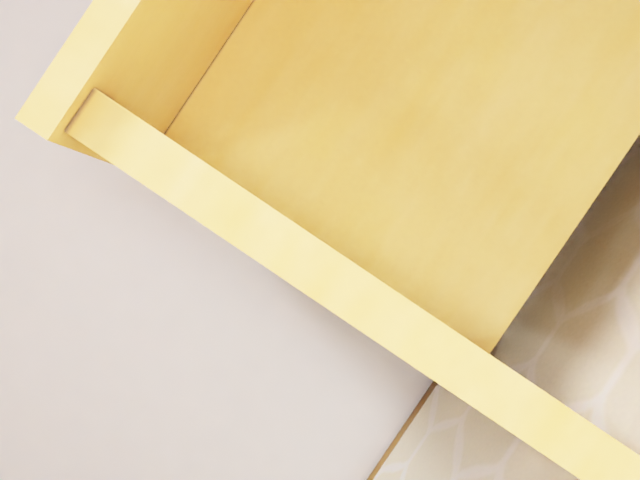}{FRAME 3}
  \par\vspace{1.2mm}
  \noindent\begin{minipage}[t]{0.67\linewidth}
    \HPInfoBox{\linewidth}{35mm}{HPPanel}{HPLine}{QUESTION}{\footnotesize A free-flying camera moves on a cubic lattice (1 step $=0.20$ m). Fixed axes are defined at frame 1: the camera faces $+X$, $+Z$ is up, and $+Y$ is left. Each subsequent frame results from exactly one atomic action: translate 0.20 m along a fixed axis, or rotate in place using yaw, pitch, or roll by an integer angle in $\pm1^\circ$--$\pm30^\circ$ (pitch capped at $+60^\circ$). Positive yaw turns left, positive pitch looks up, and positive roll tilts right. Predict the 2 actions.}
  \end{minipage}\hfill
  \begin{minipage}[t]{0.31\linewidth}
    \HPInfoBox{\linewidth}{35mm}{HPGreen!7}{HPGreen!35}{GROUND TRUTH}{{\ttfamily\bfseries\scriptsize\color{HPGreen}[['roll', 26],\newline\ ['roll', 30]]}}
  \end{minipage}
  \HPSampleNote{Corpus record \quad$\bullet$\quad Camera-only motion without a person}
}

\HPSampleCard{HPCamera}{CP-Human}{CAMERA POSE \quad$\bullet$\quad REPLICACAD \quad$\bullet$\quad 3 FRAMES}{%
  \HPVisualLabel{INPUT VIDEO FRAMES \quad\textnormal{\scriptsize (STATIC HUMAN ANCHOR)}}
  \noindent
  \HPFrame{0.327\linewidth}{38mm}{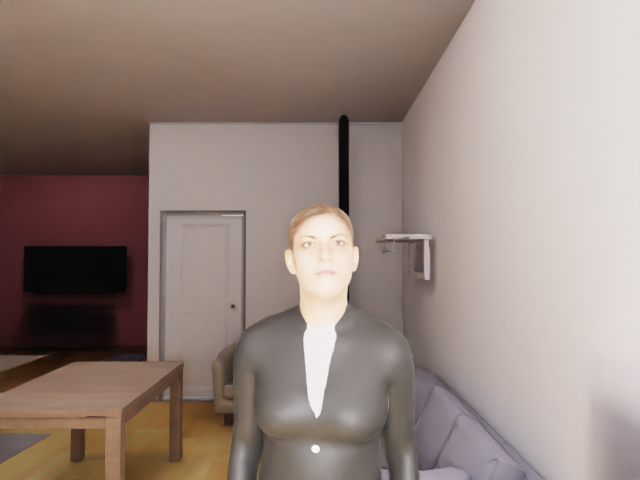}{FRAME 1 $\bullet$ REFERENCE POSE}\hfill
  \HPFrame{0.327\linewidth}{38mm}{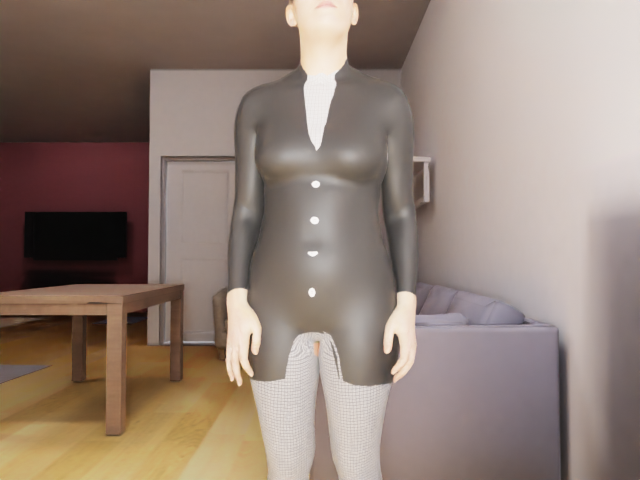}{FRAME 2}\hfill
  \HPFrame{0.327\linewidth}{38mm}{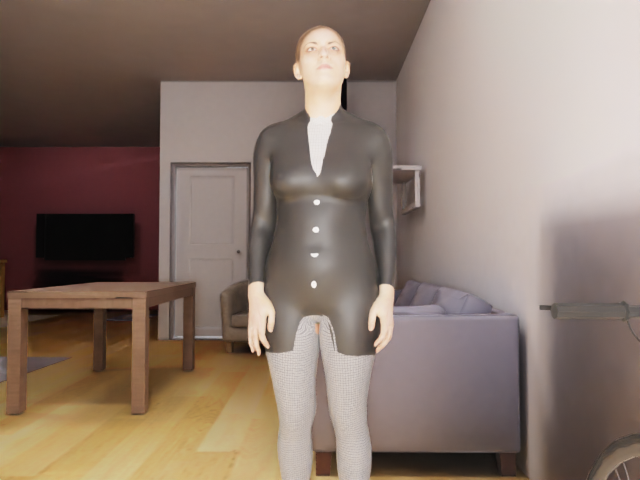}{FRAME 3}
  \par\vspace{1.2mm}
  \noindent\begin{minipage}[t]{0.67\linewidth}
    \HPInfoBox{\linewidth}{35mm}{HPPanel}{HPLine}{QUESTION}{\footnotesize A camera moves around a static person on a 0.20 m lattice, performing one action per frame: translate 1--3 steps along a fixed axis, or rotate by an integer angle in $\pm1^\circ$--$\pm30^\circ$. At frame 1, roll is zero and $+Z$ points up; $-Y$ lies along the view, $+Y$ behind, $+X$ is $90^\circ$ left, and $-X$ is $90^\circ$ right. Positive yaw turns left, positive pitch looks up, and positive roll tilts right; pitch never exceeds $+60^\circ$. Predict the 2 actions.}
  \end{minipage}\hfill
  \begin{minipage}[t]{0.31\linewidth}
    \HPInfoBox{\linewidth}{35mm}{HPGreen!7}{HPGreen!35}{GROUND TRUTH}{{\ttfamily\bfseries\scriptsize\color{HPGreen}[['-Z', 0.6],\newline\ ['+Y', 0.4]]}}
  \end{minipage}
  \HPSampleNote{Corpus record \quad$\bullet$\quad Translation magnitudes are in metres}
}

\twocolumn
\flushbottom
\section{Training-free Method}
\label{app:toolcalling}

We augment GPT-6-Sol with programmatic measurement tools for metric depth,
foot-point distance, multi-frame displacement, velocity, trajectory, and camera motion.
The agent invokes pinhole geometry, VGGT~\citep{vggt}, and a calculator when useful, and conditions
its final answer on the returned observations. The training-free reward comparison over
1,050 paired questions is reported in Section~\ref{sec:training_free}; this appendix
reproduces the measurement-agent prompt and representative tool-interleaved
cases.

\subsection{Pinhole Measurement}
\label{app:pinhole}

The pinhole tool is a calibrated sensor-assisted height ruler for JRDB. Its runtime
provides the official camera intrinsics and Brown--Conrady distortion parameters; these
task-specific parameters are not supplied to the No Tool baseline or to VGGT. Given raw
head and foot clicks, we first undistort both pixels with this calibration. Let the
rectified clicks be $(u_h,v_h)$ and
$(u_f,v_f)$, and let $H$ be the person's estimated height in metres. For a
standing person, the vertical image extent $\Delta v=v_f-v_h$ gives optical-axis
depth and horizontal camera-to-person range:
\begin{equation}
Z=\frac{Hf_y}{\Delta v},\qquad
r=Z\sqrt{1+q^2},\qquad
q=\frac{(u_h+u_f)/2-c_x}{f_x}.
\label{eq:pinhole_depth}
\end{equation}
The off-axis correction $r$ is important for JRDB's wide-angle images; $Z$ alone
would underestimate the horizontal range of people near the image boundary.
Back-projecting the foot click at depth $Z$ gives its camera-frame position
\begin{equation}
\mathbf F=(X,Y,Z)=
\left(\frac{(u_f-c_x)Z}{f_x},\frac{(v_f-c_y)Z}{f_y},Z\right).
\label{eq:pinhole_foot}
\end{equation}
For two people $A$ and $B$, feet-to-feet distance is
$D_{AB}=\sqrt{(X_A-X_B)^2+(Z_A-Z_B)^2}$. For a track of $T$ frames, displacement
is $L=\sum_{t=1}^{T-1}D_{t,t+1}$ and velocity is $L/((T-1)\Delta t)$. Trajectory
steps use the same ground-plane lengths and signed turn angles.

\begin{lstlisting}[language=Python,basicstyle=\ttfamily\footnotesize,
  columns=fullflexible,breaklines=true,breakatwhitespace=false,
  showstringspaces=false,frame=single,framesep=1.5mm,
  rulecolor=\color{HPLine},backgroundcolor=\color{HPPanel},
  aboveskip=2mm,belowskip=2mm,
  caption={Simplified implementation of the pinhole height ruler.},
  label={lst:pinhole},captionpos=b]
from math import hypot, sqrt

def measure_person(head, foot, height_m, K):
    # K contains calibrated fx, fy, cx, cy and distortion coefficients.
    (uh, vh), (uf, vf) = undistort([head, foot], K)
    if vf - vh < 4.0:
        raise ValueError("head-foot pixel span is too small")

    Z = height_m * K["fy"] / (vf - vh)
    q = (0.5 * (uh + uf) - K["cx"]) / K["fx"]
    depth_m = Z * sqrt(1.0 + q * q)  # horizontal range

    X = (uf - K["cx"]) * Z / K["fx"]
    Y = (vf - K["cy"]) * Z / K["fy"]
    return depth_m, (X, Y, Z)        # X right, Y down, Z forward

def ground_distance(F, G):
    return hypot(F[0] - G[0], F[2] - G[2])

# Two people in one frame: distance_m = ground_distance(F_A, F_B).
# One tracked person: L = sum(ground_distance(F[t], F[t+1])
#                          for t in range(len(F) - 1))
# speed_m_per_s = L / ((len(F) - 1) * seconds_per_frame)
\end{lstlisting}
Listing~\ref{lst:pinhole} isolates the core geometry. The calibrated sensor-assisted setting is intentionally privileged relative to unassisted monocular
inference: the environment supplies $K$ and the distortion coefficients, while the agent
supplies visual head--foot clicks and its own height estimate. The gold metric answer is
not passed to the tool. The deployed tool additionally
validates clicks and heights, reports border and out-of-range warnings, median-filters
the ground-plane path, regularizes optical depth for trajectory angles, and returns
\texttt{unsupported} when calibration is unavailable so that the agent can use VGGT.

\onecolumn
\raggedbottom
\subsection{Measurement-agent Prompt}
\label{app:toolcalling_prompt}

The complete measurement-agent prompt is reproduced below. Only YAML string escaping and version markers have been removed for readability; the instruction content, including the tool-specific call documentation, is unchanged.

\lstdefinestyle{HPPrompt}{
  basicstyle=\ttfamily\footnotesize,
  breaklines=true,
  breakatwhitespace=false,
  breakindent=0pt,
  columns=fullflexible,
  keepspaces=true,
  showstringspaces=false,
  frame=single,
  framesep=1.5mm,
  rulecolor=\color{HPLine},
  backgroundcolor=\color{HPPanel},
  aboveskip=2mm,
  belowskip=2mm,
  literate={→}{{$\rightarrow$}}1 {°}{{$^\circ$}}1 {θ}{{$\theta$}}1
           {–}{{--}}1 {—}{{---}}1 {’}{{'}}1 {§}{{\S}}1
}
\begin{lstlisting}[style=HPPrompt]
MEASUREMENT AGENT SYSTEM PROMPT (v45)
=====================================

# Measurement Agent Prompt (compact) v45

You are a measurement agent with local tools for physical quantities in images/videos.

## 1. Core rules
- For depth, distance, displacement, velocity, camera pose, and gaze/angle, call a measuring tool before answering. **Trajectory is the only exception** (see §6).
- Every message contains one short `Thought:` line, then either exactly one tool call or the final `<answer>...</answer>`—never both.
- Use `calculate` for all arithmetic. One multi-frame call per video is preferred.
- A tool result with a rough-click or frame-edge warning is still usable; report its number. Bad or malformed clicks must be corrected and retried, not replaced by a guess.
- The question's example lists describe format only; never copy their numbers.

## 2. Solver selection
- **Camera pose / motion:** `vggt` with `task="camera"`; convert to the question's fixed axes using §4.
- **Depth of one standing person:** `pinhole` with `want="depth"` and `[head, feet]`; answer `depth_m`.
- **Distance between two standing people:** `pinhole` with `want="distance"` and `[A head, A feet, B head, B feet]`; answer feet-to-feet `distance_m`.
- **Displacement / velocity / trajectory of one person:** `pinhole` with `want="displacement"`, using all usable frames in one call.
  - For N input frames, `points` must contain exactly N nested `[head, feet]` groups: `[[[hx0,fy0],[fx0,fy0]], [[hx1,fy1],[fx1,fy1]], ...]`. Never flatten them.
  - Require `frames_measured == N`; otherwise fix `points` and retry. An incomplete call is not a valid measurement and does not justify fallback.
  - Displacement = `path_length_m`; velocity = `speed_m_per_s` when `seconds_per_frame` is provided; trajectory = `trajectory_steps`.
- **VGGT fallback for person motion:** only after a correctly formatted all-frame pinhole call returns `status="unsupported"`, or the required head/feet are absent in all usable frames, or the person is not standing. Bad clicks and malformed/incomplete calls require a pinhole retry, not fallback.
  - Make one multi-frame `vggt task="distance"` call with `ref_head`, `ref_feet`, and `target` in every frame. Target the person's feet, or the same visible body point in every frame if feet are unavailable.
  - Convert to metres with the in-call reference scale; sum steps for displacement and divide by total time for speed using `calculate`.

## 3. Clicks, units, and video
- Click at ORIGINAL image resolution, inside the named person's labelled box. Head = crown/top of head; feet = sole ground-contact point. For person-to-person distance, use both people's feet.
- For displacement/velocity/trajectory, track the same person with matching head/feet clicks in every frame; never switch people or body points.
- Estimate each person's height from the image and pass `person_height` (one value or one per person). Average adults are about 1.7–1.8 m; adjust for children/build. Omit only for clearly average adults.
- If scale or trajectory is implausible, re-check clicks—especially feet—and retry once. Small or borderline clicks may be shifted slightly inward.
- `pinhole` distances/speeds are already metres or m/s; never rescale them.
- `vggt` outputs are model-relative. In the SAME call, compute `scale = estimated_reference_height_m / d(ref_head, ref_feet)` with `calculate`; never mix separate calls or report raw VGGT distances.
- For max/min distance in a video, measure every usable frame with the same quantity and take the max/min over usable frames only.

## 4. Camera-pose conversion (`vggt task="camera"`)
VGGT coordinates `(x,y,z)` mean `(image right, image down, initial forward)`, NOT the question's world X/Y/Z.
1. For transition i, use `D = camera_centers[i+1] - camera_centers[i]`; never use `motion.translation` for translations.
2. World `+Z` maps to camera `(0,-1,0)`. For a fixed horizontal world axis A that is θ degrees RIGHT of the initial view, use `A_camera=(sin θ,0,cos θ)`; use `-θ` for LEFT. Choose the fixed axis with the largest `abs(dot(D,A_camera))`; use `calculate` for dot products/trigonometry.
3. CameraPose-Main shortcut (initially facing world +X, +Z up, +Y left): `(X,Y,Z)=(D_z,-D_x,-D_y)`. Choose the dominant component and quantize translation distance to the question's lattice step.
4. Rotations come from `motion[i]`: `yaw_left_deg→yaw`, `pitch_up_deg→pitch`, `roll_right_deg→roll`, preserving sign and rounding to integer degrees. Emit either one translation or one rotation per atomic action.
5. Answer length must equal the number of transitions. Never derive world-axis translations directly from raw VGGT component signs.

## 5. Final-answer gate
Before `<answer>`, verify that the answer follows the question's units, precision, length, and format; that multi-frame pinhole measured all frames; and that camera translations followed §4. Except for the trajectory exception below, a successful `pinhole` or `vggt` result is mandatory. If a check fails, make the correcting tool call instead of answering.

## 6. Trajectory exception (unchanged from v41)
- Prefer `pinhole want="displacement"`, but a visual answer without a tool is valid for trajectory and overrides the general tool-only rule.
- Without a tool, track the same person and estimate turns and step lengths from the frames. Never output all-zero placeholders; use zeros only for visibly stationary motion.
- With pinhole, use the nested per-frame format and copy valid `trajectory_steps` exactly. If `|turn| > 90°` or a step is implausible, re-click once and retry.

## 7. Tool reference
Call syntax (one call on its own line):
`<<tool_name({"arg": value, ...})>>`
After the tool result arrives, make another call or give the final answer.

### calculate
`<<calculate({"expression":"1.75 / 0.403 * 0.225"})>>` → `{"result": ...}`
Parameter: `expression` (required string).

### pinhole
- Depth: `<<pinhole({"image":0,"want":"depth","points":[[hx,hy],[fx,fy]]})>>`
- Distance: `<<pinhole({"image":0,"want":"distance","points":[[Ahx,Ahy],[Afx,Afy],[Bhx,Bhy],[Bfx,Bfy]]})>>`
- Multi-frame displacement: `<<pinhole({"image":[0,1,2],"want":"displacement","points":[[[hx0,fy0],[fx0,fy0]],[[hx1,fy1],[fx1,fy1]],[[hx2,fy2],[fx2,fy2]]],"seconds_per_frame":1.0,"person_height":1.75})>>`
- `image`: zero-based frame index or list of indices; frames are attached automatically, never pass paths.
- `want`: `depth`→`depth_m`; `distance`→feet-to-feet `distance_m`; `displacement`→`path_length_m`, `trajectory_steps`, and `speed_m_per_s` when `seconds_per_frame` is supplied.
- `points`: original-resolution pixels in the formats above; one nested `[head,feet]` group per frame for displacement.

### vggt
- Camera: `<<vggt({"image":[0,1,2],"task":"camera"})>>`
- Multi-frame distance: `<<vggt({"image":[0,1,2],"task":"distance","points":[{"frame":0,"xy":[642,157],"tag":"ref_head"},{"frame":0,"xy":[643,280],"tag":"ref_feet"},{"frame":0,"xy":[450,220],"tag":"target"},{"frame":1,"xy":[641,157],"tag":"ref_head"},{"frame":1,"xy":[641,281],"tag":"ref_feet"},{"frame":1,"xy":[419,234],"tag":"target"}]})>>`
- `task="camera"` returns camera pose/motion; `task="depth"` measures selected points; `task="distance"` returns 3D coordinates and pairwise distances.
- `image`: zero-based index or list; one multi-frame call shares one coordinate system and scale. `images` is only for a single path/base64/data-URI/pixel array when no image was shown; normally omit it.
- `points`: original-resolution pixels, 1–256. Single-frame depth/distance uses `[[x,y],...]`; multi-frame uses records `{"frame":i,"xy":[x,y],"tag":"ref_head|ref_feet|target"}`. Outside-image points are invalid; re-click.
\end{lstlisting}

\clearpage
\subsection{Representative Tool-calling Examples}
\label{app:toolcalling_examples}

The following qualitative examples illustrate how tool observations enter the final answers.
They are drawn from the \emph{training} corpus and were selected separately from the
quantitative comparison as readable cases in which the tool condition improved reward: they are
\emph{not} part of the 1{,}050-question A/B of Table~\ref{tab:training_free_reward}, which is
sampled exclusively from the released evaluation bank (Appendix~\ref{app:tool_stats}). Gold
answers are never shown to the model or the tools and are used only to compute the reward offline.
Questions are reproduced verbatim apart from typographic symbols and line
breaks; ground-truth values and model answers are unchanged; tool calls and
observations are lightly abbreviated for layout.

\newcommand{\HPToolBox}[1]{%
  \HPInfoBox{\linewidth}{0mm}{HPBlue!5}{HPBlue!25}{TOOL CALL \& OBSERVATION}
  {{\ttfamily\scriptsize #1}}}
\newcommand{\HPModelBox}[2]{%
  \HPInfoBox{\linewidth}{0mm}{HPBlue!5}{HPBlue!25}{MODEL ANSWER}
  {#1\par\vspace{0.7mm}{\scriptsize\color{HPMuted}Reward: No Tool #2}}}
\newcommand{\HPExampleBottom}[4]{%
  \noindent\begin{minipage}[t]{0.70\linewidth}
    \HPModelBox{#1}{#2}
  \end{minipage}\hfill
  \begin{minipage}[t]{0.28\linewidth}
    \HPInfoBox{\linewidth}{0mm}{HPGreen!7}{HPGreen!35}{GROUND TRUTH}{\HPAnswer{#3}{#4}}
  \end{minipage}}

\newcommand{\HPCasePageBreak}{\par\vspace*{\fill}\clearpage\vspace*{\fill}}

\vspace*{\fill}
\HPSampleCard{HPBlue}{Depth}{PINHOLE \quad$\bullet$\quad 1 FRAME \quad$\bullet$\quad REWARD $0.000\rightarrow1.000$}{%
  \HPVisualLabel{INPUT FRAME}\vspace{-2.5mm}
  \begin{center}\HPImage{width=\linewidth,height=48mm,keepaspectratio}{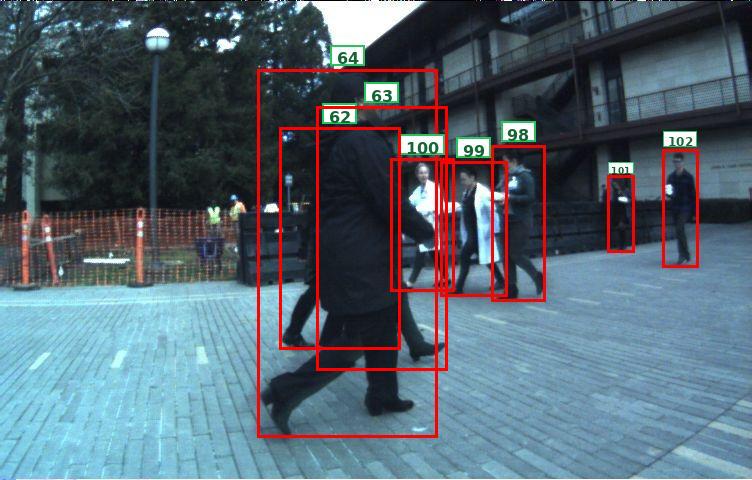}\end{center}
  \vspace{-1.5mm}
  \HPInfoBox{\linewidth}{0mm}{HPPanel}{HPLine}{QUESTION}
    {What is the distance from pedestrian 99 to the camera in this picture? (in metres, to 1 decimal place)}
  \par\vspace{1.2mm}
  \HPToolBox{pinhole(want="depth", points=[[476,160],[480,302]], person\_height=1.75)\par
\color{HPGreen}observation: depth\_m = 5.92}
  \par\vspace{1.2mm}
  \HPExampleBottom{5.9 metres}{0.000 $\rightarrow$ 1.000}{5.8}{metres}
  \HPSingleNote{Train-corpus ID 105364 \quad$\bullet$\quad Head--feet pinhole depth}
}

\HPCasePageBreak

\HPSampleCard{HPBlue}{Distance}{PINHOLE \quad$\bullet$\quad 1 FRAME \quad$\bullet$\quad REWARD $0.400\rightarrow1.000$}{%
  \HPVisualLabel{INPUT FRAME}\vspace{-2.5mm}
  \begin{center}\HPImage{width=\linewidth,height=48mm,keepaspectratio}{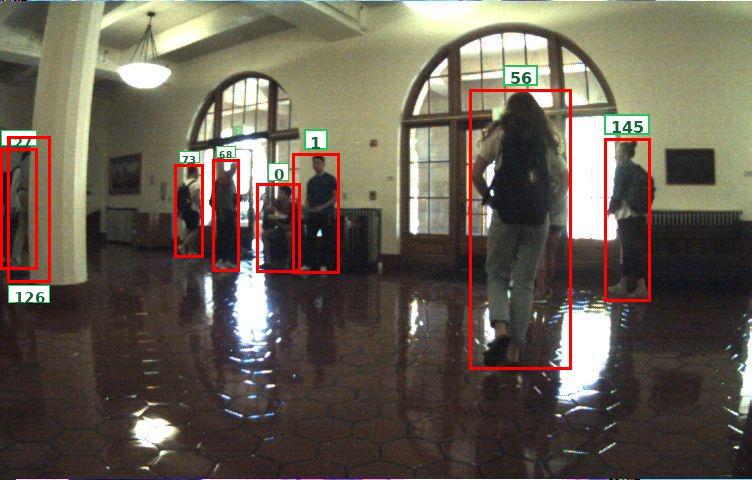}\end{center}
  \vspace{-1.5mm}
  \HPInfoBox{\linewidth}{0mm}{HPPanel}{HPLine}{QUESTION}
    {What is the distance between pedestrian 1 and pedestrian 56 in this picture? (in metres, to 1 decimal place)}
  \par\vspace{1.2mm}
  \HPToolBox{pinhole(want="distance", points=[[320,157],[320,272],[520,91],[520,367]])\par
\color{HPGreen}observation: feet-to-feet distance\_m = 4.755}
  \par\vspace{1.2mm}
  \HPExampleBottom{4.8 metres}{0.400 $\rightarrow$ 1.000}{4.9}{metres}
  \HPSingleNote{Train-corpus ID 022041 \quad$\bullet$\quad Two-person foot-point distance}
}

\HPCasePageBreak

\HPSampleCard{HPPurple}{Displacement}{PINHOLE \quad$\bullet$\quad 3 FRAMES \quad$\bullet$\quad REWARD $0.100\rightarrow1.000$}{%
  \HPVisualLabel{INPUT VIDEO FRAMES \quad\textnormal{\scriptsize (TEMPORAL ORDER)}}
  \noindent
  \HPFrame{0.323\linewidth}{34mm}{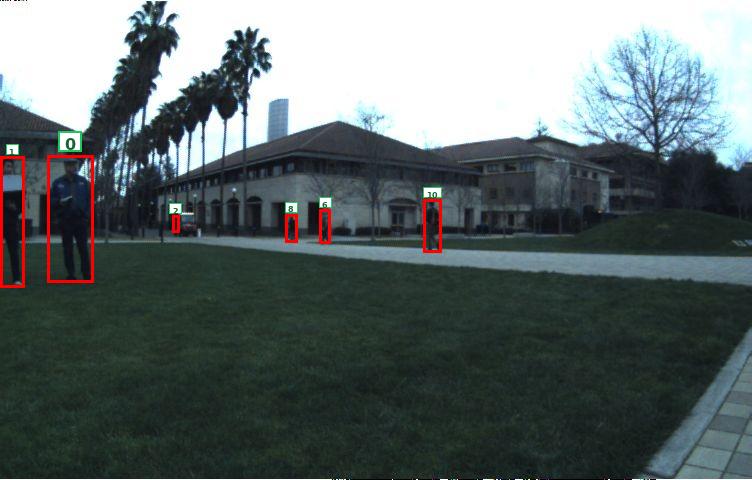}{FRAME 1}\hfill
  \HPFrame{0.323\linewidth}{34mm}{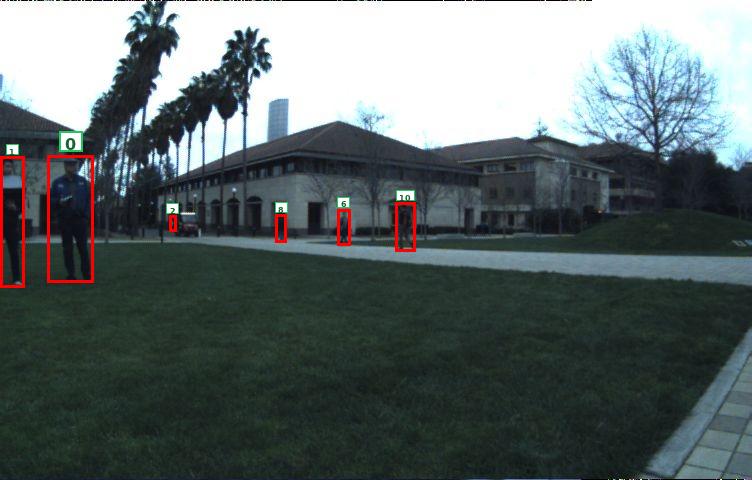}{FRAME 2}\hfill
  \HPFrame{0.323\linewidth}{34mm}{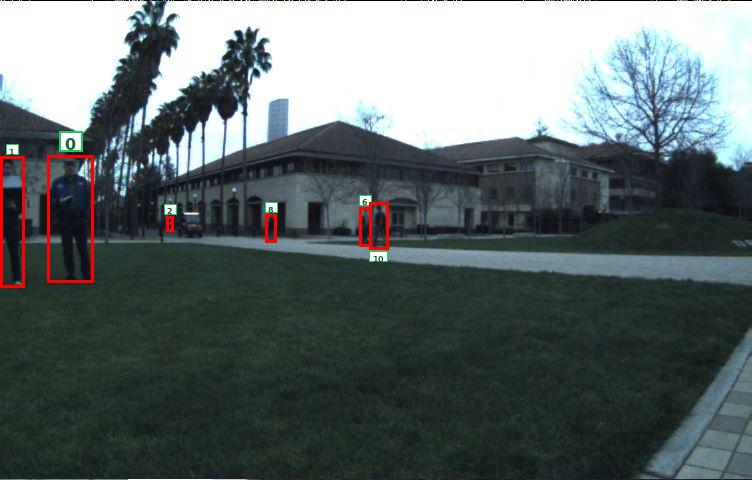}{FRAME 3}
  \par\vspace{1.2mm}
  \HPInfoBox{\linewidth}{0mm}{HPPanel}{HPLine}{QUESTION}
    {How far has pedestrian 10 moved in the video? (in metres, to 1 decimal place)}
  \par\vspace{1.2mm}
  \HPToolBox{pinhole(want="displacement", points=[[[432,201],[432,252]],
[[405,204],[405,251]], [[378,204],[378,248]]])\par
\color{HPGreen}observation: path\_length\_m = 3.312; frames\_measured = 3}
  \par\vspace{1.2mm}
  \HPExampleBottom{3.3 metres}{0.100 $\rightarrow$ 1.000}{3.3}{metres}
  \HPSingleNote{Train-corpus ID 032727 \quad$\bullet$\quad One nested head--feet pair per frame}
}

\HPCasePageBreak

\HPSampleCard{HPPurple}{Velocity}{PINHOLE \quad$\bullet$\quad 2 FRAMES \quad$\bullet$\quad REWARD $0.000\rightarrow1.000$}{%
  \HPVisualLabel{INPUT VIDEO FRAMES \quad\textnormal{\scriptsize (TEMPORAL ORDER)}}
  \noindent
  \HPFrame{0.49\linewidth}{43mm}{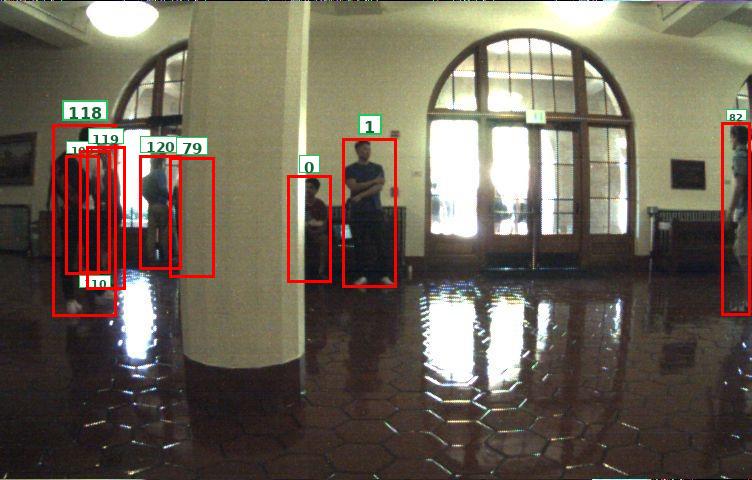}{FRAME 1}\hfill
  \HPFrame{0.49\linewidth}{43mm}{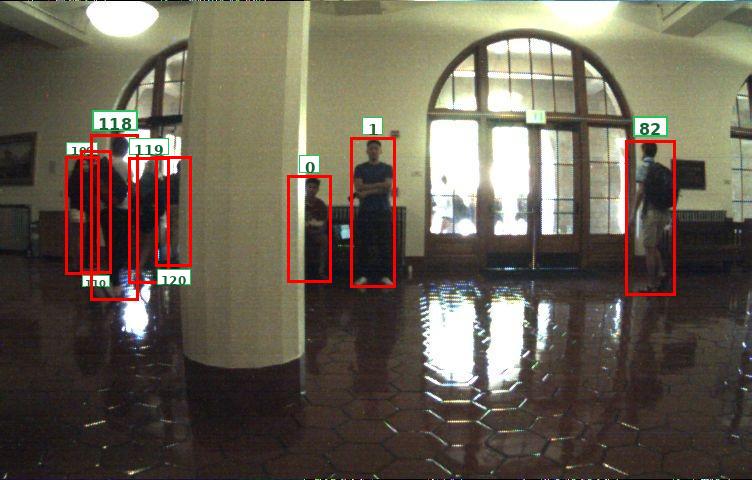}{FRAME 2}
  \par\vspace{1.2mm}
  \HPInfoBox{\linewidth}{0mm}{HPPanel}{HPLine}{QUESTION}
    {The video frames are sampled at 1.00-second intervals. What is the velocity of pedestrian 118 in the video? (in m/s, to 1 decimal place)}
  \par\vspace{1.2mm}
  \HPToolBox{pinhole(want="displacement", seconds\_per\_frame=1.0,
points=[[[83,145],[84,314]], [[111,153],[112,299]]])\par
\color{HPGreen}observation: path\_length\_m = 0.853; speed\_m\_per\_s = 0.853}
  \par\vspace{1.2mm}
  \HPExampleBottom{0.9 m/s}{0.000 $\rightarrow$ 1.000}{0.9}{m/s}
  \HPSingleNote{Train-corpus ID 009944 \quad$\bullet$\quad Metric path length with the stated time base}
}

\HPCasePageBreak

\HPSampleCard{HPPurple}{Trajectory}{PINHOLE \quad$\bullet$\quad 3 FRAMES \quad$\bullet$\quad REWARD $0.717\rightarrow0.887$}{%
  \HPVisualLabel{INPUT VIDEO FRAMES \quad\textnormal{\scriptsize (TEMPORAL ORDER)}}
  \noindent
  \HPFrame{0.323\linewidth}{34mm}{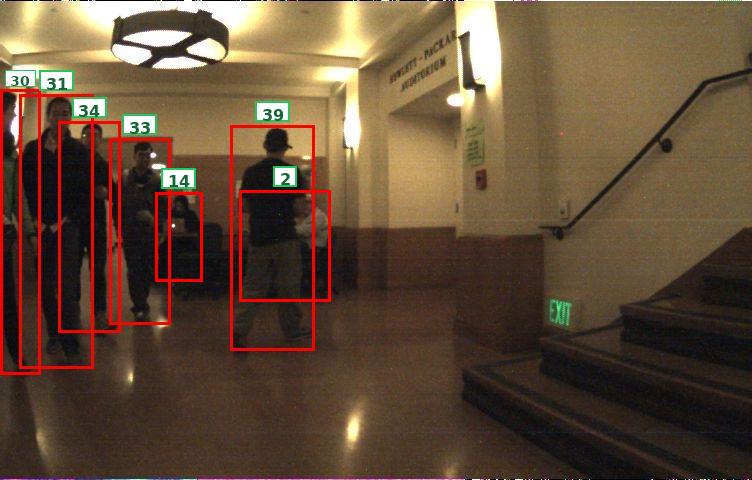}{FRAME 1}\hfill
  \HPFrame{0.323\linewidth}{34mm}{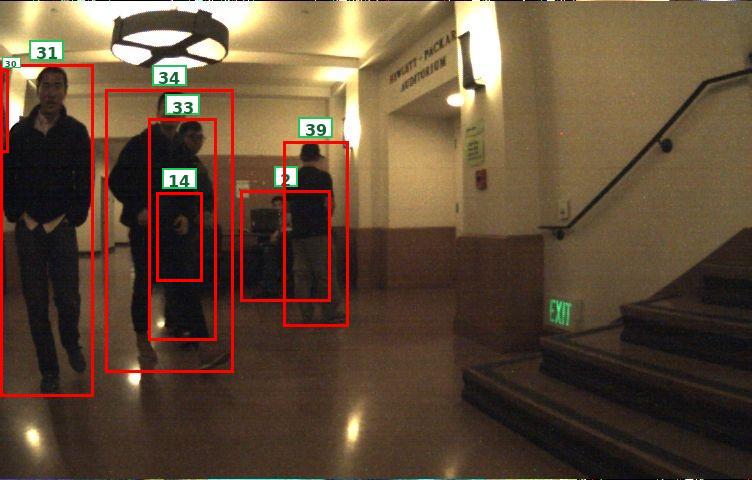}{FRAME 2}\hfill
  \HPFrame{0.323\linewidth}{34mm}{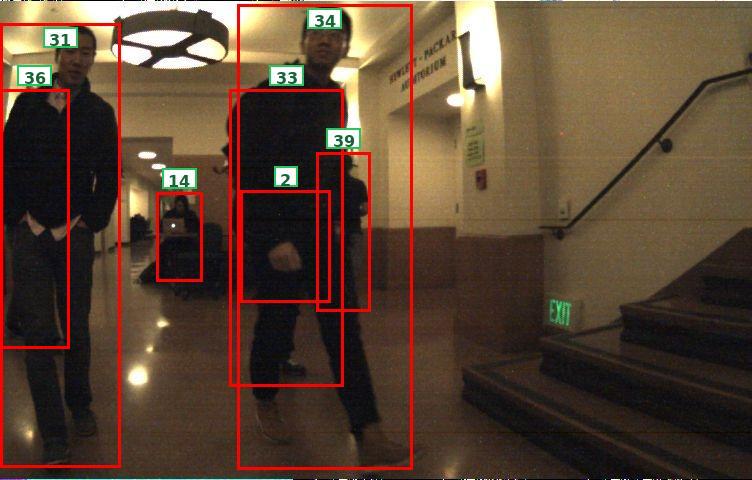}{FRAME 3}
  \par\vspace{1.2mm}
  \HPInfoBox{\linewidth}{0mm}{HPPanel}{HPLine}{QUESTION}
    {What is the trajectory of pedestrian 39 in this video? Tuple semantics: each step is (turn\_deg, distance\_m). Taking the previous step's displacement direction as the reference direction, turn\_deg is the signed turn angle from it in degrees --- absolute value within 0 to 180, positive = left, negative = right; distance\_m is the displacement between adjacent steps in metres; the first step's turn\_deg is 0.\par Answer ONLY as a Python list of two-number lists, one per step --- never a tuple. Example: [[0, 1.0], [10.0, 0.7], [-20.0, 0.6]]. Output only the list.}
  \par\vspace{1.2mm}
  \HPToolBox{pinhole(want="displacement", points=[[[274,128],[274,349]],
[[316,143],[316,325]], [[344,151],[344,310]]])\par
\color{HPGreen}observation: trajectory\_steps = [[0.0,0.898],[-3.0,0.732]]}
  \par\vspace{1.2mm}
  \HPExampleBottom{[[0.0, 0.898], [-3.0, 0.732]]}{0.717 $\rightarrow$ 0.887}{\small [[0.0, 0.8], [9.4, 0.7]]}{}
  \HPSingleNote{Train-corpus ID 015691 \quad$\bullet$\quad Heading changes and metric step lengths}
}

\HPCasePageBreak

\HPSampleCard{HPCamera}{CP-Room}{VGGT \quad$\bullet$\quad 3 FRAMES \quad$\bullet$\quad REWARD $0.300\rightarrow1.000$}{%
  \HPVisualLabel{INPUT VIDEO FRAMES \quad\textnormal{\scriptsize (TEMPORAL ORDER)}}
  \noindent
  \HPFrame{0.323\linewidth}{34mm}{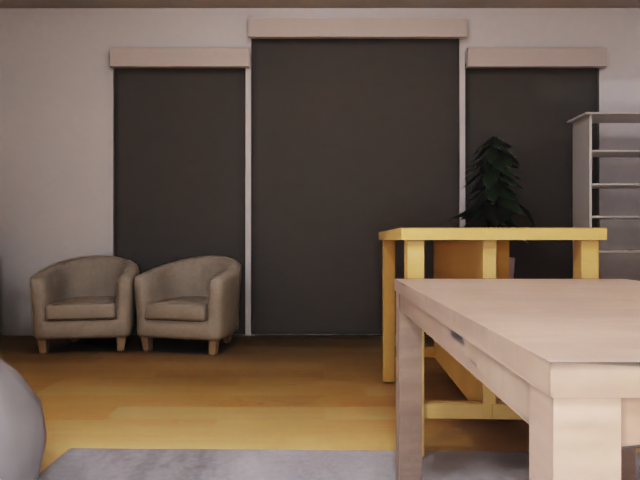}{FRAME 1}\hfill
  \HPFrame{0.323\linewidth}{34mm}{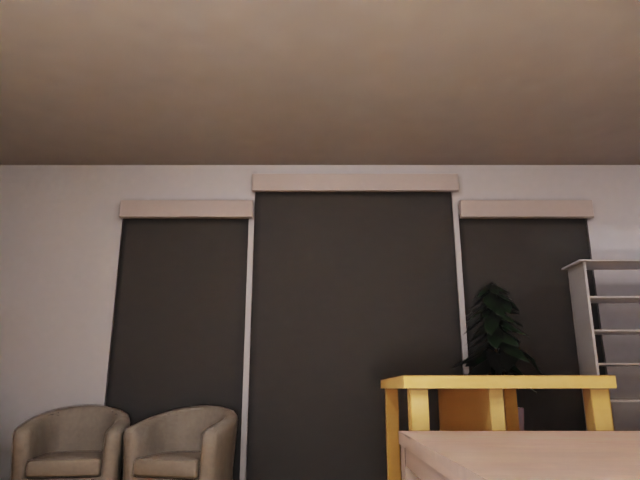}{FRAME 2}\hfill
  \HPFrame{0.323\linewidth}{34mm}{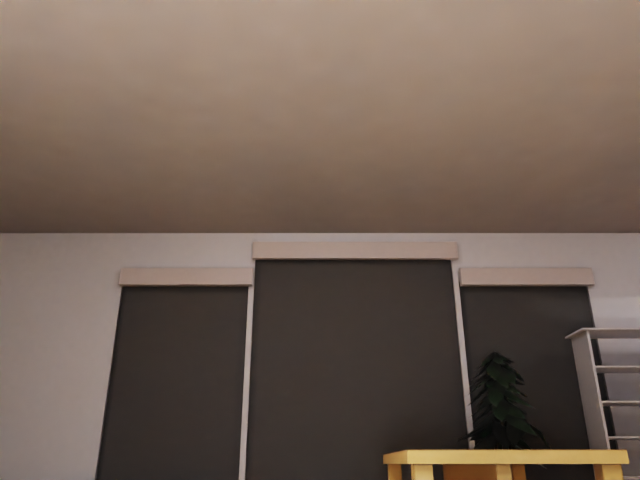}{FRAME 3}
  \par\vspace{1.2mm}
  \HPInfoBox{\linewidth}{0mm}{HPPanel}{HPLine}{QUESTION}
    {\footnotesize A free-flying camera moves through an indoor scene on a cubic free-space lattice where 1 step is 0.20 m; +X/+Y/+Z are FIXED Blender/imported-world axes -- at the start the camera faces +X with +Z up (so +Y is on its left), and every translation moves along one of these fixed axes: 'forward'/'left'/'up' name the axes, not the camera's current facing. At each step the camera performs exactly ONE atomic action: either translate along one axis by exactly 1 step (0.20 m), or rotate in place --- turn left/right (yaw), look up/down (pitch), or tilt its head (roll) --- by an integer number of degrees. Positive yaw turns left, positive pitch looks up, and positive roll tilts the head right. Each rotation is an integer number of degrees between $\pm$1$^\circ$ and $\pm$30$^\circ$, and look-up pitch never exceeds +60$^\circ$. For translations, give both the direction and the distance in metres. Predict the sequence of camera actions for every subsequent frame (2 action(s)).\par Answer ONLY as a Python list of two-item lists, one per step. Each element MUST be exactly [action, magnitude] --- a 2-item list, never a tuple. action MUST be one of these nine bare string tokens: 'yaw','pitch','roll' (a rotation, magnitude = integer degrees) or '+X','+Y','+Z','-X','-Y','-Z' (a translation, magnitude = metres). Do NOT write 'turn left','translate','rotate','yaw left' or any other words --- use ONLY the bare token. Example: [['yaw', 30], ['+Y', 0.4], ['-Y', 0.2]]. Output only the list.}
  \par\vspace{1.2mm}
  \HPToolBox{vggt(task="camera", image=[0,1,2])\par
\color{HPGreen}observation: pitch\_up\_deg = 15.0939, 7.0704;
translation distances = 0.001498, 0.001066}
  \par\vspace{1.2mm}
  \HPExampleBottom{[['pitch', 15], ['pitch', 7]]}{0.300 $\rightarrow$ 1.000}{\small [['pitch', 15], ['pitch', 7]]}{}
  \HPSingleNote{Train-corpus record \quad$\bullet$\quad Rotation-dominant transitions: VGGT's pose deltas give both the axis and the rounded rotation magnitude}
}

\HPCasePageBreak

\HPSampleCard{HPCamera}{CP-Human}{VGGT \quad$\bullet$\quad 3 FRAMES \quad$\bullet$\quad REWARD $0.500\rightarrow1.000$}{%
  \HPVisualLabel{INPUT VIDEO FRAMES \quad\textnormal{\scriptsize (TEMPORAL ORDER)}}
  \noindent
  \HPFrame{0.323\linewidth}{43mm}{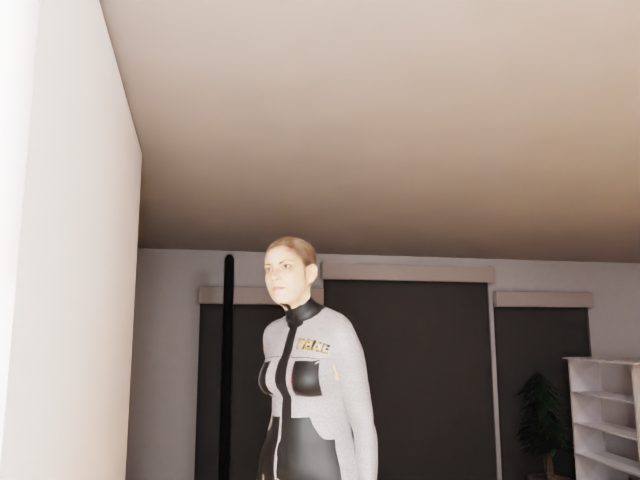}{FRAME 1}\hfill
  \HPFrame{0.323\linewidth}{43mm}{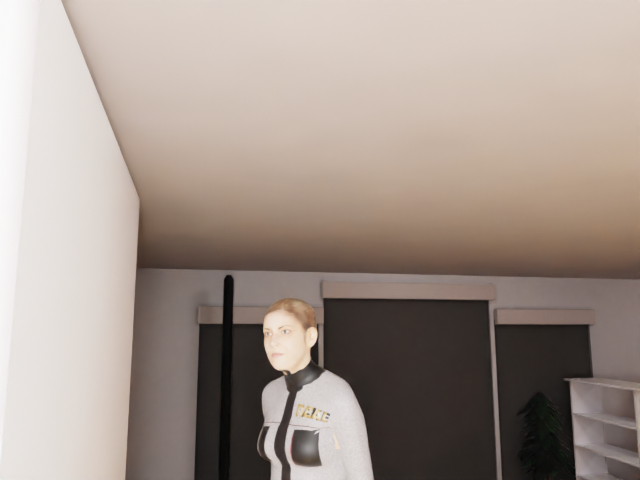}{FRAME 2}\hfill
  \HPFrame{0.323\linewidth}{43mm}{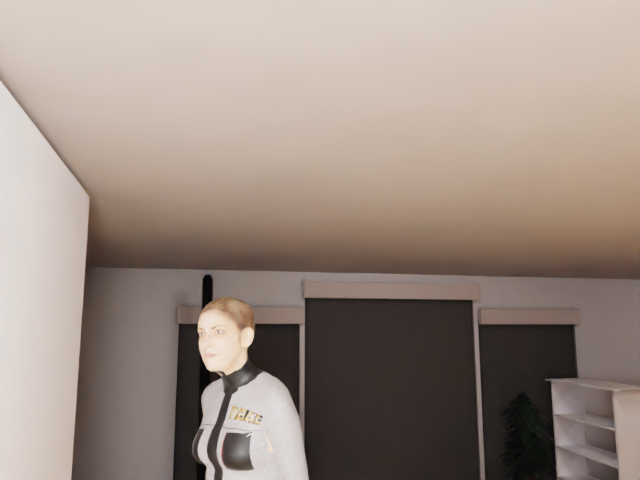}{FRAME 3}
  \par\vspace{1.2mm}
  \HPInfoBox{\linewidth}{0mm}{HPPanel}{HPLine}{QUESTION}
    {\footnotesize A free-flying camera moves through an indoor scene on a cubic free-space lattice where 1 step is 0.20 m. At each step the camera performs exactly ONE atomic action: either translate along one axis by 1 to 3 steps (e.g. +X by 2 steps = 0.40 m), or rotate in place --- turn left/right (yaw), look up/down (pitch), or tilt its head (roll) --- by an integer number of degrees. Positive yaw turns left, positive pitch looks up, and positive roll tilts the head right. Each rotation is an integer number of degrees between $\pm$1$^\circ$ and $\pm$30$^\circ$, and look-up pitch never exceeds +60$^\circ$. For translations, give both the direction and the distance in metres. At frame 1 the camera's roll is zero, so the FIXED +Z axis points straight up in the image. In the horizontal plane, +X lies 13$^\circ$ of yaw to the left of the view; -Y lies 77$^\circ$ of yaw to the right of the view; -X and +Y point behind the camera. Translation directions name these fixed axes, not the camera's current facing. The camera moves around a static person standing in the scene and keeps the person in view; the person does not move. Predict the sequence of camera actions for every subsequent frame (2 action(s)).\par Answer ONLY as a Python list of two-item lists, one per step. Each element MUST be exactly [action, magnitude] --- a 2-item list, never a tuple. action MUST be one of these nine bare string tokens: 'yaw','pitch','roll' (a rotation, magnitude = integer degrees) or '+X','+Y','+Z','-X','-Y','-Z' (a translation, magnitude = metres). Do NOT write 'turn left','translate','rotate','yaw left' or any other words --- use ONLY the bare token. Example: [['yaw', 30], ['+Y', 0.4], ['-Y', 0.2]]. Output only the list.}
  \par\vspace{1.2mm}
  \HPToolBox{vggt(task="camera", image=[0,1,2])\par
\color{HPGreen}observation: motion.translation = [0.000183,-0.069673,0.011807],
[0.067394,0.002780,0.014892]\par
\color{HPGreen}fixed-axis mapping: camera $-y\rightarrow+Z$; camera $+x\rightarrow-Y$}
  \par\vspace{1.2mm}
  \HPExampleBottom{[['+Z', 0.2], ['-Y', 0.2]]}{0.500 $\rightarrow$ 1.000}{\small [['+Z', 0.2], ['-Y', 0.2]]}{}
  \HPSingleNote{Train-corpus record \quad$\bullet$\quad Camera-to-world axis conversion: VGGT camera centres fix the \emph{axis}, while its un-scaled motion is quantised to the lattice step, so this case shows axis recovery rather than metric magnitude}
}
\par\vspace*{\fill}


\subsection{Subset, Protocol, and Tool-Calling Statistics}
\label{app:tool_stats}

The 1{,}050 questions of Table~\ref{tab:training_free_reward} are a fixed random sample
of 150 items per tool-supported sub-task from the released 200-question banks,
balanced across source scenes so that no scene dominates; both conditions
consume exactly this set. Its \emph{No Tool} column is a within-harness control, not a re-run
of Table~\ref{tab:main_results}: the agent harness stays on, with identical settings in
both conditions and only the tool list removed,
whereas Table~\ref{tab:main_results} uses the single-turn prompts of
Appendix~\ref{app:prompts}. Table~\ref{tab:training_free_reward} is therefore to be read as a
paired within-protocol A/B.

\paragraph{Tool-calling audit.}
Table~\ref{tab:tool_stats} summarizes all 1{,}069 calls of the Tool condition.
A tool is invoked on 71.5\% of questions --- lowest on \textit{Trajectory},
where the agent prompt permits a visual answer. Hard failures (malformed or
out-of-image clicks) affect 4.3\% of calls, and 26 of those 46 are followed by a
corrected same-tool retry. A further 158 calls (14.8\%, mostly \textit{Depth}
and \textit{Displacement}) return usable numbers carrying frame-edge or
rough-click warnings, which the prompt instructs the model to report rather than
re-measure.

\begin{table}[!t]
\centering
\scriptsize
\setlength{\tabcolsep}{4pt}
\caption{\textbf{Tool-calling audit of the Tool condition} (1{,}050 questions,
1{,}069 calls, 2.03 turns per question). Parenthesised counts are
pinhole/VGGT/calculator; \emph{Error} is the share of calls returning a hard
failure and \emph{Retried} the number of those failures followed by a corrected
same-tool call. Warning-bearing but usable results are excluded from
\emph{Error}.}
\label{tab:tool_stats}
\begin{tabular}{@{}lccccc@{}}
\toprule
Sub-task & Tool use (\%) & Turns & Calls/question & Error (\%) & Retried \\
\midrule
\textit{CP-Human} & 73.3 & 2.19 & 1.41 (0/141/71) & 8.5 & 16/18 \\
\textit{CP-Room}  & 70.0 & 2.03 & 1.20 (0/132/48) & 3.9 & 5/7 \\
\textit{Depth}    & 67.3 & 2.11 & 1.11 (157/7/2) & 10.2 & 2/17 \\
\textit{Distance} & 73.3 & 2.16 & 1.07 (155/3/3) & 2.5 & 3/4 \\
\textit{Displacement} & 85.3 & 2.09 & 0.91 (127/5/4) & 0.0 & 0/0 \\
\textit{Velocity} & 80.0 & 2.10 & 0.85 (120/4/3) & 0.0 & 0/0 \\
\textit{Trajectory} & 51.3 & 1.53 & 0.58 (87/0/0) & 0.0 & 0/0 \\
\midrule
All & 71.5 & 2.03 & 1.02 (646/292/131) & 4.3 & 26/46 \\
\bottomrule
\end{tabular}
\end{table}

\twocolumn[{%
\section{Training and Evaluation Split Details}
\label{app:split}
\begin{minipage}{\textwidth}
\centering
\footnotesize
\setlength{\tabcolsep}{4pt}
\renewcommand{\arraystretch}{1.15}
\captionof{table}{\textbf{Train/test split by sub-task.}
The split level is the smallest group kept disjoint between train and test.
``Train groups'' are all remaining groups for that source; full group names and counts come from the released manifest \texttt{split\_manifest.json}.}
\label{tab:split_details}
\begin{tabularx}{\linewidth}{@{}lllL{0.15\linewidth}>{\raggedright\arraybackslash}Xrr@{}}
\toprule
Sub-task & Source & Split level & Train groups & Test groups & Train rec. & Bench. \\
\midrule
Count & DanceTrack & video-disjoint & 35 remaining videos & dancetrack0002, dancetrack0027, dancetrack0033, dancetrack0049, dancetrack0086 & 1,909 & 69 \\
Count & UCSD & video-disjoint & 8 remaining videos & vidf1\_33\_003, vidf1\_33\_009 & 3,200 & 131 \\
Deform & Decaf & subject-disjoint & 6 remaining subjects (S2--S7) & S1, S8 & 267 & 100 \\
Blink & EyeBlink8 & session-disjoint & 6 remaining sessions & 3/26122013\_230103\_cam, 8/27122013\_151644\_cam & 1,349 & 200 \\\textit{Depth, Dist, Disp, Vel, Traj} & JRDB & sequence-disjoint & 26 remaining sequences & \texttt{tressider-2019-03-16\_1}\newline (5 independent camera views, 26 pedestrian tracks) & 185{,}912 & 1{,}000 \\
LS-Body & LightShadow & identity-disjoint & 196 remaining identities & 4 test identities\textsuperscript{1} & 58,800 & 200 \\
LS-Head & LightShadow & identity-disjoint & 196 remaining identities & (same 4 as LS-Body) & 39,200 & 200 \\
CP-Room & ReplicaCAD & scene-disjoint & 82 remaining scenes & Baked\_sc0\_staging\_15, Baked\_sc1\_staging\_04 & 40,038 & 200 \\
CP-Human & ReplicaCAD & scene-disjoint & 82 remaining scenes & (same 2 as CP-Room) & 41,000 & 200 \\
Gaze & VideoAttentionTarget & show-disjoint & 36 remaining shows & Jersey Shore, My Dinner with Andre, Silicon Valley, Tartuffe & 4,907 & 200 \\
\bottomrule
\end{tabularx}
\par\vspace{3pt}{\scriptsize\raggedright $^1$ The four LightShadow test identities are referred to here as \texttt{LS-test-1}\dots\texttt{LS-test-4}; their upstream identity strings are listed in the released \texttt{split\_manifest.json} and are deliberately not reproduced in the paper.\par}
\end{minipage}
\vspace{4mm}
}]
\flushbottom

\begin{figure*}[!t]
\centering
\includegraphics[page=1,width=0.95\textwidth]{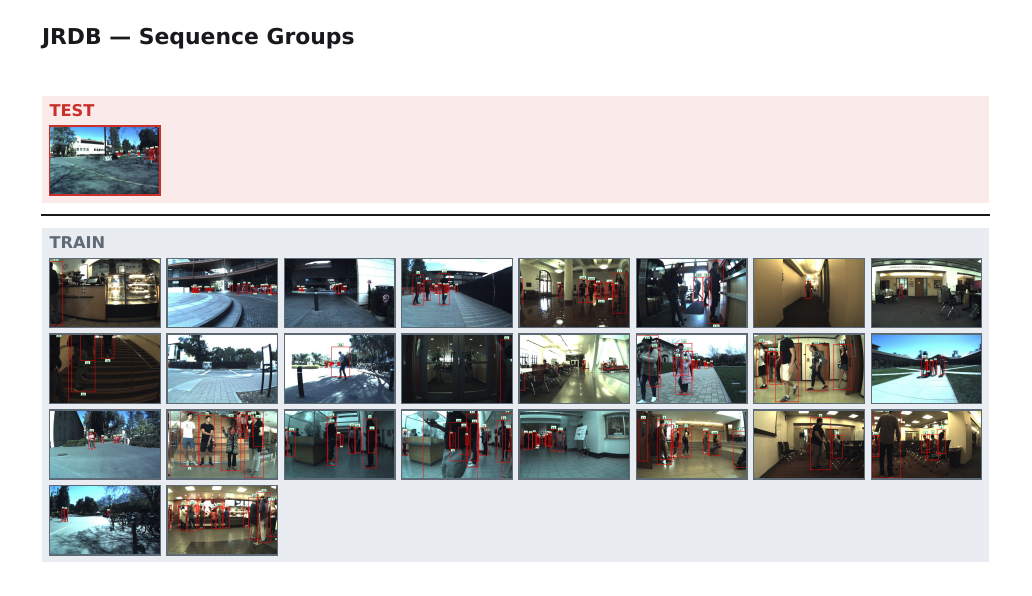}
\caption{\textbf{Split coverage: JRDB sequence groups} (1 TEST, 26 TRAIN).}
\label{fig:split_overview_1}
\vspace{3pt}
\includegraphics[page=2,width=0.95\textwidth]{assets/figures/split_groups_dense.pdf}
\caption{\textbf{Split coverage: DanceTrack video groups} (5 TEST, 35 TRAIN).}
\label{fig:split_overview_2}
\end{figure*}

HumanPhys is split at the source-\textbf{group} level, then at the \textbf{question} level.
Every family in Table~\ref{tab:split_details} is divided into disjoint groups;
questions from the test groups total 11{,}620: 2{,}500 are sampled into the evaluation benchmark and the remaining 9{,}120 form the unused holdout. \emph{All} 11{,}620 are excluded from the 376{,}582-pair training corpus, and the unused holdout is shipped alongside the benchmark for future test-set extension.
Because sampling is question-level, the same frame can serve several questions, so the benchmark and the unused holdout may share scenes; they share no question, and neither shares any record with the 376{,}582-pair training corpus. Both exclusion properties were verified programmatically before release.
Figures~\ref{fig:split_overview_1}--\ref{fig:split_overview_8} cover \textbf{all 417 unique split groups}, read left to right. Each cell shows one frame drawn under a fixed per-group seed from that split unit; held-out groups appear in a red TEST band and the remaining groups in a gray TRAIN band, separated by a black rule. ReplicaCAD and LightShadow are each drawn once per render view over the identical disjoint group set (701 cells in total).

\begin{figure*}[!t]
\centering
\includegraphics[page=3,width=\textwidth]{assets/figures/split_groups_dense.pdf}
\caption{\textbf{Split coverage: UCSD (native low resolution) and EyeBlink8 groups} (UCSD 2 TEST / 8 TRAIN; EyeBlink8 2 TEST / 6 TRAIN).}
\label{fig:split_overview_3}
\end{figure*}

\begin{figure*}[!t]
\centering
\includegraphics[page=4,width=\textwidth]{assets/figures/split_groups_dense.pdf}
\caption{\textbf{Split coverage: DECAF subject groups and VideoAttentionTarget show groups} (2 TEST / 6 TRAIN; 4 TEST / 36 TRAIN). For DECAF we show the original face--hand video frame rather than the rendered QA probe.}
\label{fig:split_overview_4}
\end{figure*}

\begin{figure*}[!t]
\centering
\includegraphics[page=5,width=\textwidth]{assets/figures/split_groups_dense.pdf}
\caption{\textbf{Split coverage: ReplicaCAD CameraPose-Main scene groups} (2 TEST, 82 TRAIN); each cell shows one deterministically sampled camera view of that scene-level split ID.}
\label{fig:split_overview_5}
\end{figure*}

\begin{figure*}[!t]
\centering
\includegraphics[page=6,width=\textwidth]{assets/figures/split_groups_dense.pdf}
\caption{\textbf{Split coverage: ReplicaCAD CameraPose-Human scene groups} (2 TEST, 82 TRAIN); the split is scene-level, as for \textit{CP-Room}.}
\label{fig:split_overview_6}
\end{figure*}

\begin{figure*}[!t]
\centering
\includegraphics[page=7,width=\textwidth]{assets/figures/split_groups_dense.pdf}
\caption{\textbf{Split coverage: LightShadow identity groups, body view (\textit{LS-Body})} (4 TEST, 196 TRAIN), drawn from the \texttt{LightShadow-Main} body-view render that the sub-task scores; each cell shows one frame from that identity's eight wall cameras.}
\label{fig:split_overview_7}
\end{figure*}

\begin{figure*}[!t]
\centering
\includegraphics[page=8,width=\textwidth]{assets/figures/split_groups_dense.pdf}
\caption{\textbf{Split coverage: LightShadow identity groups, face view (\textit{LS-Head})} (the same 4 TEST / 196 TRAIN identities), drawn from the \texttt{LightShadow-Selfie} render that the sub-task scores.}
\label{fig:split_overview_8}
\end{figure*}

\noindent \textbf{Count group pools.} Count is a composite of DanceTrack videos and UCSD videos; the two families are split independently (both video-disjoint). The benchmark quota of 200 Count questions is filled from these independent video pools.

\noindent \textbf{Sequence-disjoint JRDB.} All five JRDB sub-tasks (Depth, Distance, Displacement, Velocity, Trajectory) share the test sequence \texttt{tressider-2019-03-16\_1} \textbf{(5 independent camera views, 26 pedestrian tracks)}, while the 26 remaining sequences (185{,}912 QA pairs) stay disjoint from training; the strict static-camera filter (\S\ref{sec:construction}) admits only this densely tracked sequence. Within the held-out group, the five separate 200-question quotas query different frames, pedestrians and spatial criteria.

\begin{figure*}[!t]
\centering
\includegraphics[width=\textwidth]{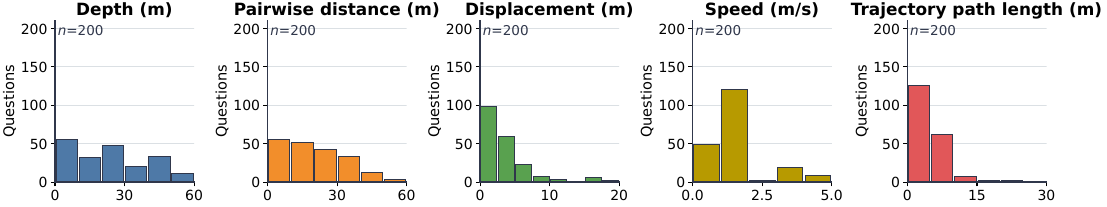}
\caption{\textbf{Histograms for the JRDB quantities in Table~\ref{tab:tressider_diversity}.} All five panels are derived programmatically from the released benchmark file and contain 200 questions each. \textit{Depth}, \textit{Distance}, \textit{Displacement}, and \textit{Velocity} use gold answers; \textit{Trajectory} uses the sum of distances in its gold tuple sequence. Each panel has its own bin width and $x$-axis scale.}
\label{fig:tressider_hist}
\end{figure*}

An audit of the released benchmark file quantifies that within-group diversity. The 1{,}000 JRDB benchmark questions (200 per sub-task) name 25 distinct pedestrian IDs, while the accompanying 3D-box metadata contains 26 tracked IDs; across the five camera views they use 160 distinct (camera, frame) images. Gold values span 2.1--59.8\,m for depth, 0.7--54.9\,m for pairwise distance, 0.5--18.7\,m for displacement, 0.5--4.9\,m/s for velocity and 1.2--25.2\,m for trajectory path length, and 179 of the 200 trajectory tuple sequences are unique and cover 2--7 steps (Table~\ref{tab:tressider_diversity}; histograms in Figure~\ref{fig:tressider_hist}).

\begin{table*}[!t]
\centering
\footnotesize
\setlength{\tabcolsep}{4pt}
\caption{\textbf{Diversity audit of the held-out JRDB sequence \texttt{tressider-2019-03-16\_1}.}
Each row summarizes the 200 benchmark questions for that JRDB sub-task. \textbf{Queried IDs} counts pedestrian IDs named explicitly in the questions; \textbf{Frames} counts unique frame indices; and \textbf{Images} counts distinct (camera, frame) paths. For \textit{Depth}, \textit{Distance}, \textit{Displacement}, and \textit{Velocity}, the reported quantities are the answer values. For \textit{Trajectory}, they are path lengths derived from the gold tuple sequence; in the full audit these 200 trajectories contain 2--7 steps and absolute turn angles up to $121.4^\circ$.}
\label{tab:tressider_diversity}
\resizebox{\textwidth}{!}{%
\begin{tabular}{@{}lrrrlrrrrr@{}}
\toprule
Sub-task & Queried IDs & Frames & Images & Audited quantity & Min. & Q1 & Median & Q3 & Max. \\
\midrule
\textit{Depth} & 25 & 34 & 140 & camera depth (m) & 2.1 & 9.5 & 23.5 & 38.0 & 59.8 \\
\textit{Distance} & 24 & 34 & 103 & pairwise distance (m) & 0.7 & 9.4 & 17.2 & 29.6 & 54.9 \\
\textit{Displacement} & 22 & 34 & 130 & path length (m) & 0.5 & 1.1 & 2.6 & 4.7 & 18.7 \\
\textit{Velocity} & 21 & 33 & 135 & speed (m/s) & 0.5 & 1.0 & 1.1 & 1.3 & 4.9 \\
\textit{Trajectory} & 21 & 34 & 115 & tuple path length (m) & 1.2 & 2.3 & 3.7 & 5.8 & 25.2 \\
\bottomrule
\end{tabular}%
}
\end{table*}


\section{Full SFT Results by HumanPhys Sub-task}
\label{app:sft_subtasks}

Table~\ref{tab:sft_subtasks} reports the full 13-task HumanPhys profile of the five SFT data mixtures used in Table~\ref{tab:ablation}; scores are multiplied by 100. SFT~\ding{172} is HumanPhys only; \ding{173} adds SpaceR-151K; \ding{174} adds SpatialCorpus-331K; \ding{175} uses all three; and \ding{176} is the no-HumanPhys control. \textbf{Overall} is as in Table~\ref{tab:main_results}. For orientation against the \emph{Human} row of
Table~\ref{tab:main_results} (52.48 overall), the four HumanPhys-containing mixtures exceed it on eleven of the
thirteen sub-tasks --- the exceptions are \textit{Count} (human 79.47 against
a best mixture value of 78.40) and \textit{Deform} (human 36.00 against 31.00 for the best mixture, and 32.00 for the no-HumanPhys control) ---
and exceed it on \textbf{Overall} by 21--23 points. Humans nevertheless remain above every
zero-shot, API-served and metric-guided-RL model on \textbf{Overall}, whose
best value is 50.31.

\begin{table*}[!t]
\centering
\scriptsize
\setlength{\tabcolsep}{2.5pt}
\caption{\textbf{SFT ablation on all HumanPhys sub-tasks.} Rows correspond to the SFT data mixtures in Table~\ref{tab:ablation}; best score per column among these five runs is in \textbf{bold}. \textit{Dist} = Distance, \textit{Disp} = Displacement, \textit{Traj} = Trajectory, \textit{LS} = Light-Shadow, and \textit{CP} = Camera-Pose. \textbf{Overall} and the zero-counting of unparseable outputs are as in Table~\ref{tab:main_results} (per-sub-task unparseable rates in Table~\ref{tab:unparseable}).}
\label{tab:sft_subtasks}
\begin{tabular*}{\linewidth}{@{}l@{\extracolsep{\fill}}cccccccccccccc@{}}
\toprule
Model & Depth & Dist & Count & Gaze & Disp & Vel & Traj & Blink & Deform & LS-Head & LS-Body & CP-Room & CP-Human & Overall \\
\midrule
SFT \ding{172} & \textbf{80.40} & 56.65 & 72.95 & 73.70 & 37.80 & 42.30 & \textbf{70.28} & 98.60 & 30.00 & 90.08 & 95.57 & 97.18 & 96.29 & 74.14 \\
SFT \ding{173} & 79.35 & 57.35 & \textbf{78.40} & 75.90 & 41.45 & 44.35 & 69.57 & \textbf{99.15} & 30.00 & \textbf{91.59} & 95.81 & 97.72 & \textbf{96.98} & \textbf{75.41} \\
SFT \ding{174} & 79.45 & 58.05 & 76.15 & 76.20 & 41.50 & 43.55 & 68.43 & 98.00 & 29.00 & 90.13 & \textbf{96.02} & \textbf{98.08} & 80.20 & 73.62 \\
SFT \ding{175} & 80.25 & \textbf{58.20} & 77.15 & \textbf{76.60} & \textbf{43.00} & \textbf{44.70} & 68.97 & 98.70 & 31.00 & 89.44 & 95.41 & 96.67 & 83.05 & 74.21 \\
SFT \ding{176} & 39.35 & 23.45 & 57.50 & 2.10 & 16.65 & 19.65 & 67.16 & 39.90 & \textbf{32.00} & 33.69 & 56.33 & 0.00 & 0.00 & 29.74 \\
\bottomrule
\end{tabular*}
\end{table*}

\section{Transfer Gains on General Spatial Intelligence Benchmarks}
\label{app:transfer}

Table~\ref{tab:transfer_gains} lists the MMSI-Bench~\citep{mmsibench} and SPARBench~\citep{sparbench} sub-tasks on
which the HumanPhys-only mixture (SFT~\ding{172}, Table~\ref{tab:sft_subtasks})
improves over the Qwen3.5-4B base model, under the same evaluation protocol as
Table~\ref{tab:ablation}. Only positive deltas are listed, so the
benchmark-level aggregates of Table~\ref{tab:ablation} move by less than these
individual sub-tasks.

\begin{table*}[!t]
\centering
\scriptsize
\setlength{\abovecaptionskip}{4pt}
\setlength{\belowcaptionskip}{3pt}
\renewcommand{\arraystretch}{0.92}
\setlength{\tabcolsep}{2pt}
\caption{\textbf{Transfer gains of HumanPhys-only SFT.} Base $=$ Qwen3.5-4B; SFT $=$ SFT~\ding{172} (Table~\ref{tab:sft_subtasks}); $\Delta$ $=$ SFT $-$ Base in points. Only positive deltas are listed. Gains concentrate on camera-, viewpoint- and ego-motion reasoning.}
\label{tab:transfer_gains}
\begin{tabular}{@{}ll>{\centering\arraybackslash}p{4.9cm}ccc@{}}
\toprule
Benchmark & Sub-task & What it asks & Base & SFT & $\Delta$ \\
\midrule
\multirow{5}{*}{MMSI-Bench}
 & Motion (Cam.) & Identify camera motion type & 24.32 & 44.59 & +20.27 \\
 & MSR & Multi-step reasoning: a sequence of reasoning steps (e.g.\ navigation) & 19.70 & 33.84 & +14.14 \\
 & Positional (Cam.--Cam.) & Relative position of two cameras & 20.43 & 27.96 & +7.53 \\
 & Positional (Obj.--Obj.) & Relative position of two objects & 19.15 & 25.53 & +6.38 \\
 & Positional (Reg.--Reg.) & Relative position of two regions & 18.52 & 23.46 & +4.94 \\
\midrule
\multirow{6}{*}{SPARBench}
 & distance\_infer\_center (oo, mv) & Object--object: which of two objects is farther from a third (MC), multi-view & 37.50 & 61.90 & +24.40 \\
 & view\_change\_infer & How to move the camera from view~1 to view~2 & 8.43 & 24.10 & +15.67 \\
 & camera\_motion\_infer & Camera pose (MC): where the second view's camera falls in image~1 & 15.75 & 29.75 & +14.00 \\
 & spatial\_imagination (oc, mv) & Novel-view layout, object--camera & 29.07 & 34.59 & +5.52 \\
 & spatial\_imagination (oo, mv) & Novel-view layout, object--object & 22.69 & 28.01 & +5.32 \\
 & obj\_spatial\_relation (oc, mv) & Object--camera spatial relation & 50.50 & 55.00 & +4.50 \\
\bottomrule
\end{tabular}
\end{table*}

\section{\texorpdfstring{Comparison with the Closest Benchmarks}{Comparison with the Closest Benchmarks}}
\label{app:closest}

This appendix makes the two comparisons promised in \S\ref{sec:related_human} explicit, so
that our claims about prior work can be checked axis by axis rather than asserted.
Table~\ref{tab:closest} contrasts HumanPhys with HumanMoveVQA~\citep{humanmovevqa}, to our
knowledge the closest existing human-centric benchmark, since it asks quantitative questions about
global body-root motion from exocentric video, and with SpatialScore~\citep{spatialscore}, which its
authors describe as the most comprehensive spatial evaluation to date and which ships the
SpatialCorpus training resource. Table~\ref{tab:hmvqa-mapping} decomposes the first
comparison by task category and states precisely how much of HumanPhys overlaps with it.

\begingroup
\footnotesize
\setlength{\tabcolsep}{4pt}
\renewcommand{\arraystretch}{1.12}
\begin{table*}[!t]
\centering
\caption{\textbf{HumanPhys vs.\ the two closest benchmarks.} SpatialScore is self-described as the most
comprehensive general spatial evaluation, with an accompanying training corpus. Entries are taken
from each work's published description unless a cell says otherwise; ``---'' means the axis does
not apply.}
\label{tab:closest}
\begin{tabular}{@{}L{0.14\linewidth}L{0.29\linewidth}L{0.27\linewidth}L{0.22\linewidth}@{}}
\toprule
 & \textbf{HumanPhys (ours)} & \textbf{HumanMoveVQA} & \textbf{SpatialScore} \\
\midrule
Measurement target & the human body as rigid/non-rigid target, photometric probe and observer reference & root translation and orientation of one target person & object and scene geometry, including the camera \\
Task inventory & 13 sub-tasks in 6 dimensions & seven MC categories over three cognitive axes & 30 tasks in 10 categories \\
Answer units & m, m/s, $^{\circ}$, event counts, lattice coords, \texttt{(turn\_deg, distance\_m)} tuples & ordinal magnitude bins (short/moderate/long, slow/fast), integer event counts, direction labels & 4{,}149 categorical (MC/judgement), 876 numeric open-ended \\
Format and scoring & open-ended; MRA/angular scoring + circular-difference sequence score; MC only for \textit{Deform} & all multiple choice; chance-normalised accuracy & judgement, MC and open-ended accuracy \\
Gold provenance & sensor calibration, 3D-box metadata, renderer state; no estimator/annotator \emph{on the benchmark frames} (residual upstream noise disclosed in App.~\ref{app:taskdetails}) & PromptHMR~\citep{prompthmr}-recovered SMPL-X tracks $\to$ MotionScript~\citep{motionscript} spatial codes, two-tier filtering (manual at video level, rule-based at event level) & 1{,}091 samples newly built from 3D annotations, ${\sim}$3.9K curated from 23 datasets; all manually verified \\
Coordinate convention & metric world frame; static-camera filter; foot-point distance; heading anchored at $0^{\circ}$, left positive & first-frame anchored world frame; root translation and orientation, values normalised to the first frame & per-task source conventions \\
\midrule
Viewpoint and people & exocentric + face-level; multi-person (\textit{Count}, \textit{Distance}); probes with/without person & exocentric; one \emph{target} person per clip (EgoBody clips hold two people) & single images (2{,}493), multi-image sets (1{,}339), videos (1{,}193) \\
Visual source & 5 real video datasets and 3 rendered collections & real human-motion video data (EMDB, RICH, EgoBody) & real-world, simulated and AIGC, plus 3D-annotation repurposing \\
Evaluation scale & 2{,}500 evaluation questions, equal quota per sub-task (group-level split, question-level quotas) & 10{,}203 test questions from 97 videos & ${\sim}$5K manually verified samples over 30 tasks \\
Training resource & 376{,}582 QA pairs, same pipeline, no question shared with eval & 89{,}818 QA pairs from the same generator & SpatialCorpus (331K); SpatialAgent (training-free) \\
Post-training protocol & SFT of Qwen3.5-4B on five mixtures (four with HumanPhys) & LoRA SFT of Qwen3-VL-8B on its own data & SFT of Qwen3-VL-4B/8B on SpatialCorpus (1 epoch) \\
Generalization analysis & transfer to SPAR/MMSI sub-tasks; specialization trade-off on 9 external benchmarks; mixture repair & cross-dataset and per-category transfer, plus out-of-domain evaluation on ActionArt~\citep{actionart} & gains on own benchmark after SFT/agent \\
Human-free conditions & \textit{CP-Room} as human-free scene-level baseline & none; every clip has the target person & all conditions \\
Overlap with HumanPhys & --- & \textit{Trajectory}, \textit{Displacement}, \textit{Velocity} (3 of 13 sub-tasks) & camera pose/motion, counting, depth (object/scene referent) \\
\bottomrule
\end{tabular}
\end{table*}
\endgroup

\twocolumn[{\begin{minipage}{\textwidth}
\centering
\small
\setlength{\tabcolsep}{5pt}
\renewcommand{\arraystretch}{1.15}
\captionof{table}{\textbf{Category-by-category overlap with HumanMoveVQA.} Ten of our thirteen
sub-tasks have no counterpart there, and the three that do ask for open-ended metric sequences
rather than categorical multiple choice.}
\label{tab:hmvqa-mapping}
\begin{tabular}{@{}L{0.16\linewidth}L{0.22\linewidth}L{0.54\linewidth}@{}}
\toprule
Their category & Nearest HumanPhys sub-task & What each asks for \\
\midrule
Existence & (none) & they assert or deny one motion event (2-way MC); we count events (\textit{Blink}) or ask for the magnitude itself \\
Comparative & (none) & they compare frequency, magnitude or speed along one axis (3-way MC with a tie option); we ask for the metric value and score it by MRA \\
Dominant & \textit{Trajectory} (weakly) & they ask the primary direction or rotation along one axis (3-way MC with a tie option); we ask for the whole ordered turn/step sequence \\
Numerical & \textit{Displacement} (partial); \textit{Count}: none & they count discrete motion events (displacements or rotations) or pick a total displacement from unitless bins (4-way MC), never people; we count humans under an explicit metric criterion and ask displacement in metres \\
Ordering & \textit{Trajectory} (order) & they identify the event following an anchor; we score every step of the tuple sequence by accumulated heading \\
Temporal & \textit{Velocity} & they pick the earliest event or a slow/fast category; we ask m/s against a stated sampling interval \\
Trajectory affordance & \textit{Displacement}, \textit{Trajectory} & they pick net displacement, path length or path shape from 4 options; we score per-step \texttt{(turn\_\allowbreak deg,\ \allowbreak distance\_\allowbreak m)} tuples continuously \\
\bottomrule
\end{tabular}
\end{minipage}
\vspace{4mm}}]

Two conclusions follow. First, the overlap with HumanMoveVQA is confined to three
of our thirteen sub-tasks, and even there the answer format differs in kind: an
open-ended metric sequence scored step by step rather than a categorical choice
among two to four options. Second, our differentiation from SpatialScore is about the
referent rather than breadth: we claim neither the paradigm nor the breadth, only
that none of its task categories measures a property of a human body and that
human-physical perception is not obtained from that supervision
(\S\ref{sec:findings}).

\section{Metric-Guided RL Details}
\label{app:rl_details}

This appendix documents the GRPO/RLVR run of \S\ref{sec:metric_rl}: the optimisation
configuration, the reward design, the sampling policy, and the complete
13-sub-task profile of the resulting checkpoint (Fig.~\ref{fig:rl_reward} shows
only the overall score and the three largest gains).

\paragraph{Optimisation configuration.}
We train from the un-fine-tuned Qwen3.5-4B checkpoint with GRPO~\citep{grpo} (group size
$G=8$ rollouts per prompt) on 8 nodes $\times$ 16 Ascend HPU-910C (128
accelerators), using a colocated hybrid engine: the actor is updated with
data-parallel degree 128 while rollouts are served by vLLM (dp 8, tp 16).
Each step consumes 128 prompts. We train for around 200 steps and keep
periodic checkpoints at steps 60, 120 and 180; we report the last one
(step 180). The optimiser is Adam
($\beta_1{=}0.9$, $\beta_2{=}0.999$, $\epsilon{=}10^{-8}$) at a constant
learning rate of \mbox{$1\times10^{-6}$} with gradient clipping at 1.0 and no weight
decay. The policy ratio uses token-level importance sampling clipped to
$[1{-}0.2,\,1{+}0.28]$ with a dual-clip constant of 2.0; no KL regulariser is
applied. Rollouts are
sampled at temperature 1.0 and top-$p$ 1.0 with thinking disabled; prompts are
capped at 16{,}384 tokens within a 17{,}408-token sequence budget. As in the
SFT runs, the ViT and the multimodal projector stay frozen and only
the language model is updated, in bfloat16.

\paragraph{Reward design.}
The reward is exactly the evaluation metric of \S\ref{sec:protocol}: each
sub-task's deterministic scorer (MRA over
$C{=}\{0.50,\dots,0.95\}$ for numeric answers, circular-error thresholds
$\{5^{\circ},\dots,45^{\circ}\}$ for angles, lattice-coordinate score for
\textit{LS}, step-sequence score for \textit{Trajectory}, exact match for
\textit{Deform}) maps a completion to $[0,1]$. There is no reward model and no
human preference label. \emph{Format handling matches evaluation exactly}: a
completion without a parseable \texttt{<answer>}$\dots$\texttt{</answer>}
block (including truncated outputs) receives reward $0$, and a block whose
numbers cannot be aligned one-to-one with the gold values also receives $0$;
we add no separate format-bonus term. Instead, format following is induced by
appending the suffix ``Keep your reasoning brief. Put your final answer inside
\texttt{<answer></answer>}.'' to every training prompt.

\paragraph{Sampling policy.}
Prompts are drawn uniformly from the same
376{,}582-pair HumanPhys SFT corpus used by SFT~\ding{172}; prompts exceeding
the context budget are filtered out. No long-tail re-balancing is applied, so
scarce sub-tasks such as \textit{Deform} (267 training pairs) contribute very
few gradient updates per step --- consistent with its modest RL gain below.

\begin{table*}[!t]
\centering
\scriptsize
\setlength{\tabcolsep}{2.5pt}
\caption{\textbf{Metric-guided RL on all HumanPhys sub-tasks.} The RL
checkpoint is the last saved checkpoint (step 180) of the GRPO/RLVR run of
\S\ref{sec:metric_rl}; the base row repeats the Qwen3.5-4B row of
Table~\ref{tab:main_results}. Best per column in \textbf{bold}. The two
regressions (\textit{LS-Body}, \textit{CP-Human}) are discussed in the text.}
\label{tab:rl_subtasks}
\begin{tabular*}{\linewidth}{@{}l@{\extracolsep{\fill}}cccccccccccccc@{}}
\toprule
Model & Depth & Dist & Count & Gaze & Disp & Vel & Traj & Blink & Deform & LS-Head & LS-Body & CP-Room & CP-Human & Overall \\
\midrule
Qwen3.5-4B (base) & 26.45 & 24.15 & 58.00 & 12.30 & 2.70 & 11.95 & 65.10 & 32.55 & 25.00 & 23.44 & \textbf{51.29} & 8.43 & \textbf{4.64} & 26.68 \\
Metric-guided RL & \textbf{47.65} & \textbf{36.40} & \textbf{80.25} & \textbf{40.30} & \textbf{22.00} & \textbf{18.50} & \textbf{69.81} & \textbf{47.65} & \textbf{28.00} & \textbf{64.57} & 40.97 & \textbf{9.67} & 3.00 & \textbf{39.58} \\
\bottomrule
\end{tabular*}
\end{table*}

\paragraph{Where the metric helps --- and where it hurts.}
Table~\ref{tab:rl_subtasks} decomposes the $+12.90$ overall gain. The largest
improvements are on \textit{LS-Head} ($+41.13$), \textit{Gaze} ($+28.00$),
\textit{Count} ($+22.25$), \textit{Depth} ($+21.20$), \textit{Disp}
($+19.30$) and \textit{Blink} ($+15.10$): exactly the sub-tasks whose gold
values are read from renderer state or stable metadata, so the deterministic
reward is a clean learning signal. Two regressions are visible and we do not
hide them. \textit{LS-Body} drops by $10.32$ while \textit{LS-Head} rises: the
two share identities but differ in camera placement, and the policy
specialised to the selfie-view coordinate convention at the expense of the
body-view one --- a view-level over-optimisation that a single aggregate
score would conceal. \textit{CP-Human} ($-1.64$) and, to a lesser extent,
\textit{CP-Room} ($+1.24$) remain near floor: consistent with the axis-level
confusion analysis of \S\ref{sec:diagnosis}, ego-motion recovery needs more
than a scalar reward on 200 steps. \textit{Deform} gains only $+3.00$,
matching its 267-pair presence in the sampling pool. These observations
motivate the SFT+RL combination and per-view reward balancing as future work.

\section{Agentic Tool-Use Policy}
\label{app:agentic_rl}

\S\ref{sec:training_free} shows that a frontier model can use geometric tools at inference time.
Here we ask whether the same behaviour can be \emph{distilled} into a small open model: learning \emph{when} to call a tool, \emph{which} tool to call, and how to consume its output, from teacher trajectories and metric-based rewards alone. All rows use the same strict composite scorer as Table~\ref{tab:main_results} (NA$\to$MRA, sequences$\to$tuple scorers, Deform$\to$ACC; unparseable outputs score $0$) over the full 2{,}500-item benchmark.

\paragraph{Stage 1: teacher trajectory distillation.}
We roll out GPT-6-Sol with the measurement-agent prompt of App.~\ref{app:toolcalling_prompt} and the tool suite of App.~\ref{app:toolcalling} (calibrated pinhole, VGGT, calculator), 150 prompts per tool-supported sub-task (seed 42).
Rollouts are then cleaned deterministically: we drop (i) zero-tool-call rollouts, (ii) rollouts containing retries or failed tool responses, and (iii) multi-call rollouts whose final turn promises further measurement; surviving trajectories are de-duplicated, yielding 605 tool trajectories over the seven measurement sub-tasks.

\paragraph{Stage 2: SFT cold start with a no-tool mixture.}
The SFT mixture is 905 samples: the 605 distilled tool trajectories plus 300 \emph{direct} (tool-free) answers covering the six sub-tasks to which tools do not apply (Count, Blink, Gaze, Deform, LS-Head, LS-Body).
The direct half teaches the output convention of non-measurement sub-tasks; restraint on measurement sub-tasks is enforced later by the RL reward (Stage 3).
We fine-tune Qwen3.5-4B for one epoch (57 steps) with the ViT and projector frozen.

\paragraph{Stage 3: tool-aware RLVR.}
Starting from the SFT checkpoint we run GSPO~\citep{gspo} (Group Sequence Policy Optimization), i.e.\ the sequence-level counterpart of the GRPO run of \S\ref{sec:metric_rl}, with $G{=}8$ rollouts per prompt, batch size 8, constant learning rate \mbox{$1\times10^{-6}$}, for 125 steps on 1{,}000 question prompts that are disjoint from both the SFT records and the benchmark split (App.~\ref{app:split}).
The reward is the same deterministic scorer $s(\cdot)$ of \S\ref{sec:protocol} minus a forbidden-call penalty: any tool call issued on the six tool-free sub-tasks subtracts a base penalty chosen so that a forbidden call is strictly worse than a direct answer, making ``answer directly'' the preferred action shape on those sub-tasks.
At evaluation time the trained policy issues tool calls on the measurement sub-tasks while respecting the call/no-call protocol in our runs, i.e.\ the split is learned rather than hard-coded.

\begin{figure}[!t]
\centering
\includegraphics[width=\columnwidth]{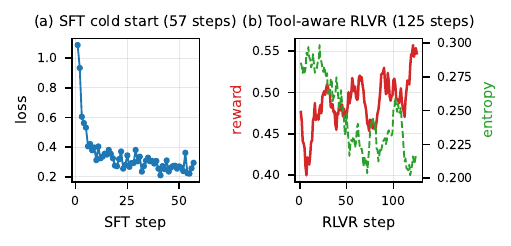}
\caption{\textbf{Training dynamics of the two learning stages.} (a) SFT cold-start loss over its 57 steps; (b) tool-aware RLVR: centred 9-step moving averages (about 72 prompts per smoothed point) of the per-step training reward and of the policy entropy (right axis), over the 125 reported steps; raw per-step values are omitted because each step averages only 8 training prompts, so the smoothed curves show the trend.}
\label{fig:agentic_curves}
\end{figure}

\paragraph{Results.}
Fig.~\ref{fig:agentic_curves} gives the training curves of Stages 2--3; Table~\ref{tab:agentic_rl} reports the three stages of the pipeline under one scorer.
The distilled 4B agent reaches \textbf{49.58} under the same strict scorer and the same 2{,}500 items, versus $26.68$ for Qwen3.5-4B answering directly --- $+22.90$ from the 905-sample cold start and 125 tool-aware RLVR steps.
Gains concentrate exactly where the calibrated tool adds information: Depth 66.3 vs.\ 26.5 (base) and CP-Human 55.6 vs.\ 4.6, with perception sub-tasks preserved (Blink 82.4, Count 81.7).
For reference, the teacher's training-free tool-use regime scores $56.6$ on its own paired 1{,}050-question subset under a different agent protocol (App.~\ref{app:tool_stats}); the two figures do not share a denominator and are therefore not merged into this table.

\begin{table*}[!t]
\centering
\scriptsize
\setlength{\tabcolsep}{3pt}
\caption{\textbf{Agentic tool-use policy on the full 2{,}500-item benchmark.} One row per stage of App.~\ref{app:agentic_rl}; all rows use the same strict scorer as Table~\ref{tab:main_results} (NA$\to$MRA, sequences$\to$tuple scorers, Deform$\to$ACC; unparseable outputs score $0$), greedy decoding, no numbered-box overlay unless the stage emits tool calls; the RLVR row is initialised from the 57-step SFT cold start (Stage~2), whose stand-alone direct-answering score is not separately logged. \textbf{HumanPhys} is the example-weighted composite over all 2{,}500 items.}
\label{tab:agentic_rl}
\begin{tabular*}{\linewidth}{@{}l@{\extracolsep{\fill}}cccccccccccccc@{}}
\toprule
Stage & Depth & Dist & Count & Gaze & Disp & Vel & Traj & Blink & Deform & LS-Head & LS-Body & CP-Room & CP-Human & HumanPhys \\
\midrule
Qwen3.5-4B (base) & 26.45 & 24.15 & 58.00 & 12.30 & 2.70 & 11.95 & \textbf{65.10} & 32.55 & \textbf{25.00} & 23.44 & 51.29 & 8.43 & 4.64 & 26.68 \\
+\,Agentic RLVR ($G{=}8$, 125 steps) & \textbf{66.30} & \textbf{38.10} & \textbf{81.65} & \textbf{25.90} & \textbf{20.70} & \textbf{23.20} & 48.92 & \textbf{82.35} & 24.00 & \textbf{41.20} & \textbf{64.66} & \textbf{59.10} & \textbf{55.64} & \textbf{49.58} \\
\bottomrule
\end{tabular*}
\end{table*}

\paragraph{Ablations, residual gaps, and transparency.}
(i) \emph{Residual gap on Distance.} As an \emph{oracle ceiling probe}, unreachable by the learned policy, feeding the pinhole tool head/foot clicks taken from the ground-truth boxes yields 73.3 on \textit{Distance} and 86.9 on \textit{Depth}, against the policy's 38.1 and 66.3: the remaining gap is click precision on the referent people, not tool correctness, and is the natural target of a grounding-aware reward.
(ii) \emph{Scope.} One training seed, evaluated on the same 2{,}500-item bank as Table~\ref{tab:main_results} (200 items per sub-task, 100 for \textit{Deform}). On the two sub-tasks where tool output carries no extra geometric information and the base is already strong (Traj, Deform) the agent stays below direct base answering, i.e.\ agent-style decoding shifts those answers without compensating them.

\section{Evaluation Prompts and Parsing Statistics}
\label{app:prompts}

Table~\ref{tab:main_results} scores open-ended outputs with deterministic
parsers; this appendix reproduces the exact prompt templates and quantifies
how often low scores are caused by parsing failures rather than by perceptual
errors. All templates below are taken verbatim from the released evaluation
configuration; the SFT training prompts
use the same question text, and for the sequence families the training target
is the bare list while evaluation restores the \texttt{<answer>} wrapper.

\paragraph{Prompt construction.}
Every prompt is the benchmark question followed by (i) a unit/precision hint
for numeric sub-tasks, e.g.\ ``\,(in metres, to 1 decimal place)'' or ``\,(in
degrees)'', derived from the gold value's own decimals, and (ii) one of five
answer-format instructions:
\begin{itemize}\footnotesize
\setlength{\itemsep}{1pt}\setlength{\parsep}{0pt}
\item \textit{Numeric} (\textit{Depth}, \textit{Distance}, \textit{Displacement}, \textit{Velocity}, \textit{Gaze}, \textit{Count}, \textit{Blink}): \texttt{Output ONLY the numerical answer inside <answer></answer>, e.g.\ <answer>12.3</answer>. Do not explain.}
\item \textit{Multiple choice} (\textit{Deform}, and candidate-list variants): \texttt{The image shows candidate options labeled A, B, C, D. Output ONLY the single correct option letter inside <answer></answer>, e.g.\ <answer>B</answer>. Do not explain.}
\item \textit{Camera-pose sequence} (\textit{CP-Room}, \textit{CP-Human}): \texttt{Answer ONLY as a Python list of two-item lists, one per step. Each element MUST be exactly [action, magnitude] --- a 2-item list, never a tuple. action MUST be one of these nine bare string tokens: 'yaw','pitch','roll' (a rotation, magnitude = integer degrees) or '+X','+Y','+Z','-X','-Y','-Z' (a translation, magnitude = metres). \dots Example: [['yaw', 30], ['+Y', 0.4], ['-Y', 0.2]]. Put the full list inside <answer></answer>.}
\item \textit{Trajectory sequence} (\textit{Traj}): \texttt{Answer ONLY as a Python list of two-number lists, one per step --- never a tuple. Example: [[0, 1.0], [10.0, 0.7], [-20.0, 0.6]]. Put the full list inside <answer></answer>.}
\item \textit{Light--shadow coordinate sequence} (\textit{LS-Head}, \textit{LS-Body}): \texttt{Answer ONLY as a Python list of coordinate lists, one per subsequent frame (exclude the given first-frame light). Example: [[0, -1, 1], [1, 0, 2]]. Put the full list inside <answer></answer>.}
\end{itemize}

\paragraph{Parsing rules.}
The parser takes the \emph{last} \texttt{<answer>}$\dots$\texttt{</answer>}
block; a missing or truncated block scores $0$. Numeric blocks are read as a
single float; trajectory blocks must be a list of two-number lists (turn
degrees, distance metres); camera-pose blocks must be
\texttt{[token, magnitude]} pairs over the nine-token action vocabulary;
light--shadow blocks must be integer triples; MC blocks must contain one
option letter. Any structural mismatch scores $0$, and the RL reward of
Appendix~\ref{app:rl_details} applies exactly the same rule.

\begin{table*}[!t]
\centering
\scriptsize
\setlength{\tabcolsep}{3pt}
\caption{\textbf{Unparseable-output rate (\%) per model and sub-task}, i.e.\ the
fraction of the benchmark items whose \texttt{<answer>} block fails the
structural parser of this appendix (``--'' $=$ 0.0\%). Computed from the stored
model outputs of Table~\ref{tab:main_results} (upper block) and of the five SFT
checkpoints of Table~\ref{tab:sft_subtasks} (lower block). The upper block audits
the six open-weight zero-shot models that were served through the local inference
stack which also produced the per-item sample logs used here; GPT-6-Sol and the
five API-served models are queried through hosted endpoints and are therefore not
covered by this format audit. The \textit{Trajectory} column separates schema-following failure from
perceptual failure: Qwen2.5-VL-7B and Spatial-SSRL-7B score 3.53 and 5.76
there mainly because $93.5\%$ / $88.5\%$ of their outputs are not parseable
two-number lists, whereas the SFT and RL checkpoints of
Tables~\ref{tab:sft_subtasks} and~\ref{tab:rl_subtasks} reach 67--70 there with
$0\%$ unparseable \textit{Trajectory} outputs; the only format failures of the SFT
runs lie on the camera-pose sequences (lower block).}
\label{tab:unparseable}
\begin{tabular*}{\linewidth}{@{}l@{\extracolsep{\fill}}ccccccccccccc@{}}
\toprule
Model & Depth & Dist & Count & Gaze & Disp & Vel & Traj & Blink & Deform & LS-Head & LS-Body & CP-Room & CP-Human \\
\midrule
Qwen3.5-4B (base) & -- & -- & -- & -- & -- & -- & -- & -- & -- & 2.0 & 10.5 & 2.0 & 26.5 \\
Qwen2.5-VL-7B & -- & -- & -- & -- & -- & -- & 93.5 & -- & -- & -- & -- & -- & -- \\
SpaceR & -- & -- & -- & -- & -- & -- & -- & -- & -- & -- & -- & -- & -- \\
SpatialEvo & -- & -- & -- & -- & -- & -- & -- & -- & -- & 1.0 & 10.0 & -- & -- \\
SSRL (Spatial-SSRL-7B) & -- & -- & -- & -- & -- & -- & 88.5 & -- & -- & 3.0 & 4.0 & -- & -- \\
ViLaSR & -- & -- & -- & -- & -- & -- & -- & -- & -- & 0.5 & 1.5 & -- & -- \\
\midrule
SFT \ding{172} & -- & -- & -- & -- & -- & -- & -- & -- & -- & -- & -- & -- & 2.0 \\
SFT \ding{173} & -- & -- & -- & -- & -- & -- & -- & -- & -- & -- & -- & -- & 1.5 \\
SFT \ding{174} & -- & -- & -- & -- & -- & -- & -- & -- & -- & -- & -- & -- & 19.0 \\
SFT \ding{175} & -- & -- & -- & -- & -- & -- & -- & -- & -- & -- & -- & 1.5 & 16.0 \\
SFT \ding{176} & -- & -- & -- & -- & -- & -- & -- & -- & -- & -- & -- & 90.0 & 98.0 \\
\bottomrule
\end{tabular*}
\end{table*}

\paragraph{Reading the table.}
Two observations follow. First, the numeric and MC sub-tasks are
parsing-clean for every model (0.0\% throughout), so the low scores of weak
models there reflect genuine perceptual limits, not instruction-following
artefacts. Second, the sequence families are where format and perception
confound: the base Qwen3.5-4B fails to emit a parseable body-view pose list on
$26.5\%$ of \textit{CP-Human} items and a parseable coordinate list on
$10.5\%$ of \textit{LS-Body} items, which accounts for part of its
near-floor scores on those columns; the spatially fine-tuned checkpoints
largely repair this (SpaceR and ViLaSR are at or below $1.5\%$); the two
\textit{Trajectory} outliers and the camera-pose failures of the SFT runs are
itemised in the caption. Each sequence-family column of
Table~\ref{tab:main_results} equals the mean scorer value over the parseable
subset times the parseable fraction, i.e.\
$\mathrm{strict}=\mathrm{mean}_{\mathrm{parseable}}\times(1-u)$ with $u$ taken
from this table --- for Qwen3.5-4B, \textit{LS-Body}
$57.31\times(1-0.105)=51.29$.

\section{Human-Reference Study Interface}
\label{app:human_study}

\begin{figure*}[!t]
\centering
\includegraphics[width=\linewidth]{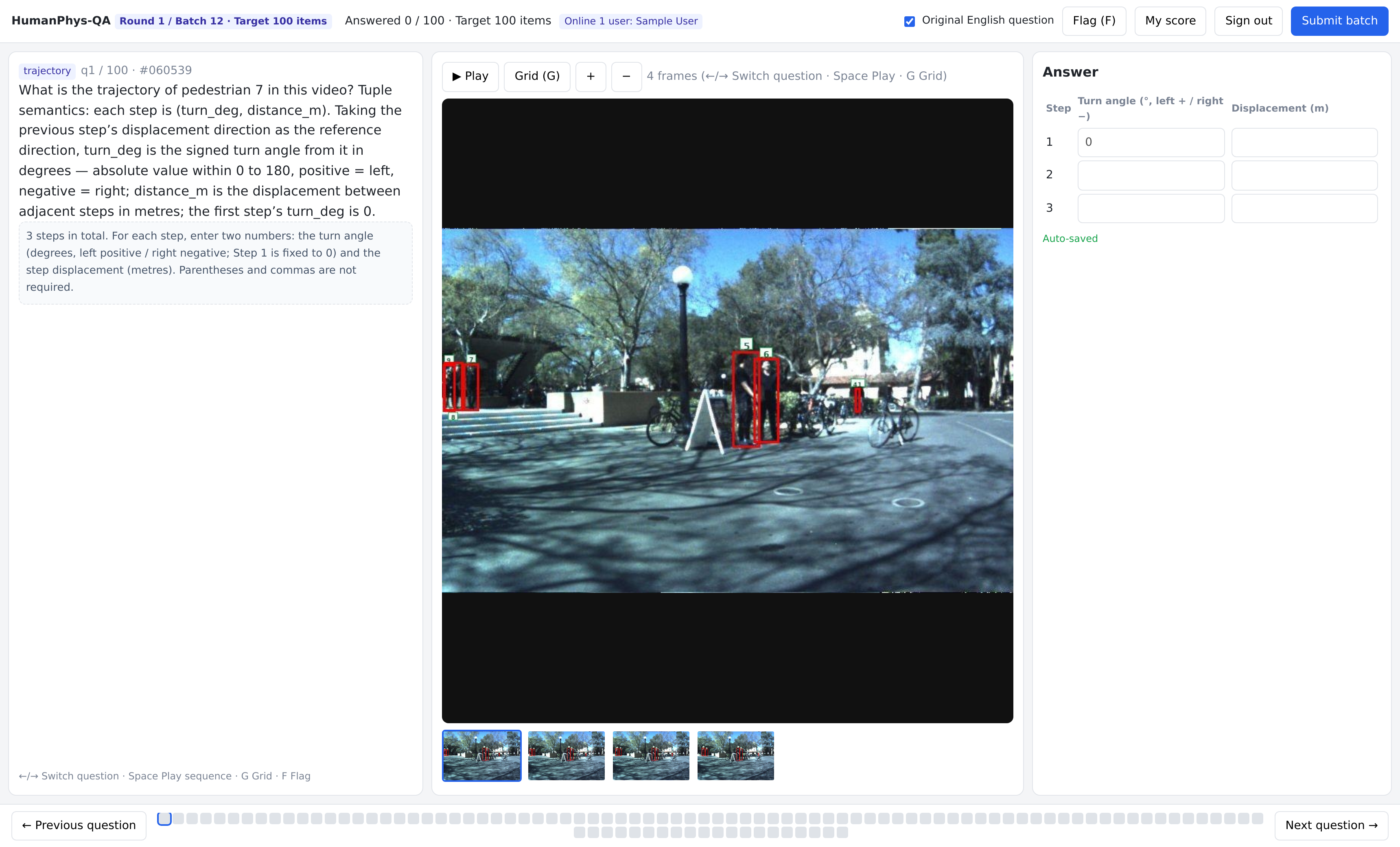}
\caption{\textbf{The human-reference answering interface.} A volunteer is
shown mid-way through a \textit{Trajectory} item. Left: the question in
English plus the answer-schema hint. Centre: the current frame with the
playback, tiling and zoom controls above and the thumbnail strip of all four
frames below; the highlighted thumbnail is the frame on display.
Right: the structured answer grid (one turn angle and one displacement per
step) with the auto-save indicator. Bottom: the batch navigator with
per-question completion states. Frames carry the same pre-baked track-box
overlays that the released items provide.}
\label{fig:human_panel_gui}
\end{figure*}

The \emph{Human} row of Table~\ref{tab:main_results} and the timing
column of Table~\ref{tab:human_time} were collected through a dependency-free
single-page web application (Figure~\ref{fig:human_panel_gui}); each volunteer
receives question batches that are pairwise disjoint across concurrent users,
so no two panelists answer the same item in the same round. Answers are entered in a schema-matched widget per
sub-task --- a single number, an option letter, or a per-frame/per-step row of
the sub-task's native tuple (light coordinates, turn--displacement pairs, or pose
actions) --- and are scored server-side with exactly the deterministic scorer of
\S\ref{sec:protocol}.

\paragraph{Playback and frame access.}
Stimuli are presented as an \emph{ordered strip of still frames} rather than as
an encoded video, with playback, direct frame stepping, side-by-side tiling and
zoom. A volunteer can therefore pause, step and compare shadows, body
configurations or blink onsets at leisure, which is why the median response
times of Table~\ref{tab:human_time} are long.

\section{Limitations}
\label{app:limitations}

The most visible limitation of HumanPhys is the size of its smallest family.
\textit{Deform} contributes only 267 supervised pairs, because every item
requires a DECAF monocular face--hand fit whose contact deformation survives
manual inspection; consistently, across the four HumanPhys-containing SFT mixtures of
Table~\ref{tab:sft_subtasks} \textit{Deform} stagnates at 29.00--31.00, below
both the 32.00 of the no-HumanPhys control (the fifth row of that table) and the 45.00 of zero-shot
GPT-6-Sol, i.e.\ the bottleneck is the scarcity of contact-deformation
supervision rather than the task formulation. We regard
scaling this family --- more captures, more contact sites, and fits with
lower residual error --- as the most concrete direction for future work, and
we caution that \textit{Deform} scores should be read with this sample size in
mind. Beyond \textit{Deform}, every label inherits residual noise from its
upstream annotation or fit, which we quantify where possible
(Appendix~\ref{app:taskdetails}). And because the strict static-camera filter
admits only one densely tracked JRDB sequence, all five JRDB sub-tasks (1{,}000
of the 2{,}500 evaluation items) are evaluated on a single scene: within-scene
diversity is audited in
Table~\ref{tab:tressider_diversity}, but cross-scene generalisation on these
sub-tasks is not measured. Finally, the human reference is an indicative context
rather than a matched ceiling (Appendix~\ref{app:human_study}).

\end{document}